\documentclass{article}
 
\usepackage{adjustbox}
\usepackage{longtable}

\usepackage{arxiv}

\usepackage{bm}
\usepackage{amssymb,amsthm,mathtools}

\usepackage[utf8]{inputenc} % allow utf-8 input
\usepackage[T1]{fontenc}    % use 8-bit T1 fonts
\usepackage{url}            % simple URL typesetting
\usepackage{booktabs}       % professional-quality tables
\usepackage{amsfonts}       % blackboard math symbols
\usepackage{nicefrac}       % compact symbols for 1/2, etc.
\usepackage{microtype}      % microtypography
\usepackage{multirow}
\usepackage{xcolor}
\usepackage{enumitem}
\usepackage{graphicx}
\usepackage{float}
\usepackage{orcidlink}
\usepackage[most]{tcolorbox}
\tcbuselibrary{breakable,listings}
\usepackage{array}
\newcolumntype{R}[1]{>{\raggedleft\arraybackslash}p{#1}}
\newcolumntype{L}[1]{>{\raggedright\arraybackslash}p{#1}}

\usepackage[numbers]{natbib}
\usepackage{caption}
\usepackage{amsmath}
\usepackage{rotating}
\usepackage{cleveref}
\crefname{appendixtable}{Appendix Table}{Appendix Tables}
\Crefname{appendixtable}{Appendix Table}{Appendix Tables}
\crefname{appendixfigure}{Appendix Figure}{Appendix Figures}
\Crefname{appendixfigure}{Appendix Figure}{Appendix Figures}

\makeatletter
\renewcommand{\footnotesize}{\@setfontsize\footnotesize{8pt}{10pt}}
\makeatother

\makeatletter
\renewcommand{\scriptsize}{\@setfontsize\scriptsize{7pt}{9pt}}
\makeatother

\definecolor{VUB_blauw}{rgb}{0.1529, 0.2667, 0.5529}
\definecolor{mygrey}{gray}{0.55}
\newtcblisting{promptbox}{
  listing only,
  breakable,
  colback=white,
  colframe=mygrey,
  boxrule=0.5pt,
  arc=1mm,
  left=1mm,
  right=1mm,
  top=1mm,
  bottom=1mm,
  listing options={
    basicstyle=\ttfamily\scriptsize,
    breaklines=true,
    breakatwhitespace=false,
    columns=fullflexible,
    keepspaces=true,
    showstringspaces=false
  }
}
\usepackage{hyperref}       % hyperlinks
\hypersetup{
    colorlinks,%
    citecolor=VUB_blauw,%
    filecolor=VUB_blauw,%
    linkcolor=VUB_blauw,%
    urlcolor=VUB_blauw
}

\AddToHook{shipout/background}{%
  \ifnum\value{page}=1 
    \pagenumbering{Roman} 
    \setcounter{page}{1} 
  \fi
  \ifnum\value{page}=2 
  \pagenumbering{arabic} 
  \setcounter{page}{2} 
  \fi
}

\title{How Much Does a Reasoning Summary Reveal? \\ An Observability Ladder for Large Language Models}
\runningtitle{An Observability Ladder for Large Language Models}

\author{
  Andres Algaba\textsuperscript{1,2,*} \\ 
  \orcidlinkc{0000-0002-0532-3066} \\
  \And
  Francesca Carlon\textsuperscript{1,2} \\ 
  \orcidlinkc{0009-0004-2152-2745} \\
  \And
  Lynn Delcon\textsuperscript{1,2} \\ 
  \orcidlinkc{0009-0005-0815-4674} \\
  \And
  Marthe Ballon\textsuperscript{1,2} \\ 
  \orcidlinkc{0009-0000-4586-234X} \\
  \And
  Bert Verbruggen\textsuperscript{1,2} \\ 
  \orcidlinkc{0000-0001-9776-2420} \\
  \And
  Vincent Ginis\textsuperscript{1,2,3} \\ 
  \orcidlinkc{0000-0003-0063-9608} \\
  \and
  \textsuperscript{1}Data Analytics Lab, Vrije Universiteit Brussel, Pleinlaan 5, 1050 Brussel, Belgium \\ 
  \textsuperscript{2}imec-SMIT, Vrije Universiteit Brussel, Pleinlaan 9, 1050 Brussel, Belgium \\ 
  \textsuperscript{3}School of Engineering and Applied Sciences, Harvard University, Cambridge, Massachusetts 02138, USA
}

\begin{document}

\maketitle
\renewcommand{\thefootnote}{}
\renewcommand{\theHfootnote}{correspondingauthor}
\footnotetext{*Corresponding author: \href{mailto:andres.algaba@vub.be}{andres.algaba@vub.be} \\}
\renewcommand{\thefootnote}{\arabic{footnote}}
\renewcommand{\theHfootnote}{\arabic{footnote}}
\thispagestyle{plain}

\begin{abstract}
Large language models often show users a final response and a short reasoning summary while the full reasoning trace stays hidden.
We introduce an observability ladder that holds each completed run fixed and varies only what a reader inspects to judge whether the answer is correct: the response, a self-summary the model writes from the trace, the trace itself, and internal signals, each with and without the prompt.
Across three benchmarks and five open-weight Qwen3 and gpt-oss models, we train matched linear correctness predictors on each access level.
Without the prompt, summaries carry most of the trace's ranking signal (mean AUROC $0.774$ versus $0.813$) and add $+0.156$ over the response alone.
With the prompt visible, the summary's gain collapses to $+0.019$, while the trace still adds $+0.041$.
Even at equal length, the trace's last words predict correctness as well as summaries, or slightly better, and carry denser and more discriminative uncertainty and self-correction cues.
On MMLU-Pro questions with both correct and incorrect runs, linear summary readers are near chance and trace readers retain only modest signal, both with and without the prompt (prompt-withheld AUROC $0.503$--$0.545$ versus $0.544$--$0.590$).
With the prompt withheld, a GPT-5-mini reader recovers substantially more signal from both summaries and traces on gpt-oss-20b, and even then the trace keeps a small $+0.034$ advantage.
Much of the linear readers' trace signal is associated with length.
In the common case where users already hold the prompt, summaries are less helpful than the full trace for monitoring correctness.
Monitorability is thus a joint property of the display and the reader, so any monitorability claim, including for faithfulness, should specify both.
\end{abstract}


\begin{figure}[t]
\centering
\includegraphics[width=\textwidth]{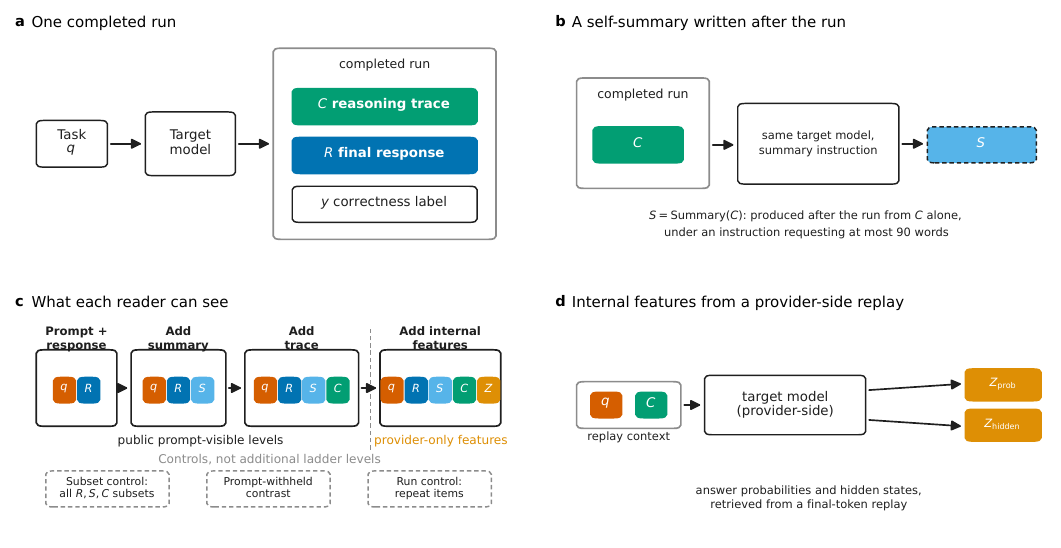}
\caption{\textbf{The observability ladder holds one completed run fixed and changes only the evidence used to predict whether its answer is correct.} \textbf{a}, The run contains prompt $q$, full reasoning trace $C$, final response $R$, and correctness label $y$; the label is only the prediction target. \textbf{b}, Afterward, the same model receives the trace alone and writes self-summary $S$ under an instruction requesting at most 90 words. \textbf{c}, Public readers first see the prompt and response, then the summary, then the trace. Prompt-withheld and repeated-run comparisons are controls, not extra access levels. \textbf{d}, A separate analysis with direct model access adds answer probabilities and hidden states, denoted $Z$.}
\label{fig:observability_ladder}
\end{figure}

\section{Introduction}

Reasoning traces can expose errors or misbehavior absent from a model's final response~\cite{korbak2025chainthoughtmonitorabilitynew,baker2025monitoring,guan2025monitoring}.
Yet interfaces differ in what they reveal.
In commercial deployments such as those from OpenAI, Google, and Anthropic, a user may receive only the final response or a short provider summary~\cite{openai2024learningreasoning,openai2026reasoningdocs,google2026geminithinking,anthropic2026extendedthinking}, while the full trace, answer probabilities, and hidden states require deeper access~\cite{openai2025detectingmisbehavior,openai2025evaluatingcotmonitorability,guo2017calibration,kadavath2022language,burns2023discovering,azaria2023internal}.
Evidence from a full trace therefore does not establish what a user can learn from a response or summary.
What any display reveals also depends on its reader, so our claims concern the tested monitoring methods rather than the text alone.

We ask three plain questions.
First, how well does a short summary predict correctness compared with the full trace?
Second, how much does the summary add when the reader does or does not see the prompt?
Third, can either text distinguish correct from incorrect runs of the same question?
To answer them, after each run the target model writes a summary from its trace alone under an instruction requesting at most 90 words.
This controlled summary is our proxy for deployed provider summaries, which may differ in context, instructions, models, or filters.
We compare matched linear correctness readers on the response, summary, full trace, and a small set of signals requiring direct model access.

We evaluate five open-weight models, Qwen3-4B, Qwen3-8B, Qwen3-14B, gpt-oss-20b, and gpt-oss-120b~\cite{qwen2025qwen3collection,yang2025qwen3,openai2025gptoss,agarwal2025gpt}, on GPQA-Diamond, MMLU-Pro, and Omni-MATH-2-Filtered~\cite{rein2023gpqa,wang2024mmlu,ballon2026benchmarks}.
The first two benchmarks are multiple choice. Omni-MATH-2-Filtered contains olympiad mathematics problems that require a written answer; GPT-5-mini grades them following the benchmark's procedure.
Open weights let us examine every access level on the same completed runs.
Under the tested linear readers with the prompt withheld, summary-only AUROC is $0.774$, compared with $0.813$ for the full trace, and adding the trace after the response and summary gives $+0.035$ AUROC.
At the same word count, extracts from the trace's end match or slightly outperform the self-summaries and carry denser and more discriminative uncertainty and self-correction cues.
Showing the prompt reduces the summary gain from $+0.156$ to $+0.019$.
A prompt-only reader already reaches AUROC $0.734$.
In repeated-run comparisons on questions with both correct and incorrect runs, linear summary readers are near chance and linear trace readers retain only modest signal, whether the prompt is visible or withheld.
For gpt-oss-20b, a GPT-5-mini reader recovers substantially more signal from both summaries and traces when the prompt is withheld, but the trace still performs slightly better than the summary.
Ranking runs across a mixed set of questions and judging one particular run are therefore different tasks.

Prior work uses intermediate reasoning to improve or check answers~\cite{wei2022chain,kojima2022zeroshotcot,cobbe2021training,lightman2024lets,ballon2025relationship}, although written reasoning need not faithfully describe the computation that produced the answer~\cite{lanham2023faithfulness,turpin2024language,lyu2023faithful,chen2025reasoning}.
Other studies ask whether traces reveal unsafe intent, cheating, obfuscation, or failures hidden from final responses~\cite{korbak2025chainthoughtmonitorabilitynew,baker2025monitoring,guan2025monitoring,wang2026monitorbench,vonarx2025recentfrontierrewardhacking,openai2025detectingmisbehavior,openai2025evaluatingcotmonitorability}.
Recent results also show that conclusions can change with the reader~\cite{emmons2025pragmatic,ballon2026benchmarks}, that structural patterns in a trace can help compare runs of the same question~\cite{lee2026reasonops}, and that shortening traces can remove useful clues~\cite{little2026length}.
Studies of reasoning summaries test whether they preserve information that another model can use~\cite{zhang2026steal}, including a model less capable than the one that produced the trace~\cite{roytburg2026legibility}.
Researchers have also used answer probabilities and hidden states to predict correctness or truthfulness~\cite{geifman2017selective,kamath2020selective,kadavath2022language,guo2017calibration,jiang2021know,varshney2023post,alain2016understanding,conneau2018cram,belinkov2022probing,burns2023discovering,marks2023geometry,azaria2023internal,zou2023representation,boppana2026reasoningtheater,mirtaheri2026rationalization,ghasemabadi2025gnosis}.

Our observability ladder changes only what a reader can inspect while keeping the completed run and correctness label fixed.
It compares the final response, self-summary, full trace, and additional internal signals available only with direct model access, both with and without the prompt.
Under the readers studied here, summaries help most when the prompt is withheld.
In the common case where users already hold the prompt, the summary is less helpful than the full trace for monitoring correctness.
For properties such as reward hacking, the decisive evidence can sit in the trace alone~\cite{baker2025monitoring,korbak2025chainthoughtmonitorabilitynew}, and judging faithfulness requires more than the visible text itself~\cite{lanham2023faithfulness,turpin2024language}.
Monitorability is thus a joint property of the display and the reader, and any monitorability claim, including for faithfulness, should state both.

\section{The observability ladder}
\label{sec:ladder_design}

For each benchmark item $i$ and target model $m$, the model produces one completed reasoning run (\Cref{fig:observability_ladder}a).
From that run we record the task prompt $q_i$, visible final response $R_{i,m}$, full reasoning trace $C_{i,m}$, post-hoc self-summary $S_{i,m}$, provider-side replay features $Z_{i,m}$, and binary correctness label
\[
y_{i,m}=\mathbf{1}[\text{final answer is correct}].
\]
Correctness is the monitored property because the labels are defined externally to the monitor and held fixed across the displays being compared.
For multiple-choice tasks, $R_{i,m}$ is the final response text, which is usually but not always a bare option letter. The extracted option is stored separately for grading.
For open-response mathematics, $R_{i,m}$ is the visible final response and the extracted boxed expression, final expression, or concise answer is stored separately.
The same target model generates $S_{i,m}$ from $C_{i,m}$ alone after the run, under an instruction requesting at most 90 words (\Cref{fig:observability_ladder}b).
The grading procedure, the verbatim summary instruction, and an example summary are in the \hyperref[sec:methods]{Methods}.
The provider-side replay feature set $Z_{i,m}$ contains answer-probability features and the last-layer hidden state at the final token, together with a small set of summaries of those quantities (\Cref{fig:observability_ladder}d).
This prompt-conditioned replay captures only a narrow subset of what model internals can reveal~\cite{belinkov2022probing,burns2023discovering,marks2023geometry,zou2023representation}.

A display is the text made available to the monitor (\Cref{fig:observability_ladder}c).
In the main prompt-visible setting, the three public access levels are
\[
\mathcal{O}^{q}_{1}(i,m)=q_i\oplus R_{i,m},
\quad
\mathcal{O}^{q}_{2}(i,m)=q_i\oplus R_{i,m}\oplus S_{i,m},
\]
\[
\mathcal{O}^{q}_{3}(i,m)=q_i\oplus R_{i,m}\oplus S_{i,m}\oplus C_{i,m}.
\]
The final, non-public level adds provider-side replay features:
\[
\mathcal{O}^{q}_{4}(i,m)=q_i\oplus R_{i,m}\oplus S_{i,m}\oplus C_{i,m}\oplus Z_{i,m}.
\]
The matched prompt-withheld contrast does not explicitly show $q_i$:
\[
\mathcal{O}_{1}(i,m)=R_{i,m},
\quad
\mathcal{O}_{2}(i,m)=R_{i,m}\oplus S_{i,m},
\quad
\mathcal{O}_{3}(i,m)=R_{i,m}\oplus S_{i,m}\oplus C_{i,m}.
\]
At the public levels, $\oplus$ joins the visible text. At the provider-side level, it appends numerical features extracted from a prompt-conditioned final-token replay. These replay features require direct model access and are not part of a public display. \Cref{tab:notation} summarizes the components and access levels.
Because $S$ is derived from $C$, an ordered increment depends on whether the summary or trace is added first.
We therefore fit public monitors to all seven subsets
\[
\{R,\;S,\;C,\;R+S,\;R+C,\;S+C,\;R+S+C\}.
\]
For each component, Shapley attribution averages its change in AUROC when added to every subset that does not already contain it.
We use the Shapley values only to describe how predictive performance beyond chance is allocated among $R$, $S$, and $C$~\cite{shapley1953value}; they are not an information decomposition.

The prompt-visible ladder represents the most realistic setting when the monitor has the prompt.
The prompt-withheld ladder remains important as an output-only contrast, especially when only responses and summaries are available. It does not include $q$ in the display, but $S$ and $C$ can still paraphrase the prompt or expose item-difficulty cues.
Additional interpretive controls remove long spans copied from the prompt, mask stated answers together with option text and prompt-derived numbers, or compare independent runs of the same MMLU-Pro item.
The within-item metric in the repeated-run comparison tests how a monitor trained across items transfers to run-level discrimination after item-level cues such as topic, format, answer structure, and difficulty are removed.
The prompt-conditioned and repeated-run controls are complementary but not equivalent. Adding $q$ tests whether a display contributes beyond an explicitly visible prompt in across-item evaluation. Holding $q$ fixed removes between-item variation, but does not reproduce a jointly prompt-conditioned monitor because the encoder can represent prompt--response and prompt--trace interactions. We therefore run the prompt-visible repeated-run controls with the primary surface-plus-Cohere monitors under two objectives, transferred across-item fitting (\Cref{tab:multirun_prompt_visible}) and direct item-conditional fitting (\Cref{tab:item_conditional_pairwise_prompt}).
We repeat both with the open surface-plus-TF-IDF monitors as encoder sensitivities (\Cref{tab:open_prompt_multirun,tab:item_conditional_pairwise_open}).

The experiment crosses five models with three benchmarks, giving 15 model--benchmark settings (\Cref{tab:task_taxonomy,tab:model_access}).
Within a setting, every access level uses the same held-out items and splits, so differences between levels are paired on a fixed set of completed runs (\Cref{tab:generation_defaults}).
For a public display, the monitor computes surface features, including word and sentence counts, digit and mathematical-symbol density, and rates of uncertainty, self-correction, and answer-framing markers~\cite{vanhoyweghen2025lexical}.
These features are concatenated with a 512-dimensional Cohere embed-v4.0 text embedding after train-only preprocessing~\cite{reimers2019sentencebert,cohere2026embed}.
The provider-side level appends replay-derived answer-probability and hidden-state features from the target model~\cite{guo2017calibration,kadavath2022language,burns2023discovering,azaria2023internal}.
An $\ell_2$-regularized logistic probe maps the resulting vector to a correctness score (\Cref{tab:probe_specification}).
AUROC is the probability that the monitor assigns a higher score to a randomly chosen correct run than to a randomly chosen incorrect run. A value of $0.5$ is chance.
The model--benchmark setting is the unit of inference, and the reported estimates average paired contrasts across the 15 settings.
The full feature specification is in \Cref{tab:feature_boundary}, and procedural details are in the \hyperref[sec:methods]{Methods}.

Our hypothesis concerns the mean paired increments; individual settings may be flat or negative.
Although the access sets are conceptually nested, each primary display is embedded and fitted separately. Finite-sample estimation and validation can therefore give a higher-access reader a lower AUROC. The ladder orders the evidence available to a reader but does not require fitted performance to increase at every step.
For the main prompt-visible ladder, we report
\[
q\Delta_S=\mathrm{AUROC}(q+R+S)-\mathrm{AUROC}(q+R),
\]
\[
q\Delta_C=\mathrm{AUROC}(q+R+S+C)-\mathrm{AUROC}(q+R+S),
\]
\[
q\Delta_Z=\mathrm{AUROC}(q+R+S+C+Z)-\mathrm{AUROC}(q+R+S+C),
\]
with $q\Delta_{\mathrm{public}}=\mathrm{AUROC}(q+R+S+C)-\mathrm{AUROC}(q+R)$ and $q\Delta_{\mathrm{full}}=\mathrm{AUROC}(q+R+S+C+Z)-\mathrm{AUROC}(q+R)$.
For the prompt-withheld contrast, we also report
\[
\Delta_S=\mathrm{AUROC}(R+S)-\mathrm{AUROC}(R),
\]
\[
\Delta_C=\mathrm{AUROC}(R+S+C)-\mathrm{AUROC}(R+S),
\]
\[
\Delta_Z=\mathrm{AUROC}(R+S+C+Z)-\mathrm{AUROC}(R+S+C),
\]
and the public and full gains $\Delta_{\mathrm{public}}=\mathrm{AUROC}(R+S+C)-\mathrm{AUROC}(R)$ and $\Delta_{\mathrm{full}}=\mathrm{AUROC}(R+S+C+Z)-\mathrm{AUROC}(R)$.
The central summary-versus-trace comparison is the trace increment after the summary is visible. A small value means that the short-summary monitor performs close to the full-trace monitor in this ordered comparison. It does not quantify how much trace information the summary contains.
Because $Z$ is computed with both $q$ and $C$ in context, it sits outside the public ladders even when appended to prompt-withheld displays.
\Cref{fig:observability_ladder} summarizes these comparisons.

\begin{figure}[t]
\centering
\includegraphics[width=\textwidth]{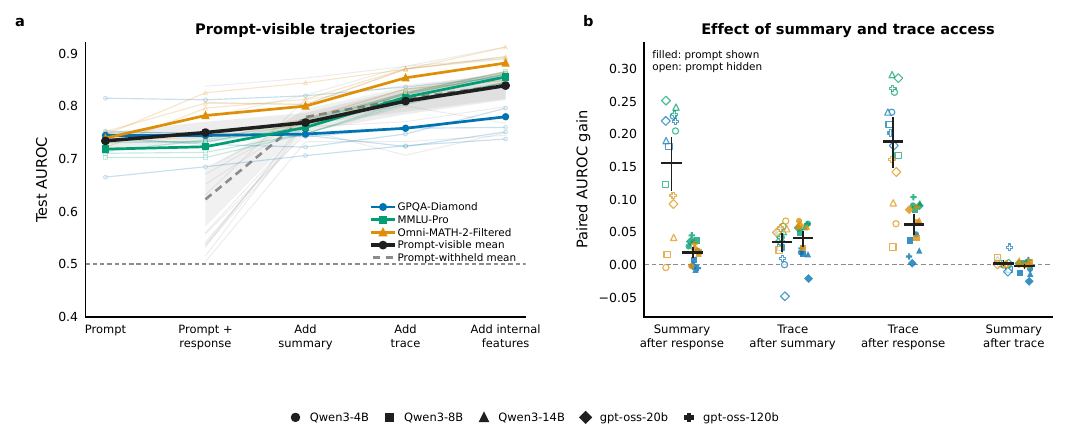}
\caption{\textbf{The summary adds little once the prompt is visible.} Results cover five models and three benchmarks. \textbf{a}, Correctness ranking as readers receive the prompt, response, summary, trace, and then internal features. Thin colored lines show the 15 settings and thick colored lines the benchmark means, all with the prompt visible. The black line is the prompt-visible mean and the dashed gray line the matched prompt-withheld mean. Shaded bands give 95\% descriptive setting-bootstrap intervals for the black prompt-visible and dashed prompt-withheld mean lines. \textbf{b}, Effects of summary and trace access. Filled points show readers that see the prompt and open points show readers that do not. Black ticks show setting means with 95\% descriptive setting-bootstrap intervals. Adding the summary after the response gains $+0.019$ with the prompt versus $+0.156$ without it. Adding the summary after the trace changes AUROC by about zero in both cases. Exact values and intervals are in \Cref{tab:prompt_ladder_contrasts,tab:ladder_delta_uncertainty}.}
\label{fig:main_ladder}
\end{figure}

\section{The summary adds little once the prompt is visible}
\label{sec:ladder_result}

\Cref{fig:main_ladder}a reports mean prompt-visible trajectories for each benchmark and overall, together with the matched prompt-withheld mean, on the same held-out items and splits, and \Cref{fig:main_ladder}b reports the paired increments within the same runs. Prompt-visible setting values are in \Cref{tab:prompt_controls_15cell}; the prompt-withheld setting values and increments are in \Cref{tab:main_ladder_15cell,tab:ladder_deltas_15cell}, and the pooled comparison is in \Cref{tab:prompt_ladder_contrasts}.

The prompt alone is already predictive, reaching mean AUROC $0.734$ across the 15 settings (\Cref{fig:main_ladder}a). Because each target model is evaluated separately, this score mainly reflects which questions tend to be easier or harder for that model.
With the prompt visible, adding the summary to the response gives a small mean gain, $q\Delta_S=+0.019$ AUROC (\Cref{fig:main_ladder}b; 95\% descriptive setting-bootstrap interval $[+0.011,+0.027]$ over the 15 observed settings, \Cref{tab:ladder_delta_uncertainty}); the mixed-model sensitivity interval for this increment includes zero (\Cref{tab:hierarchical_deltas}).
Adding the full trace after the prompt, response, and summary gives $q\Delta_C=+0.041$ (descriptive interval $[+0.028,+0.052]$).
The total public gain from $q+R$ to $q+R+S+C$ is therefore $q\Delta_{\mathrm{public}}=+0.060$ (descriptive interval $[+0.040,+0.077]$), and the replay features add $q\Delta_Z=+0.029$ (descriptive interval $[+0.023,+0.036]$), for $q\Delta_{\mathrm{full}}=+0.089$ (descriptive interval $[+0.066,+0.110]$).
The $Z$ increment measures the gain from this specific replay feature set (\Cref{fig:internal_decomposition,tab:internal_decomposition_15cell}).
These gains improve across-item correctness ranking over the held-out mixture, but they are much smaller than the prompt-withheld ladder.

The prompt-withheld contrast explains why response-only baselines can overstate the apparent value of summaries on multiple-choice tasks (\Cref{fig:main_ladder}b, open points).
Without $q$, the response is often a bare option letter and has mean AUROC $0.623$, while the prompt-only monitor reaches $0.734$.
Moving from $R$ to $R+S+C+Z$ gains $\Delta_{\mathrm{full}}=+0.217$ AUROC (95\% descriptive setting-bootstrap interval $[+0.177,+0.257]$ over the observed settings, \Cref{tab:ladder_delta_uncertainty}), positive in every setting.
The largest step is $\Delta_S=+0.156$. Prompt visibility shrinks that same summary increment to $+0.019$, a paired difference of $-0.137$ (95\% descriptive setting-bootstrap interval $[-0.177,-0.095]$; \Cref{tab:prompt_ladder_contrasts,tab:ladder_delta_uncertainty}).
By contrast, the trace increment is similar in both cases, $\Delta_C=+0.035$ prompt-withheld and $q\Delta_C=+0.041$ prompt-visible.
The order controls in \Cref{fig:main_ladder}b and \Cref{tab:prompt_ladder_contrasts} give the direct, non-cumulative contrasts. Adding the trace directly to $q+R$ gains $+0.062$, and adding the summary after the trace changes AUROC by $-0.002$. The direct $q{+}R{+}C$ versus $q{+}R{+}S$ contrast is $+0.043$, positive in all 15 settings (95\% descriptive setting-bootstrap interval $[+0.033,+0.052]$; \Cref{tab:ladder_delta_uncertainty}).
Thus, most of the large output-only summary increment overlaps with predictive information already available to a prompt-visible monitor; traces retain a small residual increment after prompt and summary are visible.
The repeated-run comparison in \Cref{sec:within_item_result} tests how well a reader distinguishes runs of the same question, which is a different quantity.

Because the summary is generated from the trace, ordered increments do not uniquely attribute the public signal to $S$ or $C$. The output-only all-subset analysis in \Cref{sec:public_monitor_result} and the following prompt-conditioned factorials address that order dependence.
The complete primary surface-plus-Cohere factorial gives mean full-display AUROC $0.809$. With $q$ as the baseline, its prompt-conditioned Shapley allocation is $+0.005$ for $R$, $+0.013$ for $S$, and $+0.057$ for $C$ (\Cref{tab:prompt_conditioned_shapley}).
A secondary open surface-plus-TF-IDF factorial reproduces this pattern, with $+0.000$ for adding $S$ after $q+R+C$ and a Shapley allocation of $+0.033$ for $S$ versus $+0.078$ for $C$ (\Cref{tab:open_prompt_factorial,tab:open_prompt_contrasts,tab:open_prompt_shapley}). These paths concern set inclusion, not physical presentation order.

The prompt-visible gains differ by benchmark (\Cref{fig:main_ladder}a,b).
Mean $q\Delta_{\mathrm{full}}$ is $+0.036$ on GPQA-Diamond, $+0.132$ on MMLU-Pro, and $+0.100$ on Omni-MATH-2-Filtered. The corresponding public gains before adding the replay features are $+0.014$, $+0.094$, and $+0.071$.
GPQA-Diamond has the smallest prompt-visible gain and the widest setting-level scatter, consistent with its smaller sample ($n=198$; \Cref{tab:task_taxonomy}).
For comparison, the matched prompt-withheld full gains are much larger on GPQA-Diamond and MMLU-Pro ($+0.230$ and $+0.295$), and closer on Omni-MATH-2-Filtered ($+0.127$), where the open-response final answer already carries substantial correctness-predictive information.
Across setting means, the prompt-visible full gain ranges from $+0.004$ to $+0.143$ (\Cref{fig:main_ladder}a), while the prompt-withheld full gain ranges from $+0.075$ to $+0.332$ and remains positive even where one intermediate increment is flat or negative (\Cref{fig:cell_heatmap}).

Mixed-model sensitivity estimates remain positive, with intervals excluding zero for every increment except $q\Delta_S$ (\Cref{tab:hierarchical_deltas}; specification and fallback in the \hyperref[sec:methods]{Methods}). With only three benchmarks and two model families, these are sensitivity summaries for the observed design rather than population-level generalization intervals.
Mean AUROC is $0.812$ for $q+C$ and $0.809$ for $q+R+S+C$ (\Cref{tab:prompt_controls_15cell}), and several setting-level steps are negative (\Cref{tab:prompt_residual_contrasts}).
A nested component-block control, which concatenates separately computed feature blocks instead of re-embedding each expanded display, changes AUROC by $-0.004$ to $+0.016$ and preserves every access increment (\Cref{tab:component_blocks}).

Across all 15 settings, surface features, Cohere embeddings, and a fully open TF-IDF monitor reproduce the public access pattern, with full-public AUROC between $0.777$ and $0.815$ across the non-primary feature families (\Cref{fig:public_features,tab:public_feature_controls}).
After train-only linear residualization of every non-length surface and Cohere feature on word and sentence counts, across-item AUROC at the full-public display remains $0.717$ prompt-withheld and $0.713$ prompt-visible (from $0.814$ and $0.809$), while the corresponding public-ladder increments shrink to $+0.101$ and $-0.014$ (\Cref{tab:length_orthogonalized}). This analysis tests sensitivity to linear associations with length; causal attribution is outside its scope.
For the prompt-withheld contrast, the mean full-ladder gain remains between $+0.178$ and $+0.262$ under leave-one-family-out, leave-one-benchmark-out, and item-weighted aggregation (\Cref{tab:robustness_loo,tab:item_weighted_increments}). Item-clustered intervals give the same qualitative result (\Cref{tab:clustered_bootstrap}), and label permutation returns every level to chance (\Cref{tab:label_permutation}).
Along the output-only ladder, higher-access monitors rank better but are less well calibrated before validation-fitted Platt scaling (\Cref{tab:calibration}). The appendix also reports secondary metrics (\Cref{tab:secondary_metrics}), answer and parser baselines (\Cref{tab:baseline_controls}), selective risk (\Cref{tab:selective_risk}), failure strata (\Cref{tab:failure_stratification}), replay quality (\Cref{tab:replay_qc}), and feature-selection checks (\Cref{tab:feature_selection_audits,tab:sparse_feature_examples}).

\begin{figure}[t]
\centering
\includegraphics[width=\textwidth]{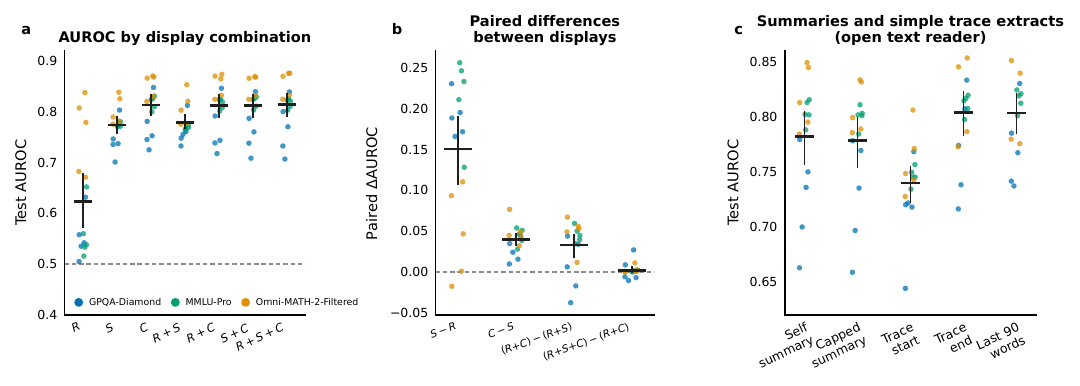}
\caption{\textbf{Across a mixed set of questions, short summaries carry most of the trace's ranking signal for the tested linear readers.} The prompt is withheld throughout. Black ticks show setting means with 95\% descriptive setting-bootstrap intervals over the 15 observed settings. \textbf{a}, Mean AUROC for each available combination of response $R$, self-summary $S$, and trace $C$. \textbf{b}, Paired differences between displays. \textbf{c}, Model-written summaries and simple trace extracts at matched word budgets under a reader based on surface and word-pattern features. Exact values are in \Cref{tab:public_lattice_15cell,tab:summary_extractive}; instruction and masking controls are in \Cref{tab:summary_ablation,tab:masked_controls}.}
\label{fig:public_signal}
\end{figure}

\section{Short summaries carry most of the trace's ranking signal for linear readers}
\label{sec:public_monitor_result}

In \Cref{fig:public_signal}, we compare summary- and trace-based displays under output-only access across a held-out mixture of questions, where monitors can use both item-level cues and evidence specific to the completed run.
With the prompt withheld, the trace-only monitor reaches mean AUROC $0.813$ across the 15 settings, while the summary-only monitor reaches $0.774$ and the full public display $R+S+C$ reaches $0.814$ (\Cref{fig:public_signal}a and \Cref{tab:public_lattice_15cell}).
We define the normalized excess-AUROC ratio as $r_{\mathrm{AUROC}}=(\mathrm{AUROC}(S)-0.5)/(\mathrm{AUROC}(C)-0.5)$.
Using the unrounded setting means, $r_{\mathrm{AUROC}}=0.874$ (95\% descriptive paired setting-bootstrap interval $[0.855,0.896]$ over the 15 observed settings; \Cref{tab:robustness_loo}). This normalized performance comparison is not an information-retention measure because AUROC is nonlinear and non-additive.
Adding the response to the trace changes little, with AUROC $0.812$ for $R+C$.
Paired contrasts place summary-containing displays $0.033$--$0.039$ AUROC below their trace counterparts, while adding the summary to $R+C$ changes AUROC by only $+0.002$ (\Cref{fig:public_signal}b).
The secondary Shapley allocation assigns mean contributions of $+0.154$ to $C$, $+0.118$ to $S$, and $+0.042$ to $R$ (\Cref{tab:public_shapley}). Because $S$ and $C$ overlap, these values allocate shared predictive performance rather than identify independent information sources.

The AUROC differences in \Cref{fig:public_signal}b also affect which runs a reader would retain. At 50\% coverage, retaining the half of test runs ranked most likely correct, the mean error rate is $0.243$ with $R$, $0.139$ with $R+S$, $0.114$ with $R+S+C$, and $0.098$ after adding $Z$ (\Cref{tab:selective_risk}).
Benchmark format explains part of the variation. GPQA-Diamond and MMLU-Pro responses are usually bare option letters, whereas the open-response Omni-MATH-2-Filtered baseline already reaches AUROC $0.755$ (\Cref{tab:secondary_metrics}). Response-only discrimination varies sharply by benchmark, model, and response format; neither selected-answer identity nor response length alone explains it (\Cref{tab:response_audit}).

In \Cref{fig:public_signal}c, we compare model-written summaries with deterministic trace extracts at the same word count.
Under the primary surface-plus-Cohere family, hard-capping the summary at 90 words or replacing it with a conclusion-region trace extract changes mean AUROC little. Summary-only AUROC is $0.774$ with the original summary, $0.775$ with the 90-word hard cap (Cap90), $0.780$ with the final-90-words extract (Last90), and $0.782$ with masked Last90; the corresponding $R+S$ values remain between $0.774$ and $0.780$, compared with $0.779$ originally (\Cref{tab:summary_primary_controls}).
In the open surface-plus-TF-IDF monitor, self-summary AUROC is $0.782$ and becomes $0.778$ after an exact 90-word cap.
Trace extracts matched to each capped summary's word count differ by position: first-word extracts are weaker ($0.740$), head--tail and random extracts are near the capped summary ($0.781$ and $0.777$), and last-word extracts are stronger, at $0.804$ versus $0.778$ for the capped summary. The paired difference on the unrounded setting means is $+0.025$ (95\% descriptive setting-bootstrap interval $[+0.013,+0.038]$ over the 15 observed settings; \Cref{tab:summary_extractive}).
Under the same open surface-plus-TF-IDF specification, the fixed Last90 extract is similar ($0.803$), and masking stated answers, option text, and prompt-derived numbers before extraction leaves it at $0.803$.
Within this open control, conclusion-region-biased extracts are especially predictive. When the requested extract is longer than the trace, it can include the whole trace.

In \Cref{tab:summary_extract_features}, we compare the primary surface features of each self-summary with its same-run Last90 extract.
Weighting the 15 settings equally, over the 80{,}763 retained pairs, Last90 contains higher rates of all three lexical marker families than the self-summary. The Last90-minus-summary differences are $+0.009$ for uncertainty, $+0.012$ for self-correction, and $+0.051$ for answer markers, and all three 95\% descriptive setting-bootstrap intervals exclude zero. The corresponding oriented single-feature AUROC differences are $+0.090$ ($[+0.064,+0.118]$), $+0.068$ ($[+0.048,+0.089]$), and $+0.072$ ($[+0.049,+0.095]$). Density results are mixed. Last90 has higher digit density but weaker digit discrimination ($+0.018$ in level and $-0.022$ in AUROC), whereas its math-character density is lower but more discriminative ($-0.010$ and $+0.044$). The two density measures move in opposite directions, so the data do not indicate that summarization consistently removes numerical or symbolic content.

A generic summary instruction changes mean AUROC for summary-containing displays by $-0.011$ to $+0.001$ (\Cref{tab:summary_ablation}).
Summary decoding also has little effect. In a balanced 48-item-per-setting sensitivity with the monitor held fixed, the sampled-decoding SD of the summary increment is $0.017$ and the greedy value sits $+0.010$ above the sampled mean (\Cref{fig:summary_sampling_sensitivity,tab:summary_sampling_sensitivity}; design in the \hyperref[sec:methods]{Methods}). Its absolute increment is not directly comparable with the full-data $+0.156$, but within this design the output-only gain is not attributable to one sampled summary draw.

To test whether whole-display features miss localized trace evidence, we divide each trace into up to eight balanced contiguous chunks and pool chunk-level features or scores (design in the \hyperref[sec:methods]{Methods}). The best pooled versions match the whole-display trace within $0.001$ AUROC in both feature families (\Cref{tab:chunked_trace,tab:chunked_trace_full_data_open}). These controls test whether the results depend on how the trace is divided and read; they do not locate the evidence causally, and a short trace can occupy a single chunk.
A masking control replaces answer spans, choice letters, option text, and prompt-derived numbers before recomputing all features.
The combined mask changes mean trace AUROC by $+0.001$, and no per-benchmark contrast moves by more than $0.012$ (\Cref{fig:hardening_controls}a and \Cref{tab:masked_controls}).
A companion control removes 8--14-word spans copied from the question and changes mean AUROC by between $-0.002$ and $+0.001$ across the six summary- and trace-containing displays (\Cref{fig:hardening_controls}b and \Cref{tab:noqcopy_controls}).
These controls restrict near-verbatim reconstruction but cannot remove paraphrase or a higher-level indication that a question is difficult.

\begin{figure}[t]
\centering
\includegraphics[width=\textwidth]{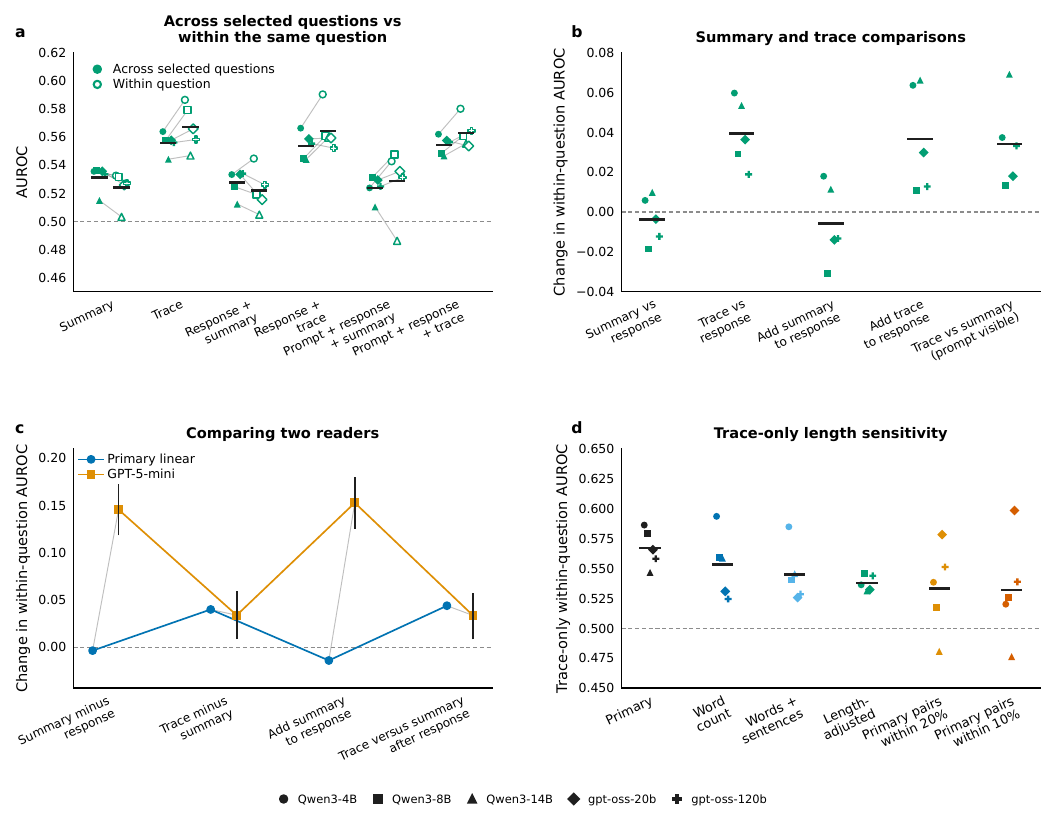}
\caption{\textbf{For repeated runs of the same question, linear summary readers are near chance, while reader choice and trace length matter.} All panels use the 939--1{,}389 outcome-discordant MMLU-Pro items per model that appeared in at least one held-out split. In panels \textbf{a}, \textbf{b}, and \textbf{d}, marker shapes identify the five target models, and black ticks show their means. Per-model intervals for the primary within-question display estimates are in \Cref{tab:multirun_within_item}. \textbf{a}, Filled points rank runs across the selected questions; open points compare correct and incorrect runs of the same question. Six summary- and trace-bearing displays are shown. \textbf{b}, Five direct within-question comparisons contrast summary and trace access; values above zero favor the display named first. \textbf{c}, For gpt-oss-20b, four selected comparisons from the ten-display analysis show how the primary linear reader and GPT-5-mini use summaries and traces differently. Black vertical lines give the paired 95\% item-bootstrap intervals for the GPT-5-mini comparisons. \textbf{d}, Trace-only within-question AUROC is shown for the primary reader, two readers using only text length, a length-adjusted reader, and pairs whose trace lengths differ by at most 20\% or 10\%. Exact values are in \Cref{tab:multirun_within_item,tab:multirun_prompt_visible,tab:llm_judge_factorial,tab:within-item-length-controls}.}
\label{fig:difficulty_signal}
\end{figure}

\section{For repeated runs of the same question, linear summary readers are near chance}
\label{sec:within_item_result}

The repeated-run control tests whether monitors separate correct from incorrect runs of the same item (\Cref{fig:difficulty_signal}). We generate two additional MMLU-Pro runs per item and model, fit monitors on the primary run, and score all three runs. Both reported AUROCs use the same 939--1{,}389 outcome-discordant items per model that appeared in at least one held-out split. Across-item AUROC pools runs across selected questions, whereas within-item AUROC compares a correct and incorrect run of the same question and model, holding item-level cues fixed (\Cref{fig:difficulty_signal}a and \Cref{tab:multirun_within_item}).
The analysis is limited to MMLU-Pro because GPQA-Diamond has too few multiply sampled items and Omni-MATH-2-Filtered has one run per item.

Under the primary linear reader, the summary displays are near chance within items, with within-item AUROC $0.503$--$0.545$ and across-item AUROC $0.512$--$0.536$ on the same discordant items (\Cref{fig:difficulty_signal}a).
The trace-containing displays retain only modest signal, with within-item AUROC $0.544$--$0.590$ and across-item AUROC $0.544$--$0.566$.
The response-only monitor is closer to chance, at $0.493$--$0.550$, with intervals covering $0.5$ in two settings.
When both measures use the same discordant items, the gap between across-item and within-item AUROC is small and changes direction across settings. We therefore base the conclusion on the absolute within-item estimates and the matched pairwise comparison. The prompt-conditioned result in \Cref{sec:ladder_result} measures a different quantity: across-item performance when the prompt is visible.

Visible text length accounts for much of the trace reader's repeated-run score (\Cref{fig:difficulty_signal}d and \Cref{tab:within-item-length-controls}). Averaged equally across models, trace word count alone reaches within-question AUROC $0.553$, compared with $0.567$ for the primary trace reader; using word and sentence counts gives $0.545$. After adjusting the other features for their fitted linear association with both counts, trace AUROC is $0.538$, compared with $0.508$ for the similarly adjusted summary reader. When correct--incorrect pairs are restricted to trace lengths within 20\% or 10\%, the primary trace reader reaches $0.533$ or $0.532$, remaining $+0.032$ or $+0.037$ above the summary reader. These restrictions retain only 3{,}412 of 10{,}608 pairs (32\%) and 1{,}701 pairs (16\%), so they are selective sensitivities rather than estimates for every discordant item.

Across models, 939--1{,}389 of the 1{,}364--2{,}062 outcome-discordant items (65.4--69.6\%) appear in at least one held-out split and enter the analysis (\Cref{tab:discordant_sensitivity}).
These selected questions are systematically harder than held-out questions whose three outcomes agree, with primary-run accuracy 23.3--29.9 percentage points lower and median traces 2.50--2.96 times longer (\Cref{tab:discordant-item-profile}). The repeated-run estimate therefore applies to this unusually difficult, unstable subset rather than to all MMLU-Pro questions.
In both strata, every prompt-withheld trace display remains above $S$ and $R{+}S$, at $0.604$--$0.611$ versus $0.548$--$0.551$ with exactly one incorrect run and $0.512$--$0.518$ versus $0.493$--$0.494$ with exactly two, and the prompt-visible trace-bearing displays show the same ordering. Most of the pooled trace discrimination is nonetheless concentrated in the one-incorrect stratum. The trace-only reader reaches $0.611$ $[0.596,0.626]$ when exactly one of three runs is incorrect but only $0.518$ $[0.502,0.535]$ when exactly two are, although the two strata contribute comparable numbers of items ($n{=}2{,}829$ versus $n{=}2{,}475$). When two of three runs are incorrect, even the trace-only reader is only slightly above chance, and the $R{+}S{+}C$ interval covers $0.5$ (\Cref{tab:discordant_sensitivity}). Model-balanced reweighting changes any pooled estimate by only 0.001 at the reported precision.
Conditional-logit coefficients for standardized monitor scores are positive for every trace-containing display and model ($\beta=0.359$--$0.815$, with all 95\% intervals above zero). Summary-only and response-plus-summary coefficients are smaller ($\beta=0.064$--$0.343$), with one interval crossing zero for each display (\Cref{tab:multirun_within_item}).

Across-item AUROC on the discordant subset ranks runs across selected questions, whereas within-item performance tests a particular run after fixing the question. The transferred across-item monitor is not an upper bound on run-specific information because it was trained across items.
We therefore train pairwise logistic monitors on correct-minus-incorrect feature differences within discordant training items and evaluate them on held-out discordant items (\Cref{tab:item_conditional_pairwise}).
In this direct control, the summary displays ($0.512$--$0.542$) remain close to the response-only range ($0.511$--$0.541$), while the trace-containing displays reach $0.534$--$0.580$.
In this MMLU-Pro comparison, summary displays remain weak at run-level diagnosis, while traces retain modest, model-dependent signal. We test only a pairwise logistic reader using surface features and Cohere embeddings, so richer item-conditional readers may perform differently.
Replacing the self-summary with the Last90 extract raises across-item AUROC for all five models and raises within-item AUROC by a mean of $+0.022$ under the primary reader (\Cref{tab:multirun_last90}), while the open version's within-item changes are mixed (\Cref{tab:multirun_last90_open}). Both Last90 controls use the whole trace when it contains fewer than 90 words and therefore do not isolate an ending-only effect.

The primary prompt-visible monitor jointly encodes $q$ in every display. On the same discordant items, summary displays remain near chance while trace displays retain modest within-item signal (\Cref{fig:difficulty_signal}a and \Cref{tab:multirun_prompt_visible}). Within item, adding $S$ to $q+R$ changes AUROC by mean $+0.026$, adding $C$ by mean $+0.060$, and adding $S$ after $q+R+C$ by mean $+0.003$ (\Cref{fig:difficulty_signal}b). The secondary open surface-plus-TF-IDF monitor is directionally similar with a similarly small summary increment (\Cref{tab:open_prompt_multirun}). The prompt is held fixed within a pair but remains part of the jointly fitted representation, so these display differences can include prompt--response or prompt--trace interactions and do not decompose the signal by source.
A direct primary item-conditional version, trained on correct-minus-incorrect run differences within discordant training items, gives the same asymmetry. Adding $C$ to $q+R$ is positive in every model ($+0.025$ to $+0.060$) and adding $S$ after $q+R+C$ ranges from $-0.020$ to $+0.011$ (\Cref{tab:item_conditional_pairwise_prompt}), and the secondary open version agrees (\Cref{tab:item_conditional_pairwise_open}).

A GPT-5-mini sensitivity with a fixed judge instruction evaluates all ten displays on the same 1{,}389 analyzed outcome-discordant gpt-oss-20b questions (\Cref{fig:difficulty_signal}c). With the prompt withheld, within-item AUROC is $0.503$ for the response, $0.649$ for the summary, $0.683$ for the trace, $0.656$ for response plus summary, and $0.690$ for response plus trace (\Cref{tab:llm_judge_factorial}).
Using the summary instead of the response changes AUROC by $+0.146$ (95\% CI $[+0.118,+0.172]$), and adding it to the response changes AUROC by $+0.153$ ($[+0.125,+0.180]$). The corresponding descriptive changes for the primary linear reader are $-0.004$ and $-0.014$. The trace still performs slightly better than the summary under both readers: the GPT-5-mini differences are $+0.034$ ($[+0.008,+0.059]$) without the response and $+0.034$ ($[+0.008,+0.057]$) with it, compared with $+0.040$ and $+0.044$ for the linear reader.
When GPT-5-mini also sees the prompt and response, its within-item AUROC is $0.800$. Adding the summary changes it by $-0.024$ ($[-0.040,-0.007]$), adding the trace by $-0.064$ ($[-0.084,-0.044]$), and adding the summary after the trace by $-0.015$ ($[-0.032,+0.002]$). Because GPT-5-mini can solve the visible task and compare its answer with the response, these comparisons mix monitoring with task solving and should not be read as evidence that the added text contains less information.
The near-chance summary result therefore applies to the tested linear readers rather than to every possible reader.
The prompt-visible result in \Cref{sec:ladder_result} provides the complementary across-item view.

\section{Discussion}
\label{sec:discussion}

Summary usefulness depends on both prompt visibility and the reader. With the prompt visible, the summary adds only $+0.019$ AUROC, compared with $+0.156$ when the prompt is withheld, while the trace still adds $+0.041$. On repeated runs of the same question with the prompt withheld, linear summary readers are near chance, whereas GPT-5-mini extracts substantially more signal. In the common case where users already hold the prompt, summaries are less helpful than the full trace for monitoring correctness. Monitorability claims, including claims about faithfulness, should therefore name both the displayed evidence and the reader.

Prompt visibility matters because the prompt already supplies much of the information in the summary. A prompt-only reader reaches AUROC $0.734$, partly because the question indicates how difficult the run is likely to be. The remaining useful evidence concerns the particular run, and more of it appears in the trace than in the summary. Reading the trace after the summary still helps, while reading the summary after the trace adds essentially nothing. Qualitative inspection points to self-corrections, contradictions, and incomplete computations as the kind of run-specific evidence involved, although these are not separated from length or correlated features.

Without the prompt, when only responses and summaries are retained, short summaries remain genuinely useful. They carry most of the trace's ranking signal across mixed questions and help rank runs by their likelihood of being correct. The trace's own ending does the same job at equal length and carries denser and more discriminative uncertainty and self-correction cues than the summary written from it, although no single feature family explains the readers' scores. Judging repeated runs of the same question is harder, because question-level cues cancel and only run-specific evidence helps. There, linear summary readers are near chance and trace readers keep only a modest edge, with or without the prompt. GPT-5-mini recovers substantial signal from summaries that the linear readers miss, yet the trace stays slightly ahead. When that same judge also sees the prompt, its scores mainly reflect its own ability to solve the task rather than what the displays reveal. This result shows why a monitorability estimate must name the reader.

Text length carries much of the signal the linear readers use. Adjusting the other features for word and sentence counts removes most of the across-question gains and turns the trace-minus-summary contrast slightly negative, from $+0.043$ to $-0.024$, while the within-item trace edge survives length-matched comparisons. The adjustment removes meaningful cues along with length, so the original and adjusted results should be read together, and neither identifies a length-independent semantic effect.

\paragraph{Limitations.}
There are four main limitations.
First, our summaries are a controlled proxy. The same model summarizes its own trace under one instruction, without the prompt, content filters, or any incentive to evade a monitor. The estimates therefore describe this summary method rather than deployed provider summaries~\cite{openai2024learningreasoning,openai2026reasoningdocs} or adversarial summaries. The results are stable under the tested instruction and decoding changes. A direct next test is to apply the paired comparison to commercial models that expose summaries.
Second, the claims are reader-relative. Most estimates use simple linear readers, the stronger-reader test covers one judge on one model and benchmark, and monitors fitted in one setting transfer imperfectly to others (\Cref{fig:transfer,tab:transfer_matrix}). We therefore state every result relative to the reader that produced it.
Third, the ladder measures final-answer correctness only. Nothing here shows that a trace or summary is a causal account of the model's computation~\cite{lanham2023faithfulness,chen2025reasoning}, and extending the design to faithfulness or misbehavior requires labels defined independently of the reader~\cite{wang2026monitorbench,openai2025detectingmisbehavior}.
Fourth, length is entangled with the linear readers' signal, and our adjustments show its influence but do not disentangle it.
Future work should also compare several summary lengths and readers, and report the monitored property, access, prompt visibility, and reader explicitly.

\clearpage
\section*{Acknowledgements}
This research was supported by funding from the Vrije
Universiteit Brussel Research Council (VUB-OZR) and the Flemish Government under the ``Onderzoeksprogramma Artifici\"ele Intelligentie (AI) Vlaanderen'' program.
Andres Algaba acknowledges support from the Francqui Foundation (Belgium) through a Francqui Start-Up Grant and a fellowship from the Research Foundation Flanders (FWO) under Grant No.1286924N. 
Vincent Ginis acknowledges support from Research Foundation Flanders under Grant No.G032822N and G0K9322N. 
The computational resources and services used in this work were provided by the VSC (Flemish Supercomputer Center), funded by the Research Foundation Flanders (FWO) and the Flemish Government - department WEWIS.

\section*{Author contributions}
A.A., B.V., and V.G.\ conceptualized the idea.
A.A.\ implemented the code and performed the analyses.
F.C., L.D., and M.B.\ verified the experiments and collaborated on the writing.
All authors read and approved the manuscript.

\section*{Data and code availability}
The code for this publication is publicly available at \url{https://github.com/AndresAlgaba/observability_ladder}. \\
Data associated with this study are available in a public repository at \url{https://doi.org/10.5281/zenodo.21770626}.

The GPQA-Diamond dataset~\cite{rein2023gpqa} is available at \url{https://huggingface.co/datasets/Idavidrein/gpqa}. \\
The MMLU-Pro dataset~\cite{wang2024mmlu} is available at \url{https://huggingface.co/datasets/TIGER-Lab/MMLU-Pro}. \\
The Omni-MATH-2-Filtered dataset~\cite{ballon2026benchmarks} is available at \url{https://huggingface.co/datasets/martheballon/Omni-MATH-2}.

The Qwen3 model family~\cite{yang2025qwen3} is available at \url{https://huggingface.co/collections/Qwen/qwen3}. \\
The gpt-oss model family~\cite{agarwal2025gpt} is available at \url{https://huggingface.co/collections/openai/gpt-oss}.

We used Python 3.13.1 with \textit{NumPy 2.4.3}, \textit{pandas 3.0.0}, \textit{pyarrow 23.0.0}, \textit{scikit-learn 1.7.2}, \textit{scipy 1.17.0}, \textit{statsmodels 0.14.6}, \textit{matplotlib 3.10.8}, \textit{vllm 0.12.0}, \textit{flashinfer-python 0.5.3}, \textit{PyTorch 2.11.0}, \textit{transformers 5.8.0}, and \textit{cohere 5.20.0}.

\clearpage
\bibliographystyle{unsrtnat}
\bibliography{references}

\clearpage
\appendix
\makeatletter
\@addtoreset{equation}{section}
\makeatother
\renewcommand{\theequation}{\arabic{equation}}
\renewcommand{\theHequation}{app.\thesection.\arabic{equation}}

\setcounter{figure}{0}
\renewcommand{\thefigure}{A\arabic{figure}}
\renewcommand{\theHfigure}{appendix.figure.\arabic{figure}}
\setcounter{table}{0}
\renewcommand{\thetable}{A\arabic{table}}
\renewcommand{\theHtable}{appendix.table.\arabic{table}}

\renewcommand{\tablename}{Appendix Table}
\renewcommand{\figurename}{Appendix Figure}
\crefalias{table}{appendixtable}
\crefalias{figure}{appendixfigure}

\section{Methods}
\label{sec:methods}

\paragraph{Data recorded for each run.}
For each benchmark item $i$ and target model $m$, we analyze the task prompt $q_i$, completed reasoning trace $C_{i,m}$, final response $R_{i,m}$, post-hoc self-summary $S_{i,m}$, provider-side replay features $Z_{i,m}$, extracted answer, correctness label, benchmark identifier, and quality-check information.
All results reported here use a single completed run per item, except the multi-run within-item control on MMLU-Pro, which adds two further independent runs per item.
For multiple-choice tasks, $R_{i,m}$ is the final response text visible after the reasoning trace, and the selected option letter is treated as the extracted answer.
For open-response mathematics, $R_{i,m}$ is the final response text, and the extracted answer span prefers a boxed answer, final expression, or concise answer.
The target is whether the final answer is correct. This binary run-level label is defined independently of the monitor, and we record which label definition was used.
We treat each model--benchmark setting as the unit of inference.

\paragraph{Models and benchmarks.}
We evaluate five open-weight target models from two families, Qwen3-4B, Qwen3-8B, and Qwen3-14B from the Qwen3 family, and gpt-oss-20b and gpt-oss-120b from the gpt-oss family~\cite{qwen2025qwen3collection, yang2025qwen3, openai2025gptoss, agarwal2025gpt}.
All target models are run in settings that permit local extraction of final hidden states and token log-probabilities.
We therefore do not evaluate closed API models.
Loading gpt-oss models uses the model-specific Transformers implementation, bfloat16 replay and summarization, and the Hugging Face kernel components required for MXFP4 inference.
The model weights are loaded from local snapshots of the Hugging Face repositories used for the run. The release documentation records the model sources, hashes of the weights and generation settings, and numeric precision.
All generation, summarization, and replay used approximately 750 GPU-hours on NVIDIA H200 GPUs.
GPQA-Diamond contains 198 four-way graduate-level science questions~\cite{rein2023gpqa}.
MMLU-Pro contains 12{,}032 multiple-choice questions with up to ten answer choices~\cite{wang2024mmlu}.
Omni-MATH-2-Filtered contains 4{,}181 open-response olympiad mathematics problems graded for response equivalence by GPT-5-mini following the benchmark procedure~\cite{ballon2026benchmarks}.
Correctness is determined by option match for GPQA-Diamond and MMLU-Pro.
For Omni-MATH-2-Filtered, correctness is determined by GPT-5-mini equivalence judgments between the model response and reference answer following Ballon et al.~\cite{ballon2026benchmarks}.
The judge is shown the problem, the model response, and the reference answer and decides whether the response is mathematically equivalent to the reference.
These judgments cover all 4{,}181 items for each reported model and run. We identify each judgment by its benchmark, model, run, and item.
We use the stored GPT-5-mini judgments for the exact responses analyzed here as the Omni-MATH-2-Filtered labels.
Against the deterministic exact/numeric checker, the judge agrees on 14{,}146/20{,}310 (69.7\%) resolved comparisons. Agreement is 7{,}859/7{,}865 (99.9\%) on deterministic exact/numeric matches, while the judge accepts 6{,}158/12{,}445 (49.5\%) deterministic mismatches as semantically equivalent. All 20{,}905 judge calls return a binary decision: 14{,}206 equivalent and 6{,}699 not equivalent, with no uncertain or error cases (\Cref{tab:judge_audit}). The release documentation includes per-model counts and details about each batch.

\paragraph{Separating reasoning traces and final responses.}
For each item, target model, and benchmark, we begin with one completed reasoning run and separate the raw completion into a reasoning trace, final response text, extracted answer, and parser status, which records whether this separation succeeded.
Target reasoning runs are generated with vLLM from benchmark system and user messages rendered through the target tokenizer chat template, with \texttt{add\_generation\_prompt=True} and \texttt{enable\_thinking=True}.
The vLLM \texttt{SamplingParams} set \texttt{max\_tokens=28000} for Qwen models and \texttt{max\_tokens=126000} for gpt-oss models, while leaving the other sampling settings at the vLLM defaults: temperature 1.0, top-$p$ 1.0, top-$k$ 0, and no random seed set.
We supply no explicit stop sequence during trace generation; generation ends when the model emits its end-of-sequence token or reaches the stated token limit, using the model's chat template. All reported analyses use the stored completions released with the paper.
Some Qwen3 completions emitted an empty thinking block, \texttt{<think></think>}, followed by content in the final response channel.
The parser assigns these runs $C=\emptyset$, and they remain in the analysis as observed model outputs.
In the primary run, empty traces occur for Qwen3-8B on GPQA-Diamond (3/198), MMLU-Pro (79/12{,}032), and Omni-MATH-2-Filtered (980/4{,}181), and for Qwen3-14B on MMLU-Pro (5/12{,}032) and Omni-MATH-2-Filtered (223/4{,}181); they do not occur for Qwen3-4B or the gpt-oss models (\Cref{tab:qc_counts}).
When $C$ is empty, the self-summary is still generated from that empty trace, so the summary can expose the absence of trace content.
Some Qwen3 completions, concentrated on Omni-MATH-2-Filtered, also omit the closing \texttt{</think>} tag.
For these runs, the parser leaves the final response $R$ empty and assigns the remaining completion text to $C$.
In the primary Omni-MATH-2-Filtered run, missing closing tags occur for Qwen3-4B (166/4{,}181), Qwen3-8B (170/4{,}181), and Qwen3-14B (142/4{,}181) (\Cref{tab:qc_counts}).
We retain this parser information for quality checks, and the runs remain in the main analysis.
For MMLU-Pro, we record the available choices explicitly because their number varies.
For Omni-MATH-2-Filtered, we mark final answers missed by the parser so extraction failures can be distinguished from incorrect model answers.
The main $Z_{\mathrm{prob}}$ and $Z_{\mathrm{hidden}}$ features are extracted from a separate prompt-conditioned replay of the completed run rather than from generation-time probability traces.
Our quality checks focus on parser and replay status.
We store further generation details to document how the traces were produced.

\paragraph{Self-summary generation.}
The primary self-summary is generated under a prompt requesting at most 90 words and is denoted $S$ throughout. Although the stored file name records a 256-token decoding limit, we define the summary by the 90-word instruction.
For each completed run, the same target model that generated the reasoning trace also generates the summary:
\[
S_{i,m}=\mathrm{Summary}_m(C_{i,m};\gamma_m),
\]
where $\gamma_m$ denotes the summary-generation settings.
The 90-word target is an instruction rather than a post-hoc length cap.
Summaries are decoded with a 256-token limit and are not truncated after generation.
Across the primary summaries ($n=82{,}055$), the mean length is 77.0 words, 74.4\% of summaries contain at most 90 words, and the maximum length is 231 words (\Cref{tab:qc_counts}).
The summary decoder uses Hugging Face \texttt{generate} with \texttt{do\_sample=True}, \texttt{use\_cache=False}, \texttt{pad\_token\_id} equal to the EOS token, and \texttt{max\_new\_tokens} set to the 256-token budget minus any final-channel prefill tokens.
No temperature or top-$p$ override is supplied in the primary run, so those settings follow the loaded model generation settings, falling back to Hugging Face defaults when unset (temperature 1.0, top-$p$ 1.0, top-$k$ 50).
Self-summary generation produces only the summary text.
Qwen summaries use \texttt{enable\_thinking=False}, while gpt-oss summaries use \texttt{reasoning\_effort="low"} and an assistant final-channel prefill so decoding begins directly in the final response channel rather than in an analysis/reasoning channel.
The summarizer receives only $C_{i,m}$, not the task prompt, final response, reference answer, correctness label, hidden states, logits, or token probabilities.
Because the summarizer receives only $C$, this setup measures how much of the trace it can compress without seeing the task prompt or final response. Supplying $q$ or $R$ could change the summary, so these results apply only to trace-conditioned summarization.
The primary prompt asks the model to compress the reasoning trace into at most 90 words while preserving uncertainty, contradictions, candidate answers, and self-corrections without adding evidence or repairing errors.
The prompt set also includes a generic version that drops the preservation clause and requests only a summary of at most 90 words. Uncertainty-only and final-answer-only versions were defined but are not analyzed here.
To test sensitivity to the instruction, we regenerate all summaries with the generic version under the same decoding settings and retrain the affected probes (\Cref{tab:summary_ablation}).
A worked appendix example shows the formatted target prompt, shortened completed response, and generated self-summary for one GPQA-Diamond run (\Cref{fig:example_qwen_prompt,fig:example_qwen_completed,fig:example_qwen_summary}).
For reproducibility, we record each summary's target model, prompt version, hash of the generation settings, sampling settings and overrides, seed identifier, input length, output length, and requested summary length. For the primary summaries analyzed here, the seed identifier was recorded but the random-number generator was not fixed.

\begin{figure}[t]
\centering
\includegraphics[width=\textwidth]{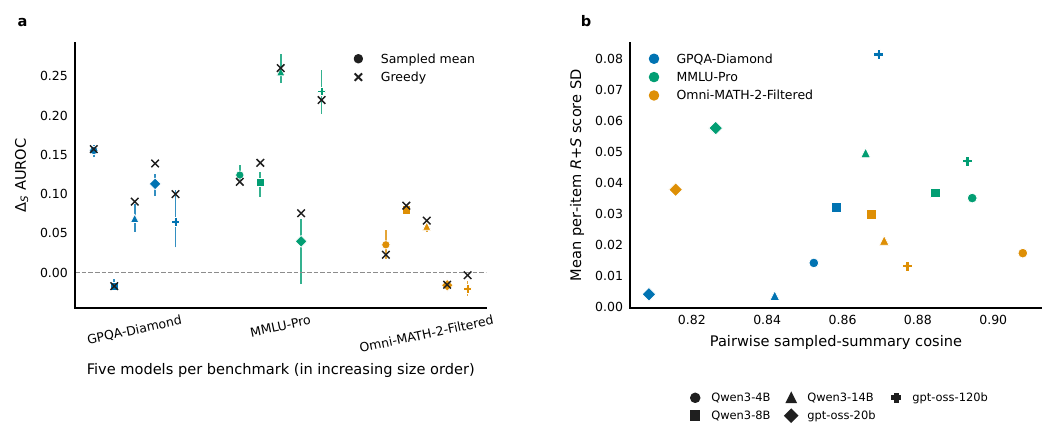}
\caption{\textbf{A fixed primary monitor is stable across alternate sampled summaries.} Each point uses a fixed surface-plus-Cohere monitor evaluated on a balanced 48-item subset in one of 15 target-model--benchmark settings; colors denote benchmarks and marker shapes denote target models. \textbf{a}, Colored symbols show the mean equal-stratum summary increment, $\Delta_S=\mathrm{AUROC}(R+S)-\mathrm{AUROC}(R)$, across three sampled summaries, vertical lines span their range, and black crosses show greedy decoding. Across settings, the sampled mean is $+0.086$ AUROC (hierarchical 95\% bootstrap interval $[+0.027,+0.146]$), and the greedy value is $+0.010$ higher on average. \textbf{b}, For each setting, mean pairwise cosine similarity among the three sampled-summary embeddings is plotted against the mean per-item standard deviation of the fixed $R+S$ monitor score across those summaries.}
\label{fig:summary_sampling_sensitivity}
\end{figure}

To measure summary-decoding sensitivity, we select 48 primary-run items per model--benchmark setting: eight for each combination of correctness and short, medium, or long traces, with the three trace-length groups defined by terciles, using a recorded random seed for item selection. For every selected trace, the target model generates three independently sampled summaries and one greedy summary. For each of $R$, $S$, and $R+S$, a primary surface-plus-Cohere monitor selects regularization on the first split's training and validation items after removing the selected items, is refitted on all remaining items, and is then held fixed across the four summary decodings. We give the six groups equal weight in the primary AUROC for this balanced subset and use inverse-selection-probability weighting as a sensitivity check. Hierarchical intervals use 2{,}000 draws that resample items within the six groups and resample the 15 observed settings (\Cref{fig:summary_sampling_sensitivity,tab:summary_sampling_sensitivity}).

\paragraph{Constructing the text shown to readers.}
We construct every visible display from the same completed run. Each display has a fixed identifier and set of visible components and is classified as either output-only or prompt-conditioned.
With the prompt visible, we evaluate every nonempty combination of $\{R,S,C\}$: $q+R$, $q+S$, $q+C$, $q+R+S$, $q+R+C$, $q+S+C$, and $q+R+S+C$. Both the primary surface-plus-Cohere and secondary open surface-plus-TF-IDF estimates cover all combinations. The main ladder follows the ordered path $q+R\rightarrow q+R+S\rightarrow q+R+S+C$, with $q$ and $q+C$ as controls.
Without the prompt, we evaluate the seven nonempty subsets of $\{R,S,C\}$, of which $R$, $R+S$, and $R+S+C$ form the prompt-withheld contrast.
Displays containing several components label each section as ``Task prompt:'', ``Response:'', ``Summary:'', or ``Reasoning trace:'' and separate them with blank lines. Single-component displays contain only the component text.
The physical section order is fixed as prompt, response, summary, and trace. Accordingly, \textit{after} in a factorial contrast and inclusion order in a Shapley calculation denote nested component sets rather than experimental permutations of text position.
Prompt-overlap-stripped (\textit{noqcopy}) versions are produced for the summary and trace of output-only displays.
All word $n$-grams of length 8--14 from the task prompt are matched case-insensitively in the visible text, matched spans are removed longest-first, overlapping removals are merged, and each removed region is replaced by a fixed ``[prompt-overlap removed]'' marker before feature extraction.

\paragraph{Display masking.}
Masked display versions replace task- and answer-revealing content with fixed markers before feature extraction.
Answer masking replaces the contents of \texttt{\textbackslash boxed\{\}} expressions and literal occurrences of the extracted answer (with a context-restricted rule for single-character answers) by an ``[ANSWER]'' marker, and masks choice letters in answer-stating contexts, including declarations such as ``the answer is X'', verb-adjacent letters, letter-to-option mapping restatements, and standalone letter lines.
Option masking replaces case-insensitive occurrences of option texts (taken from the benchmark options where available and otherwise parsed from option blocks in the prompt) by an ``[OPTION]'' marker.
Number masking replaces numeric literals of at least two characters that occur in the task prompt by a ``[NUMBER]'' marker.
Component versions apply one rule at a time to the trace display, and the combined \textit{maskall} version applies all rules to the trace, summary, and full public display, masking every visible component.
Spot-checks of sampled masked traces find no residual option-text or answer-context matches on MMLU-Pro, residual answer-stating phrasings in roughly $1\%$ of traces on GPQA-Diamond, and in roughly $18\%$ on Omni-MATH-2-Filtered, where short numeric answers escape the context patterns.
Masking removes verbatim and near-verbatim statements but cannot remove paraphrase. The masked-display results therefore show what remains after restricting direct reconstruction rather than eliminating it.
We recompute the surface features and embeddings from the masked text and train the probes in the same way on the same splits.

\paragraph{Public-text features.}
For each output-only display $P$, public monitors may compute features of the visible text $P$ but may not use the task prompt, target-model hidden states, logits, or token probabilities.
Prompt-conditioned control displays relax only the task-prompt exclusion.
The primary surface feature family includes word count, sentence count, average word length, digit density, math-character density, and per-word rates of uncertainty, self-correction, and answer-marker terms (\Cref{tab:surface_features}).
A no-length version removes word and sentence counts while retaining average word length, densities, and marker rates.
Raw marker counts, character counts, line counts, log-count transforms, and lexical-diversity features are retained only as appendix diagnostics because they are strongly length- or formatting-dependent in long reasoning traces.
The TF-IDF family combines word $n$-grams $(1,2)$ and character $n$-grams $(3,5)$, with train-only vocabulary fitting and a maximum of 5{,}000 word and 5{,}000 character features~\cite{joulin2017bag}.
The Cohere family uses 512-dimensional Cohere embed-v4.0 embeddings configured for classification and no silent truncation~\cite{reimers2019sentencebert, tunstall2022efficient, kusupati2022matryoshka, cohere2026embed, aws2026cohereembed}, an encoder chosen for its long input context window so that a full reasoning trace fits as a single whole context.
The primary combined public-text family concatenates rate-based surface features and Cohere embeddings after train-only preprocessing.
Cohere-based features define the primary monitor for the main ladder.
Across the same splits and all 15 settings, the open TF-IDF reader reproduces the mean access ordering and full-public performance (\Cref{tab:public_feature_controls}). It provides a reproducible alternative that does not depend on a proprietary encoder.
Fixed embeddings may be computed before splitting, but learned scaling, projection, or feature selection is fit only on the training split.

\paragraph{Direct comparison of summary and Last90 features.}
We match the stored surface features of each original self-summary to those of the fixed Last90 extract from the same benchmark item, model, and run, and use the paper's primary correctness label for that run. Last90 contains up to the final 90 whitespace-delimited trace words and contains the whole trace when it is shorter. We exclude pairs when either display has zero alphabetic words under the primary feature definition. This removes 1{,}292 zero-word Last90 displays and no self-summaries, leaving 80{,}763 pairs. Within each of the 15 model--benchmark settings, we compute the mean value of each feature for summaries and Last90, their paired difference, and rank-based single-feature AUROC without fitting a probe. For the seven features included in the sparse-model check, we orient AUROC using the dominant coefficient direction in \Cref{tab:sparse_feature_examples}. We treat larger average word length as positive by convention and retain the raw AUROC in the released results. The reported estimates give each setting equal weight. We form 95\% descriptive intervals with 10{,}000 bootstrap resamples of the 15 observed settings. These univariate statistics describe observable features of the summaries and Last90 extracts.

\paragraph{Length controls.}
We fit two direct baselines that use only the visible display's word count or its word and sentence counts, with the same splits, regularization search, final refit, and repeated-run scoring as the primary reader.
For the length-adjusted reader, we fit a linear projection from word and sentence counts to every other numeric surface and Cohere feature using the current training data, replace those features by their residuals, and then drop the two length variables before fitting the logistic model.
The projection is therefore learned on the training split during validation search and on train plus validation for the final model; held-out and repeated runs are transformed without refitting.
As a separate pair-matching sensitivity, we retain correct--incorrect runs of the same item only when their trace word counts differ by at most 10\% or 20\% of the longer trace, then recompute primary-reader within-item AUROC and item-bootstrap intervals.
These controls distinguish directly usable length information from variation not captured by the fitted linear association; they do not isolate a causal or length-independent effect.

\paragraph{Embeddings and long inputs.}
For each public access level, the primary embedding feature is a single embedding of exactly the visible public text at that level, $e(R)$, $e(R\Vert S)$, or $e(R\Vert S\Vert C)$.
The embedded strings are the plain ladder-display strings, including section headers such as ``Response'' and ``Summary'' where present, not chat-template text or tokenizer special tokens.
Silent truncation is disallowed.
If a display exceeds the encoder's context window, we divide it into fixed chunks and record the number and size of the chunks and how they are combined.
In the reported analysis, every full-public $R+S+C$ text fits within the encoder's context window.
\Cref{tab:embedding_qc} reports whole-context fit and chunk counts.
Because the primary display embedding replaces $e(R)$ by $e(R\Vert S)$ and then by $e(R\Vert S\Vert C)$, higher levels are not feature-nested in the embedding block.
We therefore define a nested component-block control using $[\phi(R)]$, $[\phi(R),\phi(S)]$, $[\phi(R),\phi(S),\phi(C)]$ and the analogous prompt-conditioned blocks, retaining lower-level component features unchanged when access expands.

\paragraph{Provider-side replay features.}
The provider-side replay component is
\[
Z = Z_{\mathrm{prob}}\oplus Z_{\mathrm{hidden}}.
\]
For each completed run, we replay the run to its final-response position as
\[
x_{i,m}^{\mathrm{dec}} =
\operatorname{ChatTemplate}_m(q_i)\Vert C_{i,m}\Vert s_{\mathrm{stop}},
\]
and extract
\[
h_{i,m}^{\mathrm{dec}} =
H_m(x_{i,m}^{\mathrm{dec}})_{\mathrm{last}}.
\]
The replay suffixes are
\[
s_{\mathrm{stop}}^{\mathrm{Qwen}}=\texttt{\textbackslash n</think>\textbackslash n\textbackslash n},
\qquad
\begin{aligned}
s_{\mathrm{stop}}^{\mathrm{gpt\text{-}oss}}={}&\texttt{<|end|><|start|>assistant}\\
&\texttt{<|channel|>final<|message|>}.
\end{aligned}
\]
For GPQA-Diamond, $Z_{\mathrm{prob}}$ includes $\log p(A),\ldots,\log p(D)$, entropy, margin, and maximum probability.
For MMLU-Pro, answer-choice log-probability features use the fixed answer labels $A$--$J$, with unavailable choices masked.
Choice-count and missing-choice indicators are included so probes do not interpret padding as probability mass.
Choice probabilities use the tokenizer encoding of bare uppercase labels $A$--$J$ with \texttt{add\_special\_tokens=False}. Each label is required to be a single token, and probabilities are normalized over the labels available for that item.
Thus, for multiple-choice tasks, $Z_{\mathrm{prob}}$ scores answer labels at the next-token decision point after the prompt and completed trace context $q+C$.
We also use choice-invariant summaries such as selected-choice probability, maximum probability, second-highest probability, entropy, margin, and probability variance.
For Omni-MATH-2-Filtered, $Z_{\mathrm{prob}}$ scores the model's own extracted or generated final answer span, not the reference answer.
The current open-response probability feature set scores that span with \texttt{add\_special\_tokens=False} after the same replay context and includes mean, minimum, quantile, and sum log-probabilities, answer-token and answer-character counts, and first-token entropy, margin, and maximum-probability summaries.
The count and sum features are length-sensitive and partly encode information already visible in $R$.
$Z_{\mathrm{hidden}}$ contains the raw final hidden state and scalar hidden-state summaries such as norm, mean, standard deviation, minimum, and maximum. On average in the seven-setting ablation subset, the hidden-state gain is larger with the run's own trace than with the prompt alone and is lost under shuffled-trace replay. Across this subset, $q+C$ replay differs from $C$-only replay by only $+0.006$ AUROC on average, so the prompt adds little beyond trace-conditioned replay in this ablation (\Cref{tab:z_replay_ablation}).

\paragraph{Quality checks.}
Parser status and replay status are tracked separately from the provider-side replay level.
We use this information for quality and sensitivity checks rather than as part of the main $Z$ feature set because parser and replay failures could otherwise look like replay features.
In the reported primary settings, every replay fits within the context window and produces a usable hidden-state vector (\Cref{tab:replay_qc}). If a replay is missing or too long in a diagnostic run, probability features are marked as missing, and we retain hidden-state placeholders and replay-status indicators for quality checks. We impute missing values only during probe fitting and using training data.
Completion length is visible when $C$ is visible, so direct length-derived features are treated as public text features or removed in no-length controls.
To test whether formatting failures affect AUROC, we repeat the analysis within each model--benchmark setting, access level, and split after applying two restrictions. The first excludes blank traces, and the stricter restriction also requires parser status ``ok'' and nonempty responses and extracted answers. The five per-split AUROCs are averaged within each setting before the 15 setting means receive equal weight (\Cref{tab:censoring_sensitivity}). Because these restrictions discard visible formatting failures, they test how much those failures affect the result. They do not estimate performance for hypothetical error-free outputs.

\paragraph{Probe training and splits.}
We fit $\ell_2$-regularized logistic probes for final-answer correctness~\cite{alain2016understanding, conneau2018cram, hewitt2019designing, belinkov2022probing}.
All learned preprocessing is train-only, including feature standardization, TF-IDF vocabulary fitting, and embedding-feature scaling.
The feature blocks included in each probe, and the quantities excluded by design, are summarized in \Cref{tab:probe_specification}.
For each model--benchmark setting, access level $k$, feature family $f$, and split $s$, the monitor first converts the visible display $\mathcal{O}_k(i)$ into a feature vector.
Written out by feature block, the logistic predictor is
\begin{align}
\eta_{ikfs}
=\;& \alpha_{kfs}
+ \underbrace{\sum_{r=1}^{J_{\mathrm{surf}}}
\beta^{\mathrm{surf}}_{rkfs} x^{\mathrm{surf}}_{ir}}_{\substack{\text{surface text} \\ \text{features}}}
+ \underbrace{\sum_{j=1}^{J_{\mathrm{emb}}}
\beta^{\mathrm{emb}}_{jkfs} e^{\mathrm{emb}}_{ij}}_{\substack{\text{public-text} \\ \text{embedding elements}}} \nonumber \\
&+ \underbrace{\mathbb{1}\{Z_{\mathrm{prob}}\in f\}
\sum_{a=1}^{J_{\mathrm{prob}}}
\beta^{\mathrm{prob}}_{akfs} z^{\mathrm{prob}}_{ia}}_{\substack{\text{answer-probability} \\ \text{features}}}
+ \underbrace{\mathbb{1}\{Z_{\mathrm{hidden}}\in f\}
\sum_{h=1}^{J_{\mathrm{hidden}}}
\beta^{\mathrm{hidden}}_{hkfs} z^{\mathrm{hidden}}_{ih}}_{\substack{\text{hidden-state} \\ \text{features}}},
\label{eq:probe_linear_predictor}
\end{align}
and
\[
\Pr(y_i=1\mid \mathcal{O}_k(i), f, s)
=
\sigma(\eta_{ikfs}),
\]
where $\sigma(u)=(1+\exp(-u))^{-1}$.
Here $x^{\mathrm{surf}}_{ir}$ is surface feature $r$, $e^{\mathrm{emb}}_{ij}$ is embedding element $j$, $z^{\mathrm{prob}}_{ia}$ is answer-probability feature $a$, and $z^{\mathrm{hidden}}_{ih}$ is hidden-state feature $h$.
The indicator terms state that answer-probability and hidden-state sums are included only for feature families that contain those provider-side replay blocks.
Public-only probes set those terms to zero.
Equivalently, the model can be written compactly as $\Pr(y_i=1)=\sigma(\alpha_{kfs}+\bm{\beta}_{kfs}^{\top}\bm{z}_{ikfs})$, where $\bm{z}_{ikfs}=T_{fs}(\mathcal{O}_k(i))$ is the feature vector produced by the train-fitted preprocessing map $T_{fs}$.
For a candidate inverse regularization strength $C$, the fitted coefficients solve
\[
(\hat{\alpha}_{kfs,C},\hat{\bm{\beta}}_{kfs,C})
=
\arg\min_{\alpha,\bm{\beta}}
\left[
\sum_{i\in \mathcal{D}^{\mathrm{train}}_s}
\left\{
-y_i\log p_i-(1-y_i)\log(1-p_i)
\right\}
+\frac{1}{2C}\lVert \bm{\beta}\rVert_2^2
\right],
\]
with $p_i=\sigma(\alpha+\bm{\beta}^{\top}\bm{z}_{ikfs})$.
This discriminative probe estimates held-out monitorability at a fixed access level; we do not interpret its coefficients as causal effects.
The inverse regularization strength is selected on the validation split from a prespecified set of values.
Primary across-item probes use $C\in\{10^{-4},10^{-3},10^{-2},10^{-1},1,10\}$.
Implementation uses \texttt{sklearn.linear\_model.LogisticRegression} with an $\ell_2$ penalty, \texttt{solver="liblinear"}, \texttt{max\_iter=2000}, \texttt{class\_weight=None}, median imputation and standard scaling for numeric features. We use a fixed random state for each split.
The selected $C$ is chosen by validation AUROC, and the final model is refit on train plus validation before scoring the held-out test split.
Final metrics are reported only on the held-out test split.
We use five random splits with 60/20/20 train/validation/test proportions, stratified by final-answer correctness within each model--benchmark setting.
Splits are created separately within each model--benchmark setting and reused across access levels and feature families, making ladder deltas paired.
For GPQA-Diamond, each split must contain both correctness classes in train, validation, and test.
Sparse models are used for interpretability checks, not primary estimation.
We exclude language-model judges from the main public-text analysis because they can solve the task themselves, whereas the primary analysis evaluates a fixed class of linear readers.
The fixed-instruction GPT-5-mini analysis below is reported only as a sensitivity to reader class, not as an additional ladder level.
For each of the 15 settings, a fixed selection rule retains one result file together with its paired predictions.
We use a result only when the same file contains its paired predictions; if several files provide the same access level, we use the primary result file.
Separate designated result files provide the subset, prompt-control, stripped-display, and decomposition analyses.

\paragraph{Sparse-model interpretation checks.}
We fit two sparse logistic models to ask which feature sources are repeatedly used by the public and final-level probes.
Plain LASSO uses an $\ell_1$ penalty over standardized individual features and then summarizes selected features by source group.
Sparse group LASSO combines a high-level source-group penalty with within-group sparsity, using source groups for surface text, Cohere embeddings, and answer-probability features when available.
These checks use the same target labels, access levels, train-only preprocessing, and split assignments as the main probes, but they are not used as the primary AUROC estimates.
Raw hidden-state dimensions are excluded from the sparse selection tables for tractability and interpretability.
Their contribution is instead estimated by the replay-feature sensitivity analysis.
Constant feature columns are dropped before sparse fitting.

\paragraph{Repeated runs of the same question.}
For MMLU-Pro, the target models produce two further independent runs for each of the 12{,}032 items under the generation defaults of \Cref{tab:generation_defaults}; all three runs are extracted and graded exactly as in the primary analysis.
After grading, we identify discordant items as those with at least one correct and one incorrect run. The complete three-run grades contain 1{,}364--2{,}062 such items per model among all 12{,}032 questions.
For runs 2--3, we then compute and store self-summaries, displays, text features, replay features, and monitor predictions only for this outcome-selected set of items; the original runs and grades remain available for all items.
The within-item analysis uses the subset of discordant items held out in at least one split, namely 939--1{,}389 items per model on MMLU-Pro, or 65.4--69.6\% of all outcome-discordant items.
Monitors are trained on the primary run's train and validation splits with the same split assignments, feature preprocessing, and validation-selected regularization as the primary public probes, so the repeated-run scores come from the same monitors as the main results.
For the reported repeated-run quantities, each split contributes held-out items with stored predictions for runs 1--3 and discordant outcomes. We pool item--run scores across the five splits by averaging over the splits in which the item was held out.
Reported across-item AUROC pools item--run observations only over the same post-outcome discordant items used by the within-item analysis. We retain an all-run value with mixed availability only as an internal check because summaries, features, and predictions for runs 2--3 were computed only for the discordant items.
The within-item AUROC is the pairwise concordance between correct and incorrect runs of the same item, with ties counted as one half, computed over discordant items.
Uncertainty comes from a 5{,}000-draw bootstrap that resamples per-item concordant-pair and total-pair counts at the item level, and a conditional logistic regression with item fixed effects on standardized monitor scores provides a model-based robustness check.
The primary repeated-run analysis includes the response-only, summary-only, trace-only, response-plus-summary, response-plus-trace, and full-public response-plus-summary-plus-trace displays, together with the prompt-visible set $\{q+R,q+R+S,q+R+C,q+R+S+C\}$.
It does not evaluate the provider-side replay features within item. GPQA-Diamond is excluded because too few items have multiple independent runs, and Omni-MATH-2-Filtered because only a single completed run per item is available.

A secondary open repeated-run analysis repeats the prompt-visible display set with rate-based surface features and joint word- and character-level TF-IDF. Both primary and open monitors are trained on the primary run's train and validation partitions, then score runs 1--3 under the same split pooling, post-outcome discordant-item restriction, within-item concordance estimator, item bootstrap, and conditional-logit check described above. Because the prompt is present in every jointly encoded display, it is held fixed by within-item pairing but can interact representationally with the response, summary, and trace.

The direct primary item-conditional analysis uses the same stored MMLU-Pro repeated runs and both the output-only display set $\{R,S,R+S,C,R+C,R+S+C\}$ and the prompt-visible set $\{q+R,q+R+S,q+R+C,q+R+S+C\}$.
For each model, display, and split, it restricts training, validation, and test partitions to discordant items.
Within each discordant item, every correct run is paired with every incorrect run, and the training examples are the feature differences correct-minus-incorrect and incorrect-minus-correct, labeled 1 and 0 respectively.
The monitor is an $\ell_2$-regularized logistic classifier with median imputation and standard scaling.
Regularization is selected on validation within-item pairwise AUROC from $C\in\{10^{-4},10^{-3},10^{-2},10^{-1}\}$, then the model is refit on train plus validation pairs and evaluated on held-out discordant items.
A secondary open prompt-visible version uses the display set $\{q+R,q+R+S,q+R+C,q+R+S+C\}$ and the surface-plus-TF-IDF representation, with the same within-item pair construction and held-out evaluation. It selects inverse regularization strength from $\{10^{-4},10^{-3},10^{-2},10^{-1},1,10\}$, matching the candidate values used by the open across-item probes. The reported 2.5th--97.5th percentiles of the five per-split estimates describe split sensitivity.
These analyses are CPU-only and use previously computed public features; they do not generate new runs, summaries, embeddings, replay features, or LLM judgments.

\paragraph{Comparison with a stronger reader.}
To test whether the result depends on the reader, we query \texttt{gpt-5-mini-2025-08-07} on the ten previously constructed display types for the 1{,}389 analyzed outcome-discordant gpt-oss-20b MMLU-Pro questions and three runs per question. The output-only displays are $\{R,S,C,R+S,R+C,R+S+C\}$, and the prompt-visible displays are $\{q+R,q+R+S,q+R+C,q+R+S+C\}$.
GPT-5-mini is the same judge model used for Omni-MATH-2-Filtered grading~\cite{ballon2026benchmarks}.
The three previously scored displays, $S$, $R+S$, and $q+R+S$, are reused without new API calls. We score the seven remaining displays with the same versioned instruction, \texttt{mmlu\_pro\_display\_correctness\_probability\_v1}, which requests one bare probability that the displayed run is correct. Requests use medium reasoning effort and no sampling-parameter override, and the judge receives neither the gold answer, reference answer, nor correctness label. We store each request and response with its prompt, request settings, batch, returned text, and usage information.
If an added request does not return the required bare probability, we retry it with output-token limits of 1{,}024, 5{,}120, and 128{,}000 tokens after the initial 512-token request, leaving every other request setting unchanged.
For one of the 29{,}169 new predictions, the 128{,}000-token attempt and one exact resubmission both returned the nonnumeric output \texttt{J}. We made one final call for this prediction with the same judge, display, reasoning effort, and output-token limit, but changed the formatting instruction to request a one-field JSON object and used a strict schema to require a numeric probability in $[0,1]$. We retained its value only after it passed the schema and range checks; no score was imputed.
Excluding the affected question, or replacing the recovered probability with either endpoint of the allowed range, changes any point estimate by at most $0.00075$ and preserves every contrast direction. Excluding the question also preserves whether each paired interval crosses zero.
After these checks, the analysis contains 41{,}670 display--item--run predictions: 12{,}501 reused predictions and 29{,}169 new predictions. We use the first valid response for each prediction, require every retained value to lie in $[0,1]$, and attach correctness labels only after checking the complete set.
Across-item AUROC pools 4{,}167 item--run observations per display, while within-item AUROC compares 2{,}778 correct--incorrect run pairs. We report all ten display estimates and seven paired differences: $S-R$, $C-S$, $(R+S)-R$, $(R+C)-(R+S)$, $(q+R+S)-(q+R)$, $(q+R+C)-(q+R)$, and $(q+R+S+C)-(q+R+C)$. A 5{,}000-draw item bootstrap resamples the same questions jointly across all displays, preserving the pairing in the intervals for these differences.

\paragraph{Baseline controls.}
Selected-answer and option-bias baselines test whether answer identity alone explains response-only discrimination.
The baselines are logistic models on the one-hot selected answer, a train-split answer-prior encoding of the selected answer, answer-format indicators such as answer length and digit or math-character flags, and, where replay probability features are available, the selected-answer probability alone.
They use the same splits, candidate regularization values, and held-out evaluation as the primary probes (\Cref{tab:baseline_controls}).
Verification variables computed by comparing the extracted answer against the gold answer encode the label and are excluded from all baseline and monitor inputs.

\paragraph{Transfer probes.}
Transfer experiments retrain the public monitor on one or more source settings and evaluate it, without refitting, on a held-out target setting, using the target setting's split assignments so transfer metrics remain comparable to in-setting metrics.
The transfer table reports the setups rerun under the updated Omni-MATH-2-Filtered response-label version, covering six cross-model transfers within that benchmark and ten same-model transfers between MMLU-Pro and Omni-MATH-2-Filtered.
Each setup is evaluated at $R$, $R+S$, $R+C$, $R+S+C$, and $q+R+S+C$ for five target splits (\Cref{tab:transfer_matrix}).
For same-benchmark cross-model transfer, source runs are aligned to the target item's train/validation/test split assignment before fitting, so every target-test item is excluded from source train and validation.
For cross-benchmark transfer, source and target item identifiers come from different benchmark item universes.

\paragraph{Additional comparisons and evaluation measures.}
For each setting and random split, we compute a prespecified set of paired AUROC differences for the public display subsets, prompt-conditioned controls, prompt-overlap-stripped displays, and replay-feature decomposition.
Shapley contributions for $R$, $S$, and $C$~\cite{shapley1953value} average each source's marginal AUROC contribution over all $3!=6$ source orderings, with the empty display assigned chance AUROC 0.5, so the three contributions of a setting sum to its full public AUROC minus 0.5.
The primary and secondary open prompt-conditioned Shapley controls hold $q$ fixed as the baseline and allocate $\mathrm{AUROC}(q+R+S+C)-\mathrm{AUROC}(q)$ over $R$, $S$, and $C$ using their complete factorials.
We interpret these Shapley values only as descriptive allocations of AUROC gains among displayed components; they have no information-theoretic interpretation.
Selective-risk curves rank held-out test runs by predicted correctness and report the error rate among retained runs at coverage levels $\{0.5,0.7,0.8,0.9,1.0\}$.
For post-hoc calibration, a one-dimensional Platt calibrator is fit on validation predictions for each benchmark, model, access level, feature family, split, and run, then applied to held-out test predictions.
Calibration tables report raw versus calibrated Brier score and expected calibration error, while AUROC remains the ranking metric.
A paired item bootstrap with 1{,}000 draws per setting and split resamples test items under fixed pairing to check that subset and prompt contrasts are not driven by a few items.
Its pooled intervals include between-setting heterogeneity, so setting-level bootstrap intervals remain the primary uncertainty summary.
Because the 15 settings reuse benchmarks across models and models within families, we report leave-one-family-out and leave-one-benchmark-out aggregates and an item-weighted aggregate for the prompt-withheld full ladder, which keep the gain between $+0.178$ and $+0.262$ ($+0.251$ item weighted, \Cref{tab:robustness_loo}).
We also report per-setting item-clustered bootstrap intervals that respect item-level clustering of the paired design (\Cref{tab:clustered_bootstrap}).
We fit intercept-only linear mixed models to each prompt-visible and prompt-withheld ladder increment across the 15 settings. The preferred specification includes benchmark and model-family variance components; a fit that does not converge with finite fixed-effect uncertainty uses the predefined benchmark-only fallback. Modeled estimates differ slightly from the paired means by construction.
All modeled point estimates are positive, but the 95\% interval for $q\Delta_S$ under the fallback includes zero (\Cref{tab:hierarchical_deltas}).

\paragraph{Failure-mode stratification.}
Held-out predictions of the full-public monitor are stratified by parser status, verification status, category and domain where defined, choice count, the presence of explicit self-correction markers in the trace, and within-setting trace- and summary-length terciles.
Within-stratum AUROC is reported only for strata with at least 20 test runs and both correctness classes (\Cref{tab:failure_stratification}).
Correctness-class slices and high-score wrong cases are treated as descriptive error analyses rather than AUROC strata, because AUROC is undefined within a single correctness class.

\paragraph{Metrics and aggregation.}
The primary metric is AUROC for final-answer correctness prediction.
AUROC is the probability that a randomly chosen correct run receives a higher score than a randomly chosen incorrect run, with half credit for ties. The marginal correctness rate does not mechanically shift AUROC, although it affects uncertainty, precision--recall baselines, and calibration (\Cref{tab:correctness_prevalence}).
AUPRC, expected calibration error, and Brier score are secondary.
ECE is computed with 10 equal-width probability bins.
Main secondary-metric tables use raw held-out logistic scores, and a separate post-hoc calibration table reports validation-fitted Platt calibration.
Scores above $0.8$ are therefore described as high-score errors when wrong unless the calibrated analysis is being discussed.
For each reported setting, the primary estimate is
\[
\mathrm{AUROC}_{m,b,k,f},
\]
where $m$ indexes target model, $b$ indexes benchmark, $k$ indexes access level, and $f$ indexes feature family.
We compute AUROC per split, average the five splits within each model--benchmark--level setting, compute ladder deltas within setting, and report means across the 15 model--benchmark settings.
We repeat the same calculations for every random split across the display subsets, prompt controls, stripped displays, and decomposition displays.
Standard deviations in detailed tables are descriptive summaries of between-split or between-setting variation, not confidence intervals.
For setting-level uncertainty, we resample the 15 model--benchmark settings with replacement to form nonparametric 95\% bootstrap intervals. These intervals summarize the 15 observed settings; they are not generalization intervals over arbitrary tasks or model families.
The interval for the normalized excess-AUROC ratio $r_{\mathrm{AUROC}}$ (defined in \Cref{sec:public_monitor_result}) is obtained by jointly resampling the 15 paired setting means and recomputing the ratio in each of 2{,}000 bootstrap draws.
We also report one-sided paired sign-flip tests of the setting-level mean against zero as descriptive paired-consistency checks, since the 15 settings share benchmarks, model families, prompts, and sometimes item identities.
The prompt-visible deltas $q\Delta_S$, $q\Delta_C$, $q\Delta_Z$, $q\Delta_{\mathrm{public}}$, and $q\Delta_{\mathrm{full}}$ and the prompt-withheld deltas $\Delta_S$, $\Delta_C$, $\Delta_Z$, $\Delta_{\mathrm{public}}$, and $\Delta_{\mathrm{full}}$ are defined in \Cref{sec:ladder_design}.

\paragraph{Controls.}
The no-length surface version tests whether direct length features account for public-text gains.
TF-IDF and Cohere embeddings are reported separately from the combined public-text monitor.
A no-silent-truncation check confirms whole-context fit and records chunking for public embeddings.
Replay-feature sensitivity reports answer-probability-only, hidden-state-only, combined-replay, and replay-only probes.
Surface-feature sensitivity compares the primary rate-based surface family against the no-length version, with raw marker-count and lexical-diversity versions retained as diagnostics rather than current main-text controls.
Parser and replay-status controls test whether extraction or context-fit failures explain apparent access-level gains.
Summary-length controls hard-cap primary summaries at 90 words, replace the summary with first, last, head--tail, or random trace extracts matched to each capped summary's word count, and compare the fixed Last90 extract with a masked Last90 version under the primary and open feature families. The separate fixed-monitor sensitivity varies summary decoding as described above.
The matched primary chunk control uses every primary-run trace in the 15 model--benchmark settings and partitions each trace into up to eight balanced contiguous word-position chunks targeting 512 words; a sufficiently short trace forms one chunk. Each chunk receives a fixed 512-dimensional Cohere embedding. For each of five splits, held-out item scores come from logistic probes over mean, elementwise maximum, or elementwise top-two pooled chunk embeddings, or from a linear-attention multiple-instance head, with validation-selected regularization or early stopping. A secondary open surface-plus-TF-IDF control fits individual chunks and combines held-out chunk scores by maximum, top-two mean, mean, or last-or-only chunk. These checks show whether the results depend on how the trace is divided and read; they do not identify where causal evidence appears.
Train-only preprocessing, paired split assignments, and held-out test evaluation are the main protections against leakage from labels or validation data into the reported AUROC estimates.
The reported analyses include the output-only subsets, every primary and open prompt-conditioned combination, prompt-overlap stripping, summary-instruction and decoding sensitivities, baseline controls, transfer probes, and failure stratification described above.

\clearpage
\section{Appendix Figures}
\label{sec:appendix_figures}

\begin{figure}[H]
\centering
\begin{promptbox}
<|im_start|>system
Solve the following problem. Please make sure that your response only consists of a single letter corresponding to the correct answer choice. Do not include anything else in your final response.<|im_end|>
<|im_start|>user
Among the following exoplanets, which one has the highest density?

a) An Earth-mass and Earth-radius planet.
b) A planet with 2 Earth masses and a density of approximately 5.5 g/cm^3.
c) A planet with the same composition as Earth but 5 times more massive than Earth.
d) A planet with the same composition as Earth but half the mass of Earth.

A. d
B. a
C. b
D. c<|im_end|>
<|im_start|>assistant
\end{promptbox}
\caption{\textbf{Example target prompt for Qwen3-4B on GPQA-Diamond.} The box shows the Qwen chat-template string for item 1 of GPQA-Diamond under the multiple-choice system instruction used to produce the completed run.}
\label{fig:example_qwen_prompt}
\end{figure}

\begin{figure}[H]
\centering
\begin{promptbox}
<|im_start|>assistant
<think>
Okay, let's try to figure out which exoplanet has the highest density. The question gives me four options, and I need to determine which one has the highest density. Let me start by recalling what density is. Density is mass divided by volume, right? So, the formula is density = mass/volume.

Now, the problem is about exoplanets, so I need to think about how mass and radius relate to density. For planets, if they have the same composition, like Earth, their density would depend on their mass and radius. But if the mass changes, the radius might change as well, depending on how the planet is structured.

Let me look at each option one by one.

Option a) An Earth-mass and Earth-radius planet. So, this is just Earth. Earth's density is about 5.5 g/cm^3.

Option b) A planet with 2 Earth masses and a density of approximately 5.5 g/cm^3. So, this is a planet that is twice as massive as Earth but has the same density as Earth.

[... reasoning trace omitted for display ...]

But given that the question is from a test, it's likely that the intended answer is option c. Because for a planet with same composition, more mass implies higher density, since it's more compressed. Therefore, option c has higher density than the others.

So, in that case, the answer is option c, which is D.
</think>

D
\end{promptbox}
\caption{\textbf{Example completed response from Qwen3-4B on GPQA-Diamond.} The box shows item 1 from the primary run, shortened for display. The full reasoning trace is used in the analysis. The parsed final response is $R=D$, and the extracted answer matches the GPQA-Diamond gold label.}
\label{fig:example_qwen_completed}
\end{figure}

\begin{figure}[H]
\centering
\begin{promptbox}
<|im_start|>system
Compress the reasoning trace into at most 90 words without solving the problem again. Preserve uncertainty, contradictions, candidate answers, and self-corrections. Do not add evidence, verify the answer, repair mistakes, or mention information absent from the trace.<|im_end|>
<|im_start|>user
Reasoning trace:
Okay, let's try to figure out which exoplanet has the highest density. The question gives me four options, and I need to determine which one has the highest density. Let me start by recalling what density is. Density is mass divided by volume, right?

[... full reasoning trace supplied to the summarizer ...]

So, in that case, the answer is option c, which is D.

Summary:<|im_end|>
<|im_start|>assistant
The reasoning traces uncertainty about density calculations, noting that same-composition planets with higher mass might have higher density due to compression, while others have same density. Contradictions arise between scaling assumptions and physical compression effects. Candidate answers include options a, b, c, and d, with c potentially having higher density due to compression. Self-corrections suggest uncertainty in whether same-composition mass scaling leads to same or higher density.
\end{promptbox}
\caption{\textbf{Example self-summary prompt and output for the summary level.} The summarizer receives only the reasoning trace $C$, not the task prompt, gold answer, correctness label, logits, or hidden states. The displayed prompt is shortened. The generated summary is the output for item 1 from Qwen3-4B on GPQA-Diamond. Its stored file name records a 256-token decoding limit, while the instruction requests at most 90 words.}
\label{fig:example_qwen_summary}
\end{figure}

\begin{figure}[b]
\centering
\includegraphics[width=0.82\textwidth]{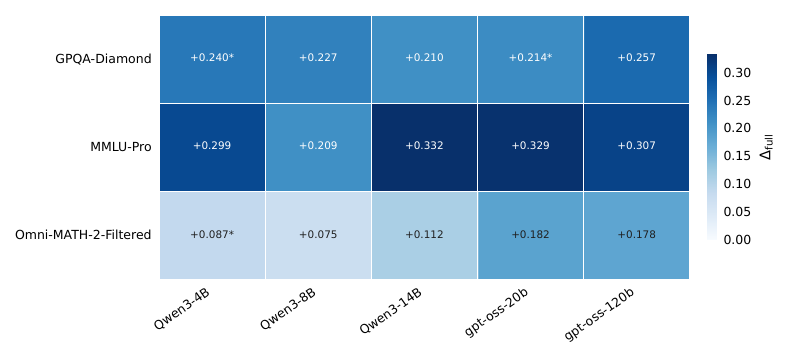}
\caption{\textbf{Full-ladder gains across the model--benchmark matrix.} Each tile is one target model and one benchmark. Color shows the paired full-ladder gain $\Delta_{\mathrm{full}}=\mathrm{AUROC}(R+S+C+Z)-\mathrm{AUROC}(R)$, also printed on each tile. An asterisk marks settings whose public-level point estimates do not satisfy $R < R+S < R+S+C$.}
\label{fig:cell_heatmap}
\end{figure}

\begin{figure}[t]
\centering
\includegraphics[width=\textwidth]{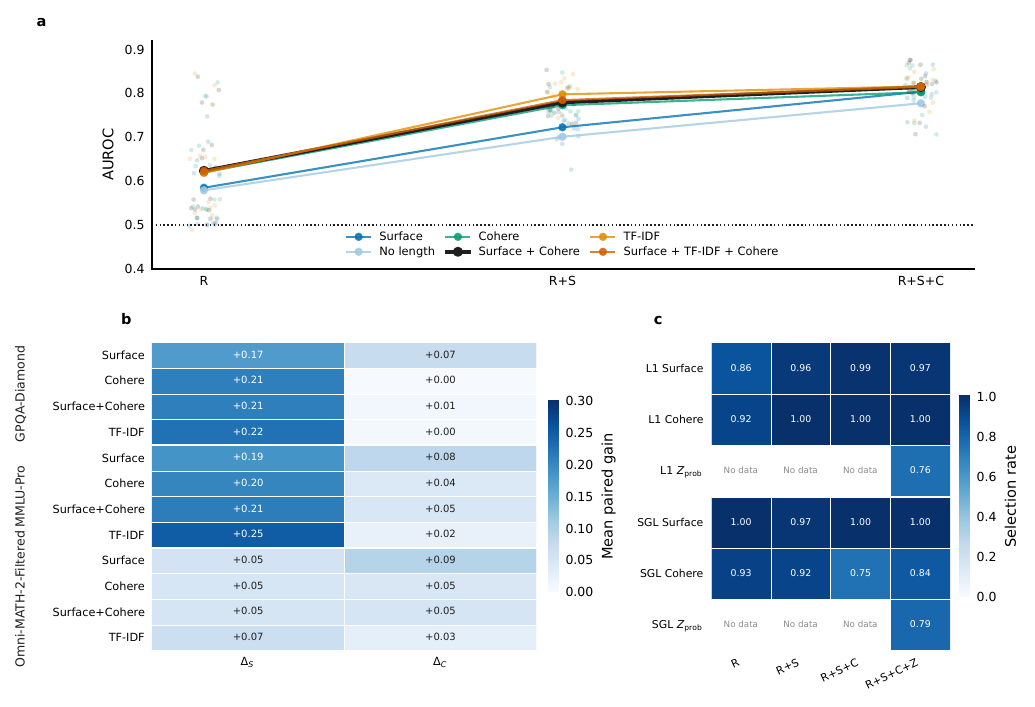}
\caption{\textbf{The mean public access pattern persists across the tested feature families.} Public monitors use only the text exposed at the corresponding level, never the target model's hidden states, logits, token probabilities, or the original question. \textbf{a}, Mean public AUROC increases across access levels for surface features, the no-length version, Cohere embed-v4.0, TF-IDF, and their combinations. Faint points show available model--benchmark settings. All displayed feature families cover the 15 model--benchmark settings. \textbf{b}, Per-benchmark paired step gains show that the summary step $\Delta_S$ is concentrated on GPQA-Diamond and MMLU-Pro, while Omni-MATH-2-Filtered spreads its public gain more evenly. \textbf{c}, Plain LASSO and sparse group LASSO selection rates show that both surface features and embeddings enter the probes. Answer-probability features appear only at the top level and are not part of the public monitor.}
\label{fig:public_features}
\end{figure}

\begin{figure}[t]
\centering
\includegraphics[width=\textwidth]{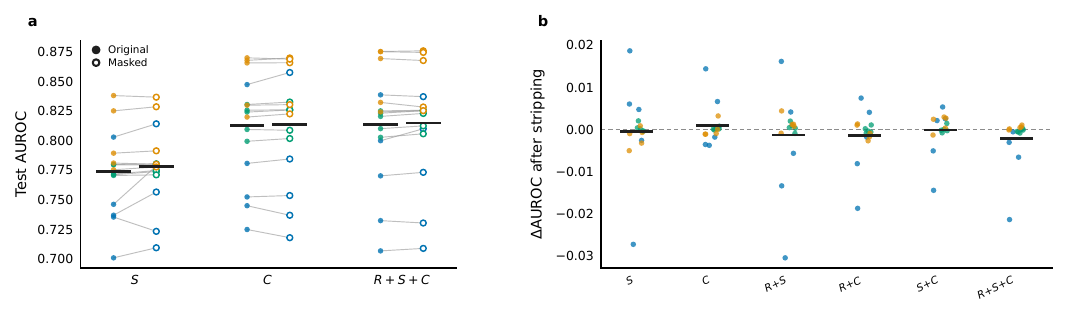}
\caption{\textbf{Masking and prompt-overlap stripping leave the public signal essentially unchanged.} \textbf{a}, Summary, trace, and full public displays after replacing stated answers, option text, and prompt-derived numbers with fixed markers (open circles) match the unmasked originals (filled circles). Gray lines connect the same setting. The combined mask shifts mean trace AUROC by $+0.001$ (\Cref{tab:masked_controls}). \textbf{b}, Removing 8--14-word spans copied from the question changes mean AUROC by between $-0.002$ and $+0.001$ across the six summary- and trace-containing displays (\Cref{tab:noqcopy_controls}).}
\label{fig:hardening_controls}
\end{figure}

\begin{figure}[t]
\centering
\includegraphics[width=\textwidth]{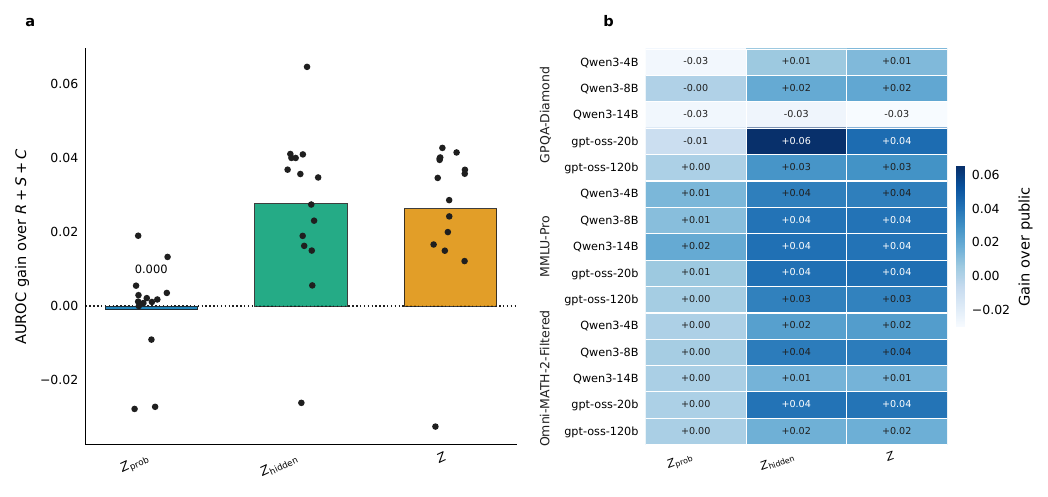}
\caption{\textbf{The provider-side replay gain is mostly a hidden-state signal.} Gains are paired AUROC differences relative to the full public display $P=R+S+C$. \textbf{a}, The average breakdown compares answer-probability, hidden-state, and combined replay features against the same full public baseline. Bars are means across settings and black dots show setting-level values; the annotated answer-probability bar has a mean gain of $0.000$. \textbf{b}, The model-by-benchmark heatmap shows where the top-level gain appears for those three replay feature families.}
\label{fig:internal_decomposition}
\end{figure}

\begin{figure}[t]
\centering
\includegraphics[width=0.6\textwidth]{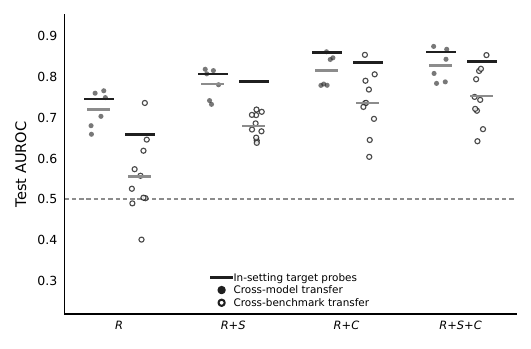}
\caption{\textbf{Public probes transfer across models and benchmarks.} Transfer tests fit public probes in one setting and evaluate them, without refitting, on held-out settings. Filled circles show cross-model transfer within Omni-MATH-2-Filtered, open circles show cross-benchmark transfer between MMLU-Pro and Omni-MATH-2-Filtered for the same model, black ticks mark matched in-setting target means, and gray ticks mark transfer means. Cross-model transfer loses modest discrimination relative to matched in-setting target probes ($0.826$ versus $0.860$ mean AUROC for $R+S+C$), while cross-benchmark transfer is lower, with mean AUROC $0.752$ for $R+S+C$ and individual settings below chance at the shortest displays.}
\label{fig:transfer}
\end{figure}

\clearpage

\section{Appendix Tables}
\label{sec:appendix_tables}

% ---- Setup and definitions ----
\begin{table}[H]
\centering
\caption{\textbf{Notation for the observability ladder.} Components are derived from one fixed completed run per item. Levels are the displays a monitor receives.}
\label{tab:notation}
\small
\begin{adjustbox}{max width=\textwidth}
\begin{tabular}{@{}L{1.6cm}L{4.4cm}L{7.2cm}@{}}
\toprule
Symbol & Object & Definition \\
\midrule
$q$ & Task prompt & Benchmark question text; visible to prompt-conditioned monitors. \\
$R$ & Final response & The model's final response text from the completed run. \\
$S$ & Self-summary & Post-hoc self-summary of the trace, generated from $C$ only by the same target model under an instruction requesting at most 90 words. \\
$C$ & Reasoning trace & The full reasoning trace of the completed run. \\
$Z$ & Provider-only replay features & Answer-probability ($Z_{\mathrm{prob}}$) and hidden-state ($Z_{\mathrm{hidden}}$) features extracted from a prompt-conditioned replay of the target model on $q$ and $C$. \\
\addlinespace
$\mathcal{O}^{q}_1$ & Prompt-visible level 1 & $q+R$ \\
$\mathcal{O}^{q}_2$ & Prompt-visible level 2 & $q+R+S$ \\
$\mathcal{O}^{q}_3$ & Prompt-visible level 3 & $q+R+S+C$ (full prompt-visible public display) \\
$\mathcal{O}^{q}_4$ & Provider-only features & $q+R+S+C+Z$ (public display plus replay-derived features) \\
\bottomrule
\end{tabular}
\end{adjustbox}
\end{table}

\begin{table}[H]
\centering
\caption{\textbf{The model set gives a within-family scale sweep and a second-family contrast.} All target models are treated as locally inspectable open-weight systems for the provider-only replay features. Public levels restrict the monitor input to model-produced visible text. The final level adds hidden states and token probabilities from a prompt-conditioned replay of the target model. gpt-oss parameter counts are total parameters, with active parameters shown in parentheses. gpt-oss inference uses the model-specific Transformers code path and locally stored Hugging Face kernel components required by the released weights.}
\label{tab:model_access}
\small
\begin{adjustbox}{max width=\textwidth}
\begin{tabular}{@{}L{3.1cm}L{1.7cm}L{2.2cm}L{2.0cm}L{2.2cm}R{1.8cm}@{}}
\toprule
Model & Family & Size & Hidden states & Token probabilities & Context window \\
\midrule
Qwen/Qwen3-4B & Qwen3 & 4B & Full & Full & 32{,}768 \\
\addlinespace
Qwen/Qwen3-8B & Qwen3 & 8B & Full & Full & 32{,}768 \\
\addlinespace
Qwen/Qwen3-14B & Qwen3 & 14B & Full & Full & 32{,}768 \\
\addlinespace
openai/gpt-oss-20b & gpt-oss & 21B (3.6B active) & Full & Full & 128{,}000 \\
\addlinespace
openai/gpt-oss-120b & gpt-oss & 117B (5.1B active) & Full & Full & 128{,}000 \\
\bottomrule
\end{tabular}
\end{adjustbox}
\end{table}

\begin{table}[H]
\centering
\caption{\textbf{Settings used to generate reasoning traces and summaries and to calculate model-based measurements.} The reported analysis can be reproduced from the saved outputs even when the original text generation was stochastic. Reasoning traces were generated with vLLM's default sampling settings except for the token limit. Summaries were generated with Hugging Face \texttt{generate} using each loaded model's generation settings, with no explicit temperature or top-$p$ override.}
\label{tab:generation_defaults}
\small
\begin{adjustbox}{max width=\textwidth}
\begin{tabular}{@{}L{2.6cm}L{1.7cm}L{1.6cm}L{1.5cm}L{2.3cm}L{2.4cm}L{3.2cm}L{2.0cm}@{}}
\toprule
Component & Temperature & Top-$p$ & Top-$k$ & Random seed & Max tokens & Stopping behavior & Deterministic? \\
\midrule
Trace generation (vLLM chat template) & 1.0 & 1.0 & 0 & None & Qwen: 28{,}000; gpt-oss: 126{,}000 & no added stopping rule; model EOS and chat-template stopping apply & No; the analysis uses saved traces \\
\addlinespace
Summary generation (HF, \texttt{do\_sample=True}, \texttt{use\_cache=False}) & model setting; default 1.0 if unset & model setting; default 1.0 if unset & model setting; default 50 if unset & seed identifier 0 recorded; random-number generator not fixed & 256 minus any final-channel prefill & EOS used for padding; summaries were not shortened afterward & No; the analysis uses saved summaries \\
\addlinespace
Calculation of $Z$ (forward pass only) & not applicable; no sampling & not applicable; no sampling & not applicable; no sampling & not applicable; deterministic forward pass & not applicable; no new text generated & model-specific final suffixes; bare-label and answer-span scores taken at the final position & Yes, apart from numeric differences across hardware and libraries \\
\bottomrule
\end{tabular}
\end{adjustbox}
\end{table}

\begin{table}[H]
\centering
\caption{\textbf{The benchmark set includes multiple-choice and open-response tasks.} GPQA-Diamond and MMLU-Pro use answer-choice log-probability features in $Z_{\mathrm{prob}}$. Omni-MATH-2-Filtered uses answer-span log-probability features for the model's own extracted final answer, including answer-token and answer-character counts and the sum of token log probabilities.}
\label{tab:task_taxonomy}
\small
\begin{adjustbox}{max width=\textwidth}
\begin{tabular}{@{}L{2.8cm}L{2.5cm}R{1.2cm}L{2.5cm}L{3.5cm}@{}}
\toprule
Benchmark & Domain & Size & Answer type & Verification and $Z_{\mathrm{prob}}$ \\
\midrule
GPQA-Diamond & Graduate science & 198 & Four-way multiple choice & Option match; log-probabilities over A--D \\
\addlinespace
MMLU-Pro & Multi-task language understanding & 12{,}032 & Up to ten-way multiple choice & Option match; log probabilities for fixed choices A--J, with unavailable choices masked out \\
\addlinespace
Omni-MATH-2-Filtered & Olympiad mathematics & 4{,}181 & Open-response numeric / expression & Stored GPT-5-mini judgment of whether the model response matches the reference answer; answer-span log-probability features for the model's extracted final answer, including answer-token and answer-character counts and the sum of token log probabilities \\
\bottomrule
\end{tabular}
\end{adjustbox}
\end{table}

\begin{table}[H]
\centering
\caption{\textbf{Correctness prevalence varies by benchmark and target model.} The table reports the binary final-answer correctness labels used in the main ladder analyses. GPQA-Diamond and MMLU-Pro use benchmark option matching. Omni-MATH-2-Filtered correctness is determined by GPT-5-mini equivalence judgments. AUROC is not mechanically shifted by these prevalences because it ranks correct against incorrect runs, but prevalence affects AUPRC baselines, calibration metrics, and uncertainty.}
\label{tab:correctness_prevalence}
\small
\begin{adjustbox}{max width=\textwidth}
\begin{tabular}{@{}llrrr@{}}
\toprule
Benchmark & Target model & $n$ & Correct & Correctness rate \\
\midrule
GPQA-Diamond & Qwen3-4B & 198 & 110 & 55.6\% \\
GPQA-Diamond & Qwen3-8B & 198 & 115 & 58.1\% \\
GPQA-Diamond & Qwen3-14B & 198 & 124 & 62.6\% \\
GPQA-Diamond & gpt-oss-20b & 198 & 124 & 62.6\% \\
GPQA-Diamond & gpt-oss-120b & 198 & 138 & 69.7\% \\
\addlinespace
MMLU-Pro & Qwen3-4B & 12{,}032 & 8{,}550 & 71.1\% \\
MMLU-Pro & Qwen3-8B & 12{,}032 & 9{,}073 & 75.4\% \\
MMLU-Pro & Qwen3-14B & 12{,}032 & 9{,}375 & 77.9\% \\
MMLU-Pro & gpt-oss-20b & 12{,}032 & 8{,}945 & 74.3\% \\
MMLU-Pro & gpt-oss-120b & 12{,}032 & 9{,}555 & 79.4\% \\
\addlinespace
Omni-MATH-2-Filtered & Qwen3-4B & 4{,}181 & 2{,}751 & 65.8\% \\
Omni-MATH-2-Filtered & Qwen3-8B & 4{,}181 & 2{,}455 & 58.7\% \\
Omni-MATH-2-Filtered & Qwen3-14B & 4{,}181 & 2{,}853 & 68.2\% \\
Omni-MATH-2-Filtered & gpt-oss-20b & 4{,}181 & 3{,}016 & 72.1\% \\
Omni-MATH-2-Filtered & gpt-oss-120b & 4{,}181 & 3{,}131 & 74.9\% \\
\bottomrule
\end{tabular}
\end{adjustbox}
\end{table}

\begin{table}[H]
\centering
\caption{\textbf{Public-text embeddings are not provider-side replay features.} Public monitors receive only the model-produced text exposed at the corresponding access level. Replay features are extracted from a prompt-conditioned forward pass of the target model on the task prompt and reasoning trace. We record answer-extraction and replay outcomes separately so that failures can be examined in sensitivity checks.}
\label{tab:feature_boundary}
\small
\begin{adjustbox}{max width=\textwidth}
\begin{tabular}{@{}L{4.0cm}L{5.2cm}R{2.7cm}L{1.5cm}@{}}
\toprule
Feature family & Input to feature extractor & Uses provider-side replay? & Level \\
\midrule
Surface text & Visible public text $P$ & no & public \\
\addlinespace
No-length surface text & Visible public text $P$ with direct length/count features removed & no & public \\
\addlinespace
TF-IDF text & Visible public text $P$ & no & public \\
\addlinespace
Cohere embed-v4.0 & Visible public text $P$ & no & public \\
\addlinespace
Task prompt $q_i$ & Benchmark prompt text, rendered into the display & no & control levels only \\
\addlinespace
Target-model answer log-probabilities & $(q_i,C_i,s_{\mathrm{stop}})$ through target model; open-response spans include length-sensitive count and sum features & yes & $Z_{\mathrm{prob}}$ \\
\addlinespace
Target-model raw hidden state & $(q_i,C_i,s_{\mathrm{stop}})$ through target model & yes & $Z_{\mathrm{hidden}}$ \\
\addlinespace
Answer extraction and replay checks & Extraction outcome, replay outcome, and replay context-fit indicator & no & Checks \\
\bottomrule
\end{tabular}
\end{adjustbox}
\end{table}

\begin{table}[H]
\centering
\caption{\textbf{Primary public surface features use rates for lexical markers.} Features are computed only from the public text visible at the corresponding access level. Raw marker counts and additional length/log-count variants are retained as appendix diagnostics, but the primary surface family uses the rate-based definitions below.}
\label{tab:surface_features}
\small
\begin{adjustbox}{max width=\textwidth}
\begin{tabular}{@{}L{3.2cm}L{6.0cm}L{4.2cm}@{}}
\toprule
Feature & Definition & Intended signal \\
\midrule
Word count & Number of alphabetic word matches in the visible text & Overall trace length \\
\addlinespace
Sentence count & Number of sentence-ending punctuation markers & Reasoning segmentation \\
\addlinespace
Average word length & Mean character length of alphabetic words & Lexical form and technical vocabulary \\
\addlinespace
Digit density & Digit characters divided by total characters & Numerical reasoning content \\
\addlinespace
Math-character density & Math-like characters divided by total characters & Formulaic or symbolic content \\
\addlinespace
Uncertainty rate & Uncertainty markers divided by word count & Explicit hesitation or low-confidence language \\
\addlinespace
Self-correction rate & Self-correction markers divided by word count & Revisions, reversals, and reconsideration \\
\addlinespace
Answer-marker rate & Answer markers divided by word count & Explicit conclusion or answer-framing language \\
\bottomrule
\end{tabular}
\end{adjustbox}
\end{table}

\begin{table}[H]
\centering
\caption{\textbf{Inputs used by each correctness probe.} All probes predict the same binary target, final-answer correctness. The reference answer and correctness label are never included as inputs. They are used only to define $y_i$. Prompt-withheld public levels exclude the task prompt as a separate input and exclude provider-side replay features. The $Z$ levels add features extracted from a prompt-conditioned replay of the task prompt and reasoning trace. Across the reported comparisons, the provider-side $Z$ features appear only alongside the full public display; they are never used as a public display or monitor by themselves. The primary hidden-state model uses the raw final hidden state with $\ell_2$ shrinkage.}
\label{tab:probe_specification}
\small
\begin{adjustbox}{max width=\textwidth}
\begin{tabular}{@{}L{3.0cm}L{5.9cm}L{5.2cm}@{}}
\toprule
Probe & Feature vector in the logistic model & Information not used \\
\midrule
Response only ($R$) &
$\bm{z}_{i}=\phi_f(R_i)$ &
Task prompt $q_i$, self-summary $S_i$, reasoning trace $C_i$, replay features $Z_i$, reference answer, correctness label \\
\addlinespace
Response + summary ($R+S$) &
$\bm{z}_{i}=\phi_f(R_i\Vert S_i)$ &
Task prompt $q_i$, full reasoning trace $C_i$, replay features $Z_i$, reference answer, correctness label \\
\addlinespace
Response + summary + trace ($R+S+C$) &
$\bm{z}_{i}=\phi_f(R_i\Vert S_i\Vert C_i)$ &
Task prompt $q_i$, replay features $Z_i$, reference answer, correctness label \\
\addlinespace
Public display subsets ($S$, $C$, $R+C$, $S+C$) &
$\bm{z}_{i}=\phi_f(\cdot)$ applied to the corresponding subset of $\{R_i,S_i,C_i\}$ &
Task prompt $q_i$, replay features $Z_i$, reference answer, correctness label; used to compare all component subsets, separately from the main prompt-visible comparison \\
\addlinespace
All prompt-conditioned public combinations ($q$, $q+R$, $q+S$, $q+C$, $q+R+S$, $q+R+C$, $q+S+C$, $q+R+S+C$) &
$\bm{z}_{i}=\phi_f(q_i\Vert\cdot)$ applied to the corresponding prompt-conditioned combination &
Replay features $Z_i$, reference answer, correctness label; the primary surface-plus-Cohere analysis follows a fixed sequence of component sets, while the secondary surface-plus-TF-IDF analysis covers all eight combinations \\
\addlinespace
Displays with prompt overlap removed (\textit{noqcopy} variants of $S$/$C$-containing levels) &
$\bm{z}_{i}=\phi_f(\cdot)$ applied to the display after removing 8--14-word spans copied from $q_i$ &
Task prompt $q_i$, replay features $Z_i$, reference answer, correctness label; tests whether the visible text reconstructs the prompt \\
\addlinespace
Full public + answer probabilities ($R+S+C+Z_{\mathrm{prob}}$) &
$\bm{z}_{i}=\phi_{\mathrm{all}}(R_i\Vert S_i\Vert C_i)\oplus Z_{\mathrm{prob},i}$ &
Hidden-state features $Z_{\mathrm{hidden},i}$, reference answer, correctness label \\
\addlinespace
Full public + hidden state ($R+S+C+Z_{\mathrm{hidden}}$) &
$\bm{z}_{i}=\phi_{\mathrm{all}}(R_i\Vert S_i\Vert C_i)\oplus Z_{\mathrm{hidden},i}$ &
Answer-probability features $Z_{\mathrm{prob},i}$, reference answer, correctness label \\
\addlinespace
Full public + all replay features ($R+S+C+Z$) &
$\bm{z}_{i}=\phi_{\mathrm{all}}(R_i\Vert S_i\Vert C_i)\oplus Z_{\mathrm{prob},i}\oplus Z_{\mathrm{hidden},i}$ &
Reference answer and correctness label \\
\addlinespace
Prompt-conditioned public + replay features ($q+R+S+C+Z_{\mathrm{hidden}}$, $q+R+S+C+Z$) &
$\bm{z}_{i}=\phi_{\mathrm{all}}(q_i\Vert R_i\Vert S_i\Vert C_i)\oplus Z_i$ or the hidden-state features alone &
Reference answer and correctness label; used to separate information in prompt-conditioned public text from information in replay features \\
\bottomrule
\end{tabular}
\end{adjustbox}
\end{table}

\clearpage
% ---- Access-ladder results: uncertainty and robustness ----
\begin{table}[H]
\centering
\caption{\textbf{Prompt-withheld observability-ladder AUROC by model--benchmark setting.} Values are mean $\pm$ descriptive SD across data splits, not confidence intervals, for the primary public feature family (rate surface plus Cohere) and the provider-side replay feature family (rate surface plus Cohere plus replay-derived features). Omni-MATH-2-Filtered correctness is determined by GPT-5-mini equivalence judgments.}
\label{tab:main_ladder_15cell}
\small
\begin{adjustbox}{max width=\textwidth}
\begin{tabular}{@{}L{2.2cm}L{2.3cm}R{1.75cm}R{1.75cm}R{1.75cm}R{1.75cm}R{1.55cm}@{}}
\toprule
Benchmark & Model & $R$ & $R+S$ & $R+S+C$ & $R+S+C+Z$ & $\Delta_{\mathrm{full}}$ \\
\midrule
GPQA-Diamond & Qwen3-4B & 0.505 $\pm$ 0.066 & 0.732 $\pm$ 0.063 & 0.732 $\pm$ 0.067 & 0.744 $\pm$ 0.059 & +0.240 $\pm$ 0.105 \\
GPQA-Diamond & Qwen3-8B & 0.631 $\pm$ 0.047 & 0.812 $\pm$ 0.082 & 0.839 $\pm$ 0.062 & 0.859 $\pm$ 0.038 & +0.227 $\pm$ 0.053 \\
GPQA-Diamond & Qwen3-14B & 0.558 $\pm$ 0.046 & 0.748 $\pm$ 0.056 & 0.800 $\pm$ 0.058 & 0.767 $\pm$ 0.078 & +0.210 $\pm$ 0.100 \\
GPQA-Diamond & gpt-oss-20b & 0.535 $\pm$ 0.137 & 0.755 $\pm$ 0.066 & 0.707 $\pm$ 0.096 & 0.749 $\pm$ 0.103 & +0.214 $\pm$ 0.074 \\
GPQA-Diamond & gpt-oss-120b & 0.542 $\pm$ 0.073 & 0.761 $\pm$ 0.053 & 0.770 $\pm$ 0.057 & 0.799 $\pm$ 0.069 & +0.257 $\pm$ 0.123 \\
\addlinespace
MMLU-Pro & Qwen3-4B & 0.560 $\pm$ 0.014 & 0.764 $\pm$ 0.009 & 0.823 $\pm$ 0.010 & 0.859 $\pm$ 0.008 & +0.299 $\pm$ 0.017 \\
MMLU-Pro & Qwen3-8B & 0.652 $\pm$ 0.014 & 0.774 $\pm$ 0.013 & 0.821 $\pm$ 0.004 & 0.861 $\pm$ 0.007 & +0.209 $\pm$ 0.011 \\
\bottomrule
\end{tabular}
\end{adjustbox}
\end{table}

\begin{table}[H]
\centering
\caption{\textbf{Paired AUROC increments along the observability ladder.} Each increment is computed within each data split before averaging. The $\pm$ values are descriptive SDs across data splits, not confidence intervals. Descriptive setting-bootstrap intervals over the observed design are reported separately in \Cref{tab:ladder_delta_uncertainty}. Omni-MATH-2-Filtered correctness is determined by GPT-5-mini equivalence judgments.}
\label{tab:ladder_deltas_15cell}
\small
\begin{adjustbox}{max width=\textwidth}
\begin{tabular}{@{}L{2.2cm}L{2.3cm}R{1.8cm}R{1.8cm}R{1.8cm}R{1.8cm}@{}}
\toprule
Benchmark & Model & $\Delta_S$ & $\Delta_C$ & $\Delta_Z$ & $\Delta_{\mathrm{full}}$ \\
\midrule
GPQA-Diamond & Qwen3-4B & +0.228 $\pm$ 0.106 & +0.000 $\pm$ 0.038 & +0.012 $\pm$ 0.023 & +0.240 $\pm$ 0.105 \\
GPQA-Diamond & Qwen3-8B & +0.181 $\pm$ 0.055 & +0.027 $\pm$ 0.023 & +0.020 $\pm$ 0.051 & +0.227 $\pm$ 0.053 \\
GPQA-Diamond & Qwen3-14B & +0.190 $\pm$ 0.057 & +0.052 $\pm$ 0.033 & $-$0.033 $\pm$ 0.051 & +0.210 $\pm$ 0.100 \\
GPQA-Diamond & gpt-oss-20b & +0.220 $\pm$ 0.123 & $-$0.049 $\pm$ 0.082 & +0.043 $\pm$ 0.125 & +0.214 $\pm$ 0.074 \\
GPQA-Diamond & gpt-oss-120b & +0.219 $\pm$ 0.117 & +0.010 $\pm$ 0.027 & +0.029 $\pm$ 0.026 & +0.257 $\pm$ 0.123 \\
\addlinespace
MMLU-Pro & Qwen3-4B & +0.205 $\pm$ 0.019 & +0.059 $\pm$ 0.002 & +0.036 $\pm$ 0.003 & +0.299 $\pm$ 0.017 \\
MMLU-Pro & Qwen3-8B & +0.122 $\pm$ 0.006 & +0.046 $\pm$ 0.012 & +0.040 $\pm$ 0.003 & +0.209 $\pm$ 0.011 \\
MMLU-Pro & Qwen3-14B & +0.241 $\pm$ 0.010 & +0.051 $\pm$ 0.009 & +0.040 $\pm$ 0.006 & +0.332 $\pm$ 0.007 \\
MMLU-Pro & gpt-oss-20b & +0.251 $\pm$ 0.010 & +0.036 $\pm$ 0.008 & +0.041 $\pm$ 0.005 & +0.329 $\pm$ 0.005 \\
MMLU-Pro & gpt-oss-120b & +0.231 $\pm$ 0.009 & +0.042 $\pm$ 0.003 & +0.035 $\pm$ 0.003 & +0.307 $\pm$ 0.009 \\
\addlinespace
Omni-MATH-2-Filtered & Qwen3-4B & $-$0.004 $\pm$ 0.008 & +0.067 $\pm$ 0.004 & +0.024 $\pm$ 0.005 & +0.087 $\pm$ 0.009 \\
Omni-MATH-2-Filtered & Qwen3-8B & +0.016 $\pm$ 0.007 & +0.022 $\pm$ 0.005 & +0.037 $\pm$ 0.013 & +0.075 $\pm$ 0.017 \\
Omni-MATH-2-Filtered & Qwen3-14B & +0.042 $\pm$ 0.008 & +0.055 $\pm$ 0.006 & +0.015 $\pm$ 0.010 & +0.112 $\pm$ 0.015 \\
Omni-MATH-2-Filtered & gpt-oss-20b & +0.093 $\pm$ 0.019 & +0.049 $\pm$ 0.013 & +0.039 $\pm$ 0.021 & +0.182 $\pm$ 0.026 \\
Omni-MATH-2-Filtered & gpt-oss-120b & +0.106 $\pm$ 0.015 & +0.056 $\pm$ 0.015 & +0.017 $\pm$ 0.013 & +0.178 $\pm$ 0.010 \\
\bottomrule
\end{tabular}
\end{adjustbox}
\end{table}

\begin{table}[H]
\centering
\caption{\textbf{Setting-level resampling describes the prompt-visible and prompt-withheld ladder contrasts.} Mean deltas and descriptive SDs are computed across the 15 model--benchmark settings after averaging the five data splits within each setting. Intervals are descriptive nonparametric 95\% bootstrap intervals over the 15 observed settings, not population-generalization intervals. The $p$ value is a one-sided paired sign-flip test of the setting-level mean against zero in the tested direction: positive for ladder increments and negative for prompt-visible-minus-prompt-withheld differences. Values within $10^{-12}$ AUROC of zero are treated as ties.}
\label{tab:ladder_delta_uncertainty}
\small
\begin{adjustbox}{max width=\textwidth}
\begin{tabular}{@{}L{2.4cm}L{2.0cm}L{3.8cm}R{1.0cm}R{1.5cm}R{2.5cm}R{1.5cm}R{1.0cm}@{}}
\toprule
Condition & Quantity & Contrast & Mean & Between-setting SD & 95\% bootstrap interval & Settings in tested direction & $p$ \\
\midrule
Prompt-visible & $q\Delta_S$ & $q{+}R$ to $q{+}R{+}S$ & +0.019 & 0.017 & [+0.011, +0.027] & 11/15 & $<0.001$ \\
\addlinespace
Prompt-visible, trace first & $q\Delta_{C\mid R}$ & $q{+}R$ to $q{+}R{+}C$ & +0.062 & 0.033 & [+0.046, +0.077] & 15/15 & $<0.001$ \\
\addlinespace
Prompt-visible & $q\Delta_C$ & $q{+}R{+}S$ to $q{+}R{+}S{+}C$ & +0.041 & 0.026 & [+0.028, +0.052] & 14/15 & $<0.001$ \\
\addlinespace
Prompt-visible, trace first & $q\Delta_{S\mid R,C}$ & $q{+}R{+}C$ to $q{+}R{+}S{+}C$ & $-$0.002 & 0.009 & [$-$0.007, +0.002] & 8/15 & 0.819 \\
\addlinespace
Prompt-visible direct comparison & $q(C{-}S)$ & $q{+}R{+}S$ versus $q{+}R{+}C$ & +0.043 & 0.020 & [+0.033, +0.052] & 15/15 & $<0.001$ \\
\addlinespace
Prompt-visible & $q\Delta_Z$ & $q{+}R{+}S{+}C$ to $q{+}R{+}S{+}C{+}Z$ & +0.029 & 0.014 & [+0.023, +0.036] & 15/15 & $<0.001$ \\
Prompt-visible & $q\Delta_{\mathrm{public}}$ & $q{+}R$ to $q{+}R{+}S{+}C$ & +0.060 & 0.039 & [+0.040, +0.077] & 14/15 & $<0.001$ \\
Prompt-visible & $q\Delta_{\mathrm{full}}$ & $q{+}R$ to $q{+}R{+}S{+}C{+}Z$ & +0.089 & 0.045 & [+0.066, +0.110] & 15/15 & $<0.001$ \\
\addlinespace
Prompt-withheld & $\Delta_S$ & $R$ to $R{+}S$ & +0.156 & 0.087 & [+0.113, +0.196] & 14/15 & $<0.001$ \\
Prompt-withheld & $\Delta_C$ & $R{+}S$ to $R{+}S{+}C$ & +0.035 & 0.030 & [+0.019, +0.048] & 13/15 (+1 tie) & $<0.001$ \\
Prompt-withheld & $\Delta_Z$ & $R{+}S{+}C$ to $R{+}S{+}C{+}Z$ & +0.026 & 0.019 & [+0.016, +0.034] & 14/15 & $<0.001$ \\
Prompt-withheld & $\Delta_{\mathrm{public}}$ & $R$ to $R{+}S{+}C$ & +0.191 & 0.080 & [+0.151, +0.228] & 15/15 & $<0.001$ \\
Prompt-withheld & $\Delta_{\mathrm{full}}$ & $R$ to $R{+}S{+}C{+}Z$ & +0.217 & 0.082 & [+0.177, +0.257] & 15/15 & $<0.001$ \\
\addlinespace
Paired condition difference & $q\Delta_S-\Delta_S$ & Prompt-visible minus prompt-withheld summary increment & $-$0.137 & 0.086 & [$-$0.177, $-$0.095] & 13/15 & $<0.001$ \\
Paired condition difference & $q\Delta_C-\Delta_C$ & Prompt-visible minus prompt-withheld trace increment & +0.006 & 0.015 & [$-$0.002, +0.013] & 2/15 & 0.920 \\
Paired condition difference & $q\Delta_Z-\Delta_Z$ & Prompt-visible minus prompt-withheld replay increment & +0.003 & 0.011 & [$-$0.002, +0.009] & 6/15 & 0.855 \\
Paired condition difference & $q\Delta_{\mathrm{public}}-\Delta_{\mathrm{public}}$ & Prompt-visible minus prompt-withheld public gain & $-$0.131 & 0.084 & [$-$0.170, $-$0.089] & 13/15 & $<0.001$ \\
Paired condition difference & $q\Delta_{\mathrm{full}}-\Delta_{\mathrm{full}}$ & Prompt-visible minus prompt-withheld full gain & $-$0.128 & 0.082 & [$-$0.166, $-$0.087] & 13/15 & $<0.001$ \\
\bottomrule
\end{tabular}
\end{adjustbox}
\end{table}

\begin{table}[H]
\centering
\caption{\textbf{The summary adds much less when the prompt is visible.} Values are mean paired AUROC changes across the 15 model--benchmark settings and five data splits ($\pm$ SD across the 75 combinations). Prompt-visible quantities include both orders in which sources are added and the direct $q{+}R{+}C$ versus $q{+}R{+}S$ comparison. The provider-only comparison adds the narrow prompt-conditioned last-token replay features $Z$. The prompt-withheld column gives the corresponding output-only comparison. Brackets are descriptive 95\% setting-bootstrap intervals after averaging the five data splits within each setting.}
\label{tab:prompt_ladder_contrasts}
\small
\begin{adjustbox}{max width=\textwidth}
\begin{tabular}{@{}L{4.0cm}R{3.0cm}R{3.0cm}R{3.2cm}@{}}
\toprule
Quantity & Prompt-visible & Prompt-withheld & Difference \\
\midrule
Summary added first & +0.019 $\pm$ 0.022 & +0.156 $\pm$ 0.099 & \shortstack[r]{$-$0.137\\\mbox{[$-$0.177, $-$0.095]}} \\
Trace added after summary & +0.041 $\pm$ 0.031 & +0.035 $\pm$ 0.038 & \shortstack[r]{+0.006\\\mbox{[$-$0.002, +0.013]}} \\
Trace added first & \shortstack[r]{+0.062 $\pm$ 0.036\\\mbox{[+0.046, +0.077]}} & \textemdash & \textemdash \\
Summary added after trace & \shortstack[r]{$-$0.002 $\pm$ 0.020\\\mbox{[$-$0.007, +0.002]}} & \textemdash & \textemdash \\
Trace minus summary & \shortstack[r]{+0.043 $\pm$ 0.026\\\mbox{[+0.033, +0.052]}} & \textemdash & \textemdash \\
\addlinespace
Provider-only replay features & +0.029 $\pm$ 0.035 & +0.026 $\pm$ 0.040 & \shortstack[r]{+0.003\\\mbox{[$-$0.002, +0.009]}} \\
Public total & +0.060 $\pm$ 0.043 & +0.191 $\pm$ 0.100 & \shortstack[r]{$-$0.131\\\mbox{[$-$0.170, $-$0.089]}} \\
Full total & +0.089 $\pm$ 0.054 & +0.217 $\pm$ 0.094 & \shortstack[r]{$-$0.128\\\mbox{[$-$0.166, $-$0.087]}} \\
\bottomrule
\end{tabular}
\end{adjustbox}
\end{table}

\begin{table}[H]
\centering
\caption{\textbf{Component-block monitors separate access from concatenation compression.} The primary public monitor embeds each rendered display once. The component-block monitor instead concatenates separately computed surface and Cohere feature blocks for the displayed components, so higher-access levels retain the lower-level representation unchanged. Values are mean AUROC across model--benchmark settings after averaging over five data splits ($\pm$ between-setting SD). $\Delta$ is component-block minus primary-display AUROC.}
\label{tab:component_blocks}
\small
\begin{adjustbox}{max width=\textwidth}
\begin{tabular}{@{}L{2.2cm}R{2.5cm}R{2.7cm}R{2.1cm}R{1.3cm}@{}}
\toprule
Display & Primary display & Component blocks & $\Delta$ & Settings \\
\midrule
$R$ & 0.623 $\pm$ 0.112 & 0.624 $\pm$ 0.112 & +0.001 $\pm$ 0.002 & 15 \\
$R{+}S$ & 0.779 $\pm$ 0.031 & 0.775 $\pm$ 0.059 & $-$0.004 $\pm$ 0.037 & 15 \\
$R{+}S{+}C$ & 0.814 $\pm$ 0.048 & 0.810 $\pm$ 0.058 & $-$0.004 $\pm$ 0.019 & 15 \\
$q$ & 0.734 $\pm$ 0.032 & 0.735 $\pm$ 0.032 & +0.001 $\pm$ 0.001 & 15 \\
$q{+}R$ & 0.750 $\pm$ 0.042 & 0.748 $\pm$ 0.051 & $-$0.001 $\pm$ 0.017 & 15 \\
$q{+}R{+}S$ & 0.769 $\pm$ 0.038 & 0.784 $\pm$ 0.049 & +0.016 $\pm$ 0.021 & 15 \\
$q{+}C$ & 0.812 $\pm$ 0.045 & 0.815 $\pm$ 0.051 & +0.003 $\pm$ 0.013 & 15 \\
$q{+}R{+}S{+}C$ & 0.809 $\pm$ 0.050 & 0.812 $\pm$ 0.056 & +0.002 $\pm$ 0.015 & 15 \\
\bottomrule
\end{tabular}
\end{adjustbox}
\end{table}

\begin{table}[H]
\centering
\caption{\textbf{Summary--trace and prompt-visible conclusions persist across the observed design.} Mean paired AUROC contrasts shown separately by benchmark, recomputed after excluding each model family or benchmark, and under item-count weighting instead of equal weighting. The excess ratio is $(\mathrm{AUROC}(S)-0.5)/(\mathrm{AUROC}(C)-0.5)$; it compares excess AUROC rather than retained information. Its paired setting-bootstrap estimate is 0.874 (95\% descriptive interval [0.855, 0.896]; 2,000 draws over the 15 observed settings).}
\label{tab:robustness_loo}
\small
\begin{adjustbox}{max width=\textwidth}
\begin{tabular}{@{}L{3.8cm}R{1.75cm}R{1.75cm}R{1.75cm}R{1.75cm}R{1.75cm}R{1.75cm}R{1.75cm}R{1.75cm}@{}}
\toprule
Aggregate & $\Delta_{\mathrm{full}}$ & $C-S$ & $(R{+}C)-(R{+}S)$ & $q\Delta_S$ & $q\Delta_C$ & $(q{+}R{+}C)-(q{+}R{+}S)$ & $q\Delta_{\mathrm{public}}$ & Excess ratio \\
\midrule
All settings (equal weight) & +0.217 & +0.039 & +0.033 & +0.019 & +0.041 & +0.043 & +0.060 & 0.874 \\
Benchmark only GPQA-Diamond & +0.230 & +0.026 & +0.006 & +0.003 & +0.011 & +0.021 & +0.014 & 0.905 \\
Benchmark only MMLU-Pro & +0.295 & +0.043 & +0.046 & +0.037 & +0.057 & +0.056 & +0.094 & 0.864 \\
Benchmark only Omni-MATH-2-Filtered & +0.127 & +0.049 & +0.047 & +0.018 & +0.054 & +0.052 & +0.071 & 0.860 \\
Excluding family gpt-oss & +0.199 & +0.043 & +0.041 & +0.017 & +0.041 & +0.044 & +0.059 & 0.868 \\
Excluding family Qwen3 & +0.245 & +0.033 & +0.021 & +0.021 & +0.040 & +0.042 & +0.061 & 0.885 \\
Excluding benchmark GPQA-Diamond & +0.211 & +0.046 & +0.046 & +0.027 & +0.056 & +0.054 & +0.083 & 0.862 \\
Excluding benchmark MMLU-Pro & +0.178 & +0.037 & +0.026 & +0.010 & +0.032 & +0.036 & +0.043 & 0.880 \\
Excluding benchmark Omni-MATH-2-Filtered & +0.262 & +0.035 & +0.026 & +0.020 & +0.034 & +0.039 & +0.054 & 0.883 \\
All settings (item weighted) & +0.251 & +0.045 & +0.046 & +0.031 & +0.056 & +0.054 & +0.087 & 0.863 \\
\bottomrule
\end{tabular}
\end{adjustbox}
\end{table}

\begin{table}[H]
\centering
\caption{\textbf{Item-count weighting preserves the headline output-only and prompt-visible increments.} Equal weighting assigns one vote to each of the 15 model--benchmark settings. Item weighting instead weights each setting by its mean held-out test count across the five data splits. The final column is the item-weighted estimate minus the equal-setting estimate.}
\label{tab:item_weighted_increments}
\small
\begin{adjustbox}{max width=\textwidth}
\begin{tabular}{@{}L{3.4cm}R{2.5cm}R{2.5cm}R{2.5cm}@{}}
\toprule
Increment & Equal-weight mean & Item-weighted mean & Change \\
\midrule
$\Delta_S$ & +0.156 & +0.169 & +0.013 \\
$\Delta_C$ & +0.035 & +0.047 & +0.012 \\
$\Delta_Z$ & +0.026 & +0.035 & +0.009 \\
$\Delta_{\mathrm{public}}$ & +0.191 & +0.216 & +0.026 \\
$\Delta_{\mathrm{full}}$ & +0.217 & +0.251 & +0.034 \\
\addlinespace
$q\Delta_S$ & +0.019 & +0.031 & +0.012 \\
$q\Delta_C$ & +0.041 & +0.056 & +0.015 \\
$q\Delta_Z$ & +0.029 & +0.036 & +0.006 \\
$q\Delta_{\mathrm{public}}$ & +0.060 & +0.087 & +0.028 \\
$q\Delta_{\mathrm{full}}$ & +0.089 & +0.123 & +0.034 \\
\bottomrule
\end{tabular}
\end{adjustbox}
\end{table}

\begin{table}[H]
\centering
\caption{\textbf{Mixed-model sensitivity estimates are positive, although the prompt-visible summary interval includes zero.} Intercept-only linear mixed models fit by REML to the 15 per-setting AUROC increments. The preferred specification includes benchmark and model-family variance components; when that fit does not converge with finite fixed-effect uncertainty, the estimate uses the predefined benchmark-only alternative. Estimates and 95\% Wald intervals are descriptive model-based sensitivities for the observed design, not population-generalization intervals. Interval endpoints are reported to four decimals to resolve near-zero bounds. Variance components are reported $\times 10^{4}$; an omitted component is shown as `--'.}
\label{tab:hierarchical_deltas}
\small
\begin{adjustbox}{max width=\textwidth}
\begin{tabular}{@{}L{4.1cm}L{2.1cm}R{1.3cm}R{2.5cm}R{1.3cm}R{1.3cm}R{1.3cm}@{}}
\toprule
Increment & Specification & Estimate & 95\% interval & $\sigma^2_{\mathrm{bench}}$ & $\sigma^2_{\mathrm{family}}$ & $\sigma^2_{\mathrm{resid}}$ \\
\midrule
$q\Delta_S$ ($q{+}R{+}S$ vs.\ $q{+}R$) & Benchmark only & +0.019 & [$-$0.0002, +0.0382] & 2.65 & -- & 1.12 \\
$q\Delta_C$ ($q{+}R{+}S{+}C$ vs.\ $q{+}R{+}S$) & Benchmark + family & +0.041 & [+0.0042, +0.0772] & 9.95 & 0.02 & 1.97 \\
$q\Delta_Z$ ($q{+}R{+}S{+}C{+}Z$ vs.\ $q{+}R{+}S{+}C$) & Benchmark + family & +0.030 & [+0.0184, +0.0415] & 0.41 & 0.22 & 1.45 \\
$q\Delta_{\mathrm{public}}$ ($q{+}R{+}S{+}C$ vs.\ $q{+}R$) & Benchmark only & +0.060 & [+0.0129, +0.1066] & 16.54 & -- & 3.10 \\
$q\Delta_{\mathrm{full}}$ ($q{+}R{+}S{+}C{+}Z$ vs.\ $q{+}R$) & Benchmark + family & +0.090 & [+0.0334, +0.1457] & 23.59 & 0.21 & 3.40 \\
$\Delta_S$ ($R{+}S$ vs.\ $R$) & Benchmark + family & +0.160 & [+0.0467, +0.2743] & 81.58 & 11.49 & 11.16 \\
$\Delta_C$ ($R{+}S{+}C$ vs.\ $R{+}S$) & Benchmark + family & +0.034 & [+0.0042, +0.0635] & 4.40 & 0.91 & 5.27 \\
$\Delta_Z$ ($R{+}S{+}C{+}Z$ vs.\ $R{+}S{+}C$) & Benchmark + family & +0.027 & [+0.0105, +0.0433] & 0.88 & 0.40 & 2.92 \\
$\Delta_{\mathrm{public}}$ ($R{+}S{+}C$ vs.\ $R$) & Benchmark + family & +0.193 & [+0.0976, +0.2876] & 62.24 & 2.89 & 18.34 \\
$\Delta_{\mathrm{full}}$ ($R{+}S{+}C{+}Z$ vs.\ $R$) & Benchmark + family & +0.221 & [+0.1164, +0.3254] & 69.64 & 8.61 & 12.99 \\
\bottomrule
\end{tabular}
\end{adjustbox}
\end{table}

\begin{table}[H]
\centering
\caption{\textbf{Item-clustered bootstrap intervals for the main ladder.} Held-out test AUROC per setting and level (mean over five data splits) with 95\% intervals from resampling test items with replacement within each setting (2{,}000 draws), so the uncertainty respects item-level clustering of the paired design rather than treating runs as independent.}
\label{tab:clustered_bootstrap}
\small
\begin{adjustbox}{max width=\textwidth}
\begin{tabular}{@{}L{2.0cm}L{1.8cm}R{2.9cm}R{2.9cm}R{2.9cm}R{2.9cm}@{}}
\toprule
Benchmark & Model & $R$ & $R+S$ & $R+S+C$ & $R+S+C+Z$ \\
\midrule
GPQA-Diamond & Qwen3-4B & 0.505 [0.408, 0.604] & 0.732 [0.638, 0.817] & 0.732 [0.636, 0.822] & 0.744 [0.651, 0.835] \\
GPQA-Diamond & Qwen3-8B & 0.631 [0.533, 0.730] & 0.812 [0.732, 0.884] & 0.839 [0.758, 0.907] & 0.859 [0.792, 0.920] \\
GPQA-Diamond & Qwen3-14B & 0.558 [0.452, 0.660] & 0.748 [0.655, 0.834] & 0.800 [0.716, 0.874] & 0.767 [0.675, 0.849] \\
GPQA-Diamond & gpt-oss-20b & 0.535 [0.438, 0.637] & 0.755 [0.665, 0.836] & 0.707 [0.612, 0.788] & 0.749 [0.650, 0.835] \\
GPQA-Diamond & gpt-oss-120b & 0.542 [0.440, 0.642] & 0.761 [0.670, 0.843] & 0.770 [0.666, 0.858] & 0.799 [0.709, 0.876] \\
\addlinespace
MMLU-Pro & Qwen3-4B & 0.560 [0.546, 0.574] & 0.764 [0.752, 0.777] & 0.823 [0.813, 0.833] & 0.859 [0.850, 0.868] \\
MMLU-Pro & Qwen3-8B & 0.652 [0.637, 0.667] & 0.774 [0.763, 0.786] & 0.821 [0.810, 0.831] & 0.861 [0.851, 0.870] \\
MMLU-Pro & Qwen3-14B & 0.533 [0.518, 0.549] & 0.775 [0.762, 0.787] & 0.825 [0.814, 0.836] & 0.865 [0.855, 0.875] \\
MMLU-Pro & gpt-oss-20b & 0.515 [0.500, 0.531] & 0.767 [0.754, 0.780] & 0.803 [0.791, 0.814] & 0.844 [0.834, 0.855] \\
MMLU-Pro & gpt-oss-120b & 0.538 [0.521, 0.555] & 0.769 [0.755, 0.782] & 0.810 [0.798, 0.822] & 0.845 [0.834, 0.856] \\
\addlinespace
Omni-MATH-2-Filtered & Qwen3-4B & 0.807 [0.789, 0.825] & 0.803 [0.785, 0.820] & 0.870 [0.855, 0.884] & 0.894 [0.879, 0.907] \\
Omni-MATH-2-Filtered & Qwen3-8B & 0.837 [0.821, 0.853] & 0.853 [0.838, 0.868] & 0.875 [0.861, 0.889] & 0.912 [0.899, 0.924] \\
Omni-MATH-2-Filtered & Qwen3-14B & 0.779 [0.758, 0.799] & 0.821 [0.803, 0.838] & 0.876 [0.861, 0.891] & 0.891 [0.877, 0.904] \\
Omni-MATH-2-Filtered & gpt-oss-20b & 0.682 [0.658, 0.706] & 0.775 [0.754, 0.795] & 0.824 [0.807, 0.841] & 0.864 [0.847, 0.879] \\
Omni-MATH-2-Filtered & gpt-oss-120b & 0.671 [0.645, 0.696] & 0.776 [0.756, 0.796] & 0.832 [0.815, 0.850] & 0.849 [0.833, 0.865] \\
\bottomrule
\end{tabular}
\end{adjustbox}
\end{table}

\begin{table}[H]
\centering
\caption{\textbf{Label permutation removes the monitor signal at every level.} Held-out test AUROC of the primary probes on real labels (mean over per-setting means $\pm$ between-setting SD) versus probes retrained on label-permuted training data (mean $\pm$ SD over all permutation runs across model--benchmark settings and data splits). Permuted AUROC concentrating at chance confirms chance performance after randomizing the training labels; it is not an independent proof against every form of leakage.}
\label{tab:label_permutation}
\small
\begin{adjustbox}{max width=\textwidth}
\begin{tabular}{@{}L{2.6cm}R{3.0cm}R{3.0cm}R{2.4cm}@{}}
\toprule
Level & Real labels & Permuted labels & Permutation runs \\
\midrule
$R$ & 0.623 $\pm$ 0.112 & 0.503 $\pm$ 0.062 & 375 \\
$R+S$ & 0.779 $\pm$ 0.031 & 0.496 $\pm$ 0.057 & 375 \\
$R+S+C$ & 0.814 $\pm$ 0.048 & 0.494 $\pm$ 0.053 & 375 \\
$R+S+C+Z$ & 0.840 $\pm$ 0.052 & 0.496 $\pm$ 0.055 & 375 \\
\bottomrule
\end{tabular}
\end{adjustbox}
\end{table}

\begin{table}[H]
\centering
\caption{\textbf{AUPRC and Brier generally improve with access, whereas uncalibrated ECE does not.} AUROC, AUPRC, Brier score, and expected calibration error (ECE) for the prompt-withheld ladder levels, averaged over per-setting means ($\pm$ between-setting SD; per-setting means average five data splits). Lower is better for Brier and ECE. ECE uses 10 equal-width probability bins on the held-out test predictions and is reported for the raw logistic-regression scores without post-hoc calibration.}
\label{tab:secondary_metrics}
\small
\begin{adjustbox}{max width=\textwidth}
\begin{tabular}{@{}L{2.6cm}L{2.0cm}R{2.1cm}R{2.1cm}R{2.1cm}R{2.1cm}@{}}
\toprule
Benchmark & Level & AUROC & AUPRC & Brier & ECE \\
\midrule
GPQA-Diamond & $R$ & 0.554 $\pm$ 0.047 & 0.688 $\pm$ 0.053 & 0.249 $\pm$ 0.018 & 0.124 $\pm$ 0.018 \\
 & $R+S$ & 0.762 $\pm$ 0.030 & 0.843 $\pm$ 0.038 & 0.226 $\pm$ 0.011 & 0.200 $\pm$ 0.028 \\
 & $R+S+C$ & 0.770 $\pm$ 0.053 & 0.842 $\pm$ 0.051 & 0.222 $\pm$ 0.016 & 0.201 $\pm$ 0.013 \\
 & $R+S+C+Z$ & 0.784 $\pm$ 0.047 & 0.851 $\pm$ 0.057 & 0.209 $\pm$ 0.020 & 0.204 $\pm$ 0.020 \\
\addlinespace
MMLU-Pro & $R$ & 0.560 $\pm$ 0.054 & 0.793 $\pm$ 0.044 & 0.190 $\pm$ 0.019 & 0.064 $\pm$ 0.037 \\
 & $R+S$ & 0.770 $\pm$ 0.005 & 0.907 $\pm$ 0.017 & 0.160 $\pm$ 0.011 & 0.087 $\pm$ 0.009 \\
 & $R+S+C$ & 0.816 $\pm$ 0.010 & 0.931 $\pm$ 0.011 & 0.151 $\pm$ 0.007 & 0.096 $\pm$ 0.019 \\
 & $R+S+C+Z$ & 0.855 $\pm$ 0.010 & 0.946 $\pm$ 0.009 & 0.143 $\pm$ 0.008 & 0.118 $\pm$ 0.038 \\
\addlinespace
Omni-MATH-2-Filtered & $R$ & 0.755 $\pm$ 0.075 & 0.855 $\pm$ 0.028 & 0.181 $\pm$ 0.016 & 0.107 $\pm$ 0.033 \\
 & $R+S$ & 0.806 $\pm$ 0.033 & 0.895 $\pm$ 0.011 & 0.173 $\pm$ 0.010 & 0.115 $\pm$ 0.031 \\
 & $R+S+C$ & 0.855 $\pm$ 0.025 & 0.924 $\pm$ 0.013 & 0.154 $\pm$ 0.012 & 0.116 $\pm$ 0.036 \\
 & $R+S+C+Z$ & 0.882 $\pm$ 0.025 & 0.937 $\pm$ 0.006 & 0.139 $\pm$ 0.018 & 0.112 $\pm$ 0.048 \\
\addlinespace
All benchmarks & $R$ & 0.623 $\pm$ 0.112 & 0.779 $\pm$ 0.082 & 0.207 $\pm$ 0.035 & 0.099 $\pm$ 0.039 \\
 & $R+S$ & 0.779 $\pm$ 0.031 & 0.882 $\pm$ 0.037 & 0.186 $\pm$ 0.031 & 0.134 $\pm$ 0.055 \\
 & $R+S+C$ & 0.814 $\pm$ 0.048 & 0.899 $\pm$ 0.051 & 0.176 $\pm$ 0.036 & 0.138 $\pm$ 0.053 \\
 & $R+S+C+Z$ & 0.840 $\pm$ 0.052 & 0.911 $\pm$ 0.054 & 0.164 $\pm$ 0.037 & 0.145 $\pm$ 0.055 \\
\bottomrule
\end{tabular}
\end{adjustbox}
\end{table}

\begin{table}[H]
\centering
\caption{\textbf{Platt calibration improves the mean probability estimates while leaving aggregate AUROC essentially unchanged.} Calibrators are fit on validation predictions and applied to held-out test predictions for each benchmark, model, data split, access level, and feature family. Each estimate averages 75 combinations of 15 model--benchmark settings and five data splits. AUROC is shown before calibration; Brier score and expected calibration error (ECE) are shown before and after calibration. An individual fitted calibrator can reverse a ranking when its validation slope is negative.}
\label{tab:calibration}
\small
\begin{adjustbox}{max width=\textwidth}
\begin{tabular}{@{}L{2.2cm}R{1.7cm}R{1.7cm}R{1.7cm}R{1.7cm}R{1.7cm}@{}}
\toprule
Level & AUROC & Brier before & Brier after & ECE before & ECE after \\
\midrule
$q{+}R$ & 0.750 & 0.204 & 0.176 & 0.153 & 0.068 \\
$q{+}R{+}S$ & 0.769 & 0.190 & 0.172 & 0.132 & 0.068 \\
$q{+}R{+}S{+}C$ & 0.809 & 0.176 & 0.159 & 0.130 & 0.077 \\
$q{+}R{+}S{+}C{+}Z$ & 0.839 & 0.165 & 0.147 & 0.145 & 0.079 \\
\addlinespace
$R$ & 0.623 & 0.207 & 0.196 & 0.099 & 0.050 \\
$R{+}S$ & 0.779 & 0.186 & 0.171 & 0.134 & 0.077 \\
$R{+}S{+}C$ & 0.814 & 0.176 & 0.158 & 0.138 & 0.073 \\
$R{+}S{+}C{+}Z$ & 0.840 & 0.164 & 0.147 & 0.145 & 0.080 \\
\bottomrule
\end{tabular}
\end{adjustbox}
\end{table}

\begin{table}[H]
\centering
\caption{\textbf{Selected-answer and option-bias baselines do not explain the public monitor signal.} Held-out test AUROC for logistic baselines that see only the selected answer identity, an answer-prior encoding, answer-format indicators, or the selected-answer probability, averaged over five data splits ($\pm$ SD across data splits). The two rightmost columns repeat the public monitor at $R$ and $R{+}S{+}C$ for reference. Verification fields computed against the gold answer encode the label and are excluded from all baseline inputs.}
\label{tab:baseline_controls}
\small
\begin{adjustbox}{max width=\textwidth}
\begin{tabular}{@{}L{2.0cm}L{1.8cm}R{1.7cm}R{1.7cm}R{1.7cm}R{1.7cm}R{1.7cm}R{1.7cm}@{}}
\toprule
Benchmark & Model & Answer prior & Selected answer & Answer format & Selected prob. & $R$ monitor & $R{+}S{+}C$ monitor \\
\midrule
GPQA-Diamond & Qwen3-4B & 0.518 $\pm$ 0.072 & 0.518 $\pm$ 0.072 & 0.510 $\pm$ 0.024 & 0.489 $\pm$ 0.137 & 0.505 $\pm$ 0.066 & 0.732 $\pm$ 0.067 \\
GPQA-Diamond & Qwen3-8B & 0.544 $\pm$ 0.123 & 0.544 $\pm$ 0.123 & 0.589 $\pm$ 0.072 & 0.576 $\pm$ 0.055 & 0.631 $\pm$ 0.047 & 0.839 $\pm$ 0.062 \\
GPQA-Diamond & Qwen3-14B & 0.545 $\pm$ 0.058 & 0.545 $\pm$ 0.058 & 0.516 $\pm$ 0.017 & 0.658 $\pm$ 0.084 & 0.558 $\pm$ 0.046 & 0.800 $\pm$ 0.058 \\
GPQA-Diamond & gpt-oss-20b & 0.526 $\pm$ 0.127 & 0.560 $\pm$ 0.146 & 0.500 $\pm$ 0.000 & 0.625 $\pm$ 0.142 & 0.535 $\pm$ 0.137 & 0.707 $\pm$ 0.096 \\
GPQA-Diamond & gpt-oss-120b & 0.532 $\pm$ 0.063 & 0.538 $\pm$ 0.056 & 0.500 $\pm$ 0.000 & 0.662 $\pm$ 0.109 & 0.542 $\pm$ 0.073 & 0.770 $\pm$ 0.057 \\
\addlinespace
MMLU-Pro & Qwen3-4B & 0.532 $\pm$ 0.018 & 0.531 $\pm$ 0.020 & 0.507 $\pm$ 0.006 & 0.754 $\pm$ 0.011 & 0.560 $\pm$ 0.014 & 0.823 $\pm$ 0.010 \\
MMLU-Pro & Qwen3-8B & 0.523 $\pm$ 0.016 & 0.524 $\pm$ 0.017 & 0.557 $\pm$ 0.014 & 0.688 $\pm$ 0.015 & 0.652 $\pm$ 0.014 & 0.821 $\pm$ 0.004 \\
MMLU-Pro & Qwen3-14B & 0.524 $\pm$ 0.010 & 0.523 $\pm$ 0.012 & 0.508 $\pm$ 0.002 & 0.760 $\pm$ 0.007 & 0.533 $\pm$ 0.004 & 0.825 $\pm$ 0.007 \\
MMLU-Pro & gpt-oss-20b & 0.512 $\pm$ 0.010 & 0.511 $\pm$ 0.011 & 0.504 $\pm$ 0.001 & 0.652 $\pm$ 0.018 & 0.515 $\pm$ 0.009 & 0.803 $\pm$ 0.015 \\
MMLU-Pro & gpt-oss-120b & 0.532 $\pm$ 0.004 & 0.522 $\pm$ 0.016 & 0.501 $\pm$ 0.001 & 0.590 $\pm$ 0.119 & 0.538 $\pm$ 0.006 & 0.810 $\pm$ 0.005 \\
\addlinespace
Omni-MATH-2-Filtered & Qwen3-4B & 0.597 $\pm$ 0.012 & 0.593 $\pm$ 0.009 & 0.762 $\pm$ 0.008 & 0.626 $\pm$ 0.017 & 0.807 $\pm$ 0.015 & 0.870 $\pm$ 0.015 \\
Omni-MATH-2-Filtered & Qwen3-8B & 0.608 $\pm$ 0.023 & 0.607 $\pm$ 0.022 & 0.712 $\pm$ 0.025 & 0.620 $\pm$ 0.012 & 0.837 $\pm$ 0.027 & 0.875 $\pm$ 0.022 \\
Omni-MATH-2-Filtered & Qwen3-14B & 0.610 $\pm$ 0.015 & 0.606 $\pm$ 0.010 & 0.671 $\pm$ 0.021 & 0.639 $\pm$ 0.012 & 0.779 $\pm$ 0.010 & 0.876 $\pm$ 0.008 \\
Omni-MATH-2-Filtered & gpt-oss-20b & 0.563 $\pm$ 0.015 & 0.567 $\pm$ 0.016 & 0.639 $\pm$ 0.027 & 0.604 $\pm$ 0.020 & 0.682 $\pm$ 0.019 & 0.824 $\pm$ 0.022 \\
Omni-MATH-2-Filtered & gpt-oss-120b & 0.568 $\pm$ 0.018 & 0.568 $\pm$ 0.020 & 0.618 $\pm$ 0.013 & 0.607 $\pm$ 0.015 & 0.671 $\pm$ 0.012 & 0.832 $\pm$ 0.019 \\
\bottomrule
\end{tabular}
\end{adjustbox}
\end{table}

\clearpage
% ---- Public and summary signal ----
\begin{table}[H]
\centering
\caption{\textbf{Public display-subset AUROC by model--benchmark setting.} Each entry is the held-out test AUROC of the public monitor (rate surface + Cohere) for the corresponding visible display, averaged over five data splits ($\pm$ SD across data splits). The final row averages the per-setting means across all 15 settings ($\pm$ between-setting SD).}
\label{tab:public_lattice_15cell}
\small
\begin{adjustbox}{max width=\textwidth}
\begin{tabular}{@{}L{2.2cm}L{2.0cm}R{1.75cm}R{1.75cm}R{1.75cm}R{1.75cm}R{1.75cm}R{1.75cm}R{1.75cm}@{}}
\toprule
Benchmark & Model & $R$ & $S$ & $C$ & $R+S$ & $R+C$ & $S+C$ & $R+S+C$ \\
\midrule
GPQA-Diamond & Qwen3-4B & 0.505 $\pm$ 0.066 & 0.735 $\pm$ 0.058 & 0.745 $\pm$ 0.065 & 0.732 $\pm$ 0.063 & 0.738 $\pm$ 0.078 & 0.738 $\pm$ 0.062 & 0.732 $\pm$ 0.067 \\
GPQA-Diamond & Qwen3-8B & 0.631 $\pm$ 0.047 & 0.803 $\pm$ 0.067 & 0.848 $\pm$ 0.058 & 0.812 $\pm$ 0.082 & 0.846 $\pm$ 0.065 & 0.839 $\pm$ 0.065 & 0.839 $\pm$ 0.062 \\
GPQA-Diamond & Qwen3-14B & 0.558 $\pm$ 0.046 & 0.746 $\pm$ 0.084 & 0.781 $\pm$ 0.068 & 0.748 $\pm$ 0.056 & 0.791 $\pm$ 0.072 & 0.787 $\pm$ 0.046 & 0.800 $\pm$ 0.058 \\
GPQA-Diamond & gpt-oss-20b & 0.535 $\pm$ 0.137 & 0.701 $\pm$ 0.122 & 0.725 $\pm$ 0.084 & 0.755 $\pm$ 0.066 & 0.717 $\pm$ 0.095 & 0.708 $\pm$ 0.104 & 0.707 $\pm$ 0.096 \\
GPQA-Diamond & gpt-oss-120b & 0.542 $\pm$ 0.073 & 0.737 $\pm$ 0.062 & 0.752 $\pm$ 0.052 & 0.761 $\pm$ 0.053 & 0.743 $\pm$ 0.062 & 0.760 $\pm$ 0.046 & 0.770 $\pm$ 0.057 \\
\addlinespace
MMLU-Pro & Qwen3-4B & 0.560 $\pm$ 0.014 & 0.771 $\pm$ 0.008 & 0.824 $\pm$ 0.011 & 0.764 $\pm$ 0.009 & 0.824 $\pm$ 0.010 & 0.824 $\pm$ 0.009 & 0.823 $\pm$ 0.010 \\
MMLU-Pro & Qwen3-8B & 0.652 $\pm$ 0.014 & 0.780 $\pm$ 0.010 & 0.831 $\pm$ 0.005 & 0.774 $\pm$ 0.013 & 0.819 $\pm$ 0.005 & 0.829 $\pm$ 0.005 & 0.821 $\pm$ 0.004 \\
\bottomrule
\end{tabular}
\end{adjustbox}
\end{table}

\begin{table}[H]
\centering
\caption{\textbf{Order-independent Shapley attribution allocates the largest mean AUROC contribution to the trace.} Shapley AUROC contributions of the response $R$, self-summary $S$, and reasoning trace $C$ to the full public display, computed over all $3!$ source orderings with an empty-display baseline of 0.5, averaged over five data splits ($\pm$ SD across data splits). The contributions of a setting sum to its full public AUROC minus 0.5. The final row averages per-setting means across all 15 settings ($\pm$ between-setting SD).}
\label{tab:public_shapley}
\small
\begin{adjustbox}{max width=\textwidth}
\begin{tabular}{@{}L{2.2cm}L{2.0cm}R{1.9cm}R{1.9cm}R{1.9cm}R{2.1cm}@{}}
\toprule
Benchmark & Model & $\phi(R)$ & $\phi(S)$ & $\phi(C)$ & $R{+}S{+}C$ AUROC \\
\midrule
GPQA-Diamond & Qwen3-4B & $-$0.002 $\pm$ 0.022 & +0.113 $\pm$ 0.031 & +0.121 $\pm$ 0.053 & 0.732 \\
GPQA-Diamond & Qwen3-8B & +0.045 $\pm$ 0.018 & +0.127 $\pm$ 0.030 & +0.167 $\pm$ 0.020 & 0.839 \\
GPQA-Diamond & Qwen3-14B & +0.026 $\pm$ 0.018 & +0.118 $\pm$ 0.035 & +0.157 $\pm$ 0.047 & 0.800 \\
GPQA-Diamond & gpt-oss-20b & +0.019 $\pm$ 0.064 & +0.097 $\pm$ 0.068 & +0.090 $\pm$ 0.072 & 0.707 \\
GPQA-Diamond & gpt-oss-120b & +0.020 $\pm$ 0.018 & +0.126 $\pm$ 0.030 & +0.125 $\pm$ 0.042 & 0.770 \\
\addlinespace
MMLU-Pro & Qwen3-4B & +0.019 $\pm$ 0.004 & +0.124 $\pm$ 0.005 & +0.181 $\pm$ 0.007 & 0.823 \\
MMLU-Pro & Qwen3-8B & +0.045 $\pm$ 0.005 & +0.114 $\pm$ 0.005 & +0.162 $\pm$ 0.008 & 0.821 \\
MMLU-Pro & Qwen3-14B & +0.010 $\pm$ 0.002 & +0.134 $\pm$ 0.004 & +0.182 $\pm$ 0.007 & 0.825 \\
MMLU-Pro & gpt-oss-20b & +0.004 $\pm$ 0.003 & +0.134 $\pm$ 0.007 & +0.165 $\pm$ 0.007 & 0.803 \\
MMLU-Pro & gpt-oss-120b & +0.012 $\pm$ 0.002 & +0.130 $\pm$ 0.003 & +0.169 $\pm$ 0.004 & 0.810 \\
\addlinespace
Omni-MATH-2-Filtered & Qwen3-4B & +0.107 $\pm$ 0.005 & +0.096 $\pm$ 0.006 & +0.167 $\pm$ 0.006 & 0.870 \\
Omni-MATH-2-Filtered & Qwen3-8B & +0.116 $\pm$ 0.011 & +0.119 $\pm$ 0.009 & +0.140 $\pm$ 0.006 & 0.875 \\
Omni-MATH-2-Filtered & Qwen3-14B & +0.096 $\pm$ 0.006 & +0.116 $\pm$ 0.002 & +0.164 $\pm$ 0.006 & 0.876 \\
Omni-MATH-2-Filtered & gpt-oss-20b & +0.061 $\pm$ 0.007 & +0.108 $\pm$ 0.008 & +0.155 $\pm$ 0.017 & 0.824 \\
Omni-MATH-2-Filtered & gpt-oss-120b & +0.057 $\pm$ 0.004 & +0.111 $\pm$ 0.011 & +0.164 $\pm$ 0.012 & 0.832 \\
\addlinespace
Mean &  & +0.042 $\pm$ 0.038 & +0.118 $\pm$ 0.012 & +0.154 $\pm$ 0.025 & 0.814 $\pm$ 0.048 \\
\bottomrule
\end{tabular}
\end{adjustbox}
\end{table}

\begin{table}[H]
\centering
\caption{\textbf{Mean AUROC is similar under generic and preservation-tuned summary instructions.} Held-out test AUROC of the public monitor for summary-containing displays when the self-summary is generated with the primary preservation instruction (Pres.) versus a generic instruction requesting compression to at most 90 words (Gen.), averaged over five data splits ($\pm$ SD across data splits).}
\label{tab:summary_ablation}
\small
\begin{adjustbox}{max width=\textwidth}
\begin{tabular}{@{}L{2.0cm}L{1.8cm}R{1.6cm}R{1.6cm}R{1.6cm}R{1.6cm}R{1.6cm}R{1.6cm}R{1.6cm}R{1.6cm}@{}}
\toprule
Benchmark & Model & $S$ Pres. & $S$ Gen. & $R+S$ Pres. & $R+S$ Gen. & $S+C$ Pres. & $S+C$ Gen. & $R+S+C$ Pres. & $R+S+C$ Gen. \\
\midrule
GPQA-Diamond & Qwen3-4B & 0.735 $\pm$ 0.058 & 0.693 $\pm$ 0.046 & 0.732 $\pm$ 0.063 & 0.704 $\pm$ 0.047 & 0.738 $\pm$ 0.062 & 0.745 $\pm$ 0.060 & 0.732 $\pm$ 0.067 & 0.741 $\pm$ 0.070 \\
GPQA-Diamond & Qwen3-8B & 0.803 $\pm$ 0.067 & 0.821 $\pm$ 0.072 & 0.812 $\pm$ 0.082 & 0.810 $\pm$ 0.075 & 0.839 $\pm$ 0.065 & 0.839 $\pm$ 0.055 & 0.839 $\pm$ 0.062 & 0.838 $\pm$ 0.059 \\
GPQA-Diamond & Qwen3-14B & 0.746 $\pm$ 0.084 & 0.728 $\pm$ 0.123 & 0.748 $\pm$ 0.056 & 0.761 $\pm$ 0.109 & 0.787 $\pm$ 0.046 & 0.787 $\pm$ 0.073 & 0.800 $\pm$ 0.058 & 0.780 $\pm$ 0.067 \\
GPQA-Diamond & gpt-oss-20b & 0.701 $\pm$ 0.122 & 0.726 $\pm$ 0.098 & 0.755 $\pm$ 0.066 & 0.686 $\pm$ 0.105 & 0.708 $\pm$ 0.104 & 0.718 $\pm$ 0.082 & 0.707 $\pm$ 0.096 & 0.718 $\pm$ 0.086 \\
GPQA-Diamond & gpt-oss-120b & 0.737 $\pm$ 0.062 & 0.765 $\pm$ 0.036 & 0.761 $\pm$ 0.053 & 0.749 $\pm$ 0.038 & 0.760 $\pm$ 0.046 & 0.767 $\pm$ 0.053 & 0.770 $\pm$ 0.057 & 0.766 $\pm$ 0.056 \\
\addlinespace
MMLU-Pro & Qwen3-4B & 0.771 $\pm$ 0.008 & 0.761 $\pm$ 0.013 & 0.764 $\pm$ 0.009 & 0.758 $\pm$ 0.012 & 0.824 $\pm$ 0.009 & 0.822 $\pm$ 0.010 & 0.823 $\pm$ 0.010 & 0.822 $\pm$ 0.011 \\
MMLU-Pro & Qwen3-8B & 0.780 $\pm$ 0.010 & 0.778 $\pm$ 0.014 & 0.774 $\pm$ 0.013 & 0.771 $\pm$ 0.016 & 0.829 $\pm$ 0.005 & 0.830 $\pm$ 0.006 & 0.821 $\pm$ 0.004 & 0.822 $\pm$ 0.006 \\
\bottomrule
\end{tabular}
\end{adjustbox}
\end{table}

\begin{table}[H]
\centering
\caption{\textbf{A fixed primary monitor is stable across three sampled summaries and one greedy summary.} For every setting and access level, regularization is selected on the first data split's training and validation items after excluding the balanced 48-item subset, then the surface-plus-Cohere monitor is refit once on the full complement. AUROC columns report the mean across the three sampled summaries and observed settings; brackets are hierarchical 95\% bootstrap intervals that resample items within the six correctness-by-trace-length strata and resample observed settings. SD, range, and MAPD summarize the three sampled $\Delta_S$ values, where MAPD is the mean absolute pairwise difference. Greedy $-$ sampled compares the greedy $\Delta_S$ with the sampled mean. Equal-stratum AUROC is primary; IPW uses the recorded inverse selection probabilities.}
\label{tab:summary_sampling_sensitivity}
\small
\begin{adjustbox}{max width=\textwidth}
\begin{tabular}{@{}L{2.1cm}L{1.4cm}R{2.2cm}R{2.2cm}R{2.2cm}R{1.0cm}R{1.0cm}R{1.0cm}R{2.2cm}R{1.2cm}@{}}
\toprule
Scope & Weighting & $S$ AUROC & $R{+}S$ AUROC & $\Delta_S$ & SD & Range & MAPD & Greedy $-$ sampled & Settings \\
\midrule
GPQA-Diamond & Equal-stratum & 0.686 [0.596, 0.780] & 0.707 [0.622, 0.786] & +0.077 [$-$0.015, +0.171] & 0.017 & 0.033 & 0.022 & +0.017 [$-$0.001, +0.038] & 5 \\
 & IPW & 0.779 [0.695, 0.859] & 0.797 [0.736, 0.859] & +0.149 [+0.055, +0.248] & 0.015 & 0.027 & 0.018 & +0.015 [$-$0.001, +0.033] & 5 \\
\addlinespace
MMLU-Pro & Equal-stratum & 0.757 [0.670, 0.834] & 0.744 [0.661, 0.827] & +0.153 [+0.052, +0.254] & 0.024 & 0.045 & 0.030 & +0.009 [$-$0.024, +0.044] & 5 \\
 & IPW & 0.787 [0.711, 0.855] & 0.769 [0.695, 0.840] & +0.181 [+0.075, +0.284] & 0.031 & 0.059 & 0.039 & +0.007 [$-$0.028, +0.042] & 5 \\
\addlinespace
Omni-MATH-2-Filtered & Equal-stratum & 0.653 [0.555, 0.763] & 0.675 [0.552, 0.799] & +0.027 [$-$0.039, +0.089] & 0.009 & 0.017 & 0.012 & +0.003 [$-$0.014, +0.020] & 5 \\
 & IPW & 0.770 [0.685, 0.848] & 0.781 [0.674, 0.881] & +0.051 [$-$0.015, +0.117] & 0.009 & 0.018 & 0.012 & +0.001 [$-$0.019, +0.018] & 5 \\
\addlinespace
All settings & Equal-stratum & 0.699 [0.635, 0.756] & 0.708 [0.650, 0.765] & +0.086 [+0.027, +0.146] & 0.017 & 0.032 & 0.021 & +0.010 [$-$0.005, +0.024] & 15 \\
 & IPW & 0.779 [0.732, 0.820] & 0.782 [0.731, 0.830] & +0.127 [+0.066, +0.187] & 0.018 & 0.034 & 0.023 & +0.008 [$-$0.008, +0.023] & 15 \\
\bottomrule
\end{tabular}
\end{adjustbox}
\end{table}

\begin{table}[H]
\centering
\caption{\textbf{The exact 90-word cap and conclusion-region controls use the primary surface-plus-Cohere monitor.} Each AUROC entry is the mean $\pm$ between-setting SD of the setting-level held-out test AUROC, where each setting mean averages the five prespecified data splits. $\Delta$ is paired within setting against the original uncapped preservation summary. Cap90 deterministically truncates that summary to at most 90 whitespace-delimited words; Last=$n_S$ uses up to the final $n_S$ trace words, where $n_S$ is the capped summary length; Last90 uses up to the final 90 trace words and therefore the whole trace when it is shorter; and Last90 masked removes answer, option, and prompt-number cues before extraction. The final row reports Last90 minus Pres. for the $S$ display; brackets are descriptive 95\% setting-bootstrap intervals after averaging the five data splits within each of the 15 observed settings.}
\label{tab:summary_primary_controls}
\small
\begin{adjustbox}{max width=\textwidth}
\begin{tabular}{@{}L{2.8cm}R{2.0cm}R{2.0cm}R{2.0cm}R{2.0cm}R{0.9cm}@{}}
\toprule
Summary slot & $S$ AUROC & $S$ $\Delta$ & $R{+}S$ AUROC & $R{+}S$ $\Delta$ & $n$ \\
\midrule
Pres. (uncapped) & 0.774 $\pm$ 0.035 & 0.000 $\pm$ 0.000 & 0.779 $\pm$ 0.031 & 0.000 $\pm$ 0.000 & 15 \\
Cap90 & 0.775 $\pm$ 0.026 & +0.001 $\pm$ 0.018 & 0.774 $\pm$ 0.034 & $-$0.005 $\pm$ 0.007 & 15 \\
Last=$n_S$ & 0.784 $\pm$ 0.028 & +0.010 $\pm$ 0.017 & 0.779 $\pm$ 0.029 & 0.000 $\pm$ 0.012 & 15 \\
Last90 & 0.780 $\pm$ 0.032 & +0.006 $\pm$ 0.019 & 0.778 $\pm$ 0.033 & $-$0.001 $\pm$ 0.014 & 15 \\
Last90 masked & 0.782 $\pm$ 0.031 & +0.009 $\pm$ 0.022 & 0.780 $\pm$ 0.032 & +0.001 $\pm$ 0.017 & 15 \\
\addlinespace
Last90 $-$ Pres. & \textemdash & \shortstack[r]{+0.006\\\mbox{[$-$0.003, +0.016]}} & \textemdash & \textemdash & 15 \\
\bottomrule
\end{tabular}
\end{adjustbox}
\end{table}

\begin{table}[H]
\centering
\caption{\textbf{Length-matched extractive controls compare summaries with trace extracts at the same word count.} Held-out test AUROC of the open surface-plus-TF-IDF monitor for the summary-only display. Pres. uses the original preservation-tuned self-summary; Cap90 hard-truncates that summary to at most 90 words; First=nS, Last=nS, Head+Tail=nS, and Random=nS use trace words matched to the capped summary length $n_S$; Last90 uses up to the final 90 trace words and therefore the whole trace when it is shorter; Last90 masked applies answer, option, and prompt-number masking before the same extraction. Values average five data splits ($\pm$ SD across data splits). The matched mean averages settings where every displayed variant is available. The two contrasts report Last=$n_S$ minus Cap90 and Last90 minus Pres.; brackets are descriptive 95\% setting-bootstrap intervals after averaging the five data splits within each of the 15 matched settings.}
\label{tab:summary_extractive}
\small
\begin{adjustbox}{max width=\textwidth}
\begin{tabular}{@{}L{2.0cm}L{1.8cm}R{1.5cm}R{1.5cm}R{1.5cm}R{1.5cm}R{1.5cm}R{1.5cm}R{1.5cm}R{1.5cm}@{}}
\toprule
Benchmark & Model & $S$ Pres. & $S$ Cap90 & $S$ First=nS & $S$ Last=nS & $S$ Head+Tail=nS & $S$ Random=nS & $S$ Last90 & $S$ Last90 masked \\
\midrule
GPQA-Diamond & Qwen3-4B & 0.663 $\pm$ 0.035 & 0.659 $\pm$ 0.034 & 0.644 $\pm$ 0.066 & 0.716 $\pm$ 0.058 & 0.646 $\pm$ 0.118 & 0.769 $\pm$ 0.092 & 0.741 $\pm$ 0.071 & 0.720 $\pm$ 0.075 \\
GPQA-Diamond & Qwen3-8B & 0.750 $\pm$ 0.047 & 0.769 $\pm$ 0.050 & 0.768 $\pm$ 0.092 & 0.833 $\pm$ 0.073 & 0.829 $\pm$ 0.054 & 0.771 $\pm$ 0.038 & 0.830 $\pm$ 0.080 & 0.842 $\pm$ 0.058 \\
GPQA-Diamond & Qwen3-14B & 0.779 $\pm$ 0.081 & 0.778 $\pm$ 0.073 & 0.720 $\pm$ 0.053 & 0.774 $\pm$ 0.094 & 0.715 $\pm$ 0.065 & 0.820 $\pm$ 0.104 & 0.785 $\pm$ 0.088 & 0.773 $\pm$ 0.084 \\
GPQA-Diamond & gpt-oss-20b & 0.700 $\pm$ 0.080 & 0.697 $\pm$ 0.074 & 0.722 $\pm$ 0.054 & 0.738 $\pm$ 0.114 & 0.713 $\pm$ 0.108 & 0.762 $\pm$ 0.084 & 0.737 $\pm$ 0.117 & 0.742 $\pm$ 0.113 \\
GPQA-Diamond & gpt-oss-120b & 0.736 $\pm$ 0.100 & 0.735 $\pm$ 0.099 & 0.718 $\pm$ 0.090 & 0.807 $\pm$ 0.036 & 0.714 $\pm$ 0.084 & 0.760 $\pm$ 0.093 & 0.767 $\pm$ 0.047 & 0.784 $\pm$ 0.066 \\
\addlinespace
MMLU-Pro & Qwen3-4B & 0.802 $\pm$ 0.006 & 0.802 $\pm$ 0.006 & 0.734 $\pm$ 0.005 & 0.814 $\pm$ 0.009 & 0.796 $\pm$ 0.005 & 0.772 $\pm$ 0.009 & 0.824 $\pm$ 0.010 & 0.826 $\pm$ 0.010 \\
MMLU-Pro & Qwen3-8B & 0.815 $\pm$ 0.009 & 0.803 $\pm$ 0.010 & 0.745 $\pm$ 0.008 & 0.820 $\pm$ 0.006 & 0.815 $\pm$ 0.010 & 0.795 $\pm$ 0.009 & 0.821 $\pm$ 0.007 & 0.823 $\pm$ 0.006 \\
MMLU-Pro & Qwen3-14B & 0.813 $\pm$ 0.006 & 0.811 $\pm$ 0.007 & 0.745 $\pm$ 0.005 & 0.817 $\pm$ 0.007 & 0.805 $\pm$ 0.010 & 0.777 $\pm$ 0.009 & 0.819 $\pm$ 0.004 & 0.821 $\pm$ 0.005 \\
MMLU-Pro & gpt-oss-20b & 0.788 $\pm$ 0.014 & 0.784 $\pm$ 0.015 & 0.749 $\pm$ 0.014 & 0.798 $\pm$ 0.013 & 0.790 $\pm$ 0.012 & 0.774 $\pm$ 0.010 & 0.801 $\pm$ 0.011 & 0.805 $\pm$ 0.010 \\
MMLU-Pro & gpt-oss-120b & 0.802 $\pm$ 0.008 & 0.801 $\pm$ 0.009 & 0.756 $\pm$ 0.005 & 0.807 $\pm$ 0.009 & 0.789 $\pm$ 0.004 & 0.788 $\pm$ 0.005 & 0.812 $\pm$ 0.009 & 0.814 $\pm$ 0.009 \\
\addlinespace
Omni-MATH-2-Filtered & Qwen3-4B & 0.813 $\pm$ 0.009 & 0.799 $\pm$ 0.009 & 0.748 $\pm$ 0.013 & 0.845 $\pm$ 0.013 & 0.839 $\pm$ 0.014 & 0.767 $\pm$ 0.011 & 0.851 $\pm$ 0.006 & 0.838 $\pm$ 0.016 \\
Omni-MATH-2-Filtered & Qwen3-8B & 0.849 $\pm$ 0.020 & 0.834 $\pm$ 0.018 & 0.806 $\pm$ 0.025 & 0.874 $\pm$ 0.014 & 0.868 $\pm$ 0.011 & 0.810 $\pm$ 0.020 & 0.866 $\pm$ 0.014 & 0.871 $\pm$ 0.013 \\
Omni-MATH-2-Filtered & Qwen3-14B & 0.845 $\pm$ 0.006 & 0.832 $\pm$ 0.006 & 0.771 $\pm$ 0.007 & 0.853 $\pm$ 0.017 & 0.845 $\pm$ 0.015 & 0.773 $\pm$ 0.007 & 0.840 $\pm$ 0.014 & 0.849 $\pm$ 0.019 \\
Omni-MATH-2-Filtered & gpt-oss-20b & 0.784 $\pm$ 0.005 & 0.785 $\pm$ 0.008 & 0.727 $\pm$ 0.018 & 0.773 $\pm$ 0.016 & 0.766 $\pm$ 0.021 & 0.755 $\pm$ 0.025 & 0.780 $\pm$ 0.012 & 0.767 $\pm$ 0.014 \\
Omni-MATH-2-Filtered & gpt-oss-120b & 0.795 $\pm$ 0.021 & 0.789 $\pm$ 0.021 & 0.743 $\pm$ 0.021 & 0.786 $\pm$ 0.015 & 0.789 $\pm$ 0.014 & 0.768 $\pm$ 0.016 & 0.775 $\pm$ 0.021 & 0.772 $\pm$ 0.014 \\
\addlinespace
Matched mean (15 settings) &  & 0.782 & 0.778 & 0.740 & 0.804 & 0.781 & 0.777 & 0.803 & 0.803 \\
\addlinespace
\multicolumn{2}{l}{Last=$n_S$ $-$ Cap90} & \multicolumn{8}{r}{\shortstack[r]{+0.025\\\mbox{[+0.013, +0.038]}}} \\
\multicolumn{2}{l}{Last90 $-$ Pres.} & \multicolumn{8}{r}{\shortstack[r]{+0.021\\\mbox{[+0.008, +0.036]}}} \\
\bottomrule
\end{tabular}
\end{adjustbox}
\end{table}

\begin{table}[H]
\centering
\caption{\textbf{Trace-native chunk monitors are compared directly with the matched summary and whole-trace monitors.} AUROC values are means $\pm$ between-setting SD after first averaging the five prespecified data splits within each of the 15 model--benchmark settings. Deltas are paired by setting against the primary whole-display summary ($S$) and trace ($C$) monitors. Each full trace is divided into balanced contiguous word chunks targeting 512 words and capped at eight chunks. The Cohere variants pool fixed 512-dimensional chunk vectors or fit a linear-attention multiple-instance head. Optional local TF--IDF variants are labelled separately as weak-label alternatives.}
\label{tab:chunked_trace}
\small
\begin{adjustbox}{max width=\textwidth}
\begin{tabular}{@{}L{4.0cm}R{2.2cm}R{2.2cm}R{2.2cm}R{1.0cm}@{}}
\toprule
Monitor & AUROC & $\Delta$ vs $S$ & $\Delta$ vs $C$ & Settings \\
\midrule
Primary whole-display $S$ & 0.774 $\pm$ 0.035 & +0.000 $\pm$ 0.000 & $-$0.039 $\pm$ 0.017 & 15 \\
Primary whole-display $C$ & 0.813 $\pm$ 0.045 & +0.039 $\pm$ 0.017 & +0.000 $\pm$ 0.000 & 15 \\
Cohere chunks: mean pooling & 0.812 $\pm$ 0.048 & +0.038 $\pm$ 0.020 & $-$0.001 $\pm$ 0.008 & 15 \\
Cohere chunks: elementwise max & 0.811 $\pm$ 0.044 & +0.038 $\pm$ 0.019 & $-$0.002 $\pm$ 0.010 & 15 \\
Cohere chunks: elementwise top-2 & 0.813 $\pm$ 0.043 & +0.040 $\pm$ 0.018 & +0.000 $\pm$ 0.011 & 15 \\
Cohere chunks: attention MIL & 0.792 $\pm$ 0.049 & +0.018 $\pm$ 0.021 & $-$0.021 $\pm$ 0.013 & 15 \\
\bottomrule
\end{tabular}
\end{adjustbox}
\end{table}

\begin{table}[H]
\centering
\caption{\textbf{Scores from trace segments are compared with scores from complete summaries and traces on the full dataset.} AUROC values first average the five prespecified random splits within each model--benchmark combination and then report the mean $\pm$ SD across all 15 combinations. Deltas compare values within the same combination with the complete summary and trace, using the same surface features and TF--IDF features. The segment and complete-text classifiers use the same six candidate inverse-regularization values, $C\in\{0.0001,0.001,0.01,0.1,1,10\}$. Each saved trace is divided into up to eight contiguous segments of similar length, targeting 512 words. The segment classifier is trained on all items in the saved split rather than the earlier 500-item subset based on one random split. Scores are combined only for held-out items. Within each combination and random split, $C$ is chosen once according to validation item-level AUROC after taking the maximum segment score for each item. The top-two mean, mean, and last-segment results use that same value of $C$. These alternatives test sensitivity to how traces are divided and segment scores are combined; they do not locate causal evidence. A short trace may form one segment; during fitting, each item is weighted once for every segment it contains; and maximum or top-two aggregation gives longer traces more opportunities to receive a high score.}
\label{tab:chunked_trace_full_data_open}
\small
\begin{adjustbox}{max width=\textwidth}
\begin{tabular}{@{}L{4.2cm}R{2.2cm}R{2.4cm}R{2.4cm}R{1.2cm}@{}}
\toprule
Input and score & AUROC & $\Delta$ vs $S$ & $\Delta$ vs $C$ & Settings \\
\midrule
Whole-display summary $S$ & 0.782 $\pm$ 0.051 & +0.000 $\pm$ 0.000 & $-$0.036 $\pm$ 0.021 & 15 \\
Whole-display trace $C$ & 0.818 $\pm$ 0.051 & +0.036 $\pm$ 0.021 & +0.000 $\pm$ 0.000 & 15 \\
Segment scores: maximum & 0.799 $\pm$ 0.051 & +0.017 $\pm$ 0.029 & $-$0.019 $\pm$ 0.016 & 15 \\
Segment scores: top-two mean & 0.803 $\pm$ 0.051 & +0.021 $\pm$ 0.024 & $-$0.015 $\pm$ 0.013 & 15 \\
Segment scores: mean & 0.818 $\pm$ 0.046 & +0.036 $\pm$ 0.026 & +0.000 $\pm$ 0.012 & 15 \\
Last (or only) segment & 0.817 $\pm$ 0.043 & +0.035 $\pm$ 0.026 & $-$0.000 $\pm$ 0.015 & 15 \\
\bottomrule
\end{tabular}
\end{adjustbox}
\end{table}

\begin{table}[H]
\centering
\caption{\textbf{Public-text feature comparisons.} AUROC values are mean $\pm$ descriptive SD across the model--benchmark settings and data splits available at each public access level. These SDs are not confidence intervals. All displayed feature families cover the 15 model--benchmark settings. The final column gives the number of model--benchmark--split estimates.}
\label{tab:public_feature_controls}
\small
\begin{adjustbox}{max width=\textwidth}
\begin{tabular}{@{}L{2.2cm}L{4.0cm}R{2.0cm}R{1.2cm}@{}}
\toprule
Access level & Feature family & AUROC & Estimates \\
\midrule
$R$ & Surface & 0.584 $\pm$ 0.098 & 75 \\
$R$ & No length & 0.579 $\pm$ 0.092 & 75 \\
$R$ & Cohere & 0.620 $\pm$ 0.113 & 75 \\
$R$ & Surface + Cohere & 0.623 $\pm$ 0.117 & 75 \\
$R$ & TF-IDF & 0.618 $\pm$ 0.121 & 75 \\
$R$ & Surface + TF-IDF + Cohere & 0.623 $\pm$ 0.117 & 75 \\
\addlinespace
$R+S$ & Surface & 0.722 $\pm$ 0.054 & 75 \\
$R+S$ & No length & 0.701 $\pm$ 0.052 & 75 \\
$R+S$ & Cohere & 0.773 $\pm$ 0.046 & 75 \\
$R+S$ & Surface + Cohere & 0.779 $\pm$ 0.047 & 75 \\
$R+S$ & TF-IDF & 0.797 $\pm$ 0.050 & 75 \\
$R+S$ & Surface + TF-IDF + Cohere & 0.784 $\pm$ 0.047 & 75 \\
\addlinespace
$R+S+C$ & Surface & 0.803 $\pm$ 0.050 & 75 \\
$R+S+C$ & No length & 0.777 $\pm$ 0.043 & 75 \\
$R+S+C$ & Cohere & 0.802 $\pm$ 0.059 & 75 \\
$R+S+C$ & Surface + Cohere & 0.814 $\pm$ 0.060 & 75 \\
$R+S+C$ & TF-IDF & 0.815 $\pm$ 0.048 & 75 \\
$R+S+C$ & Surface + TF-IDF + Cohere & 0.814 $\pm$ 0.060 & 75 \\
\bottomrule
\end{tabular}
\end{adjustbox}
\end{table}

\begin{table}[H]
\centering
\caption{\textbf{Correctness ranking remains after adjusting the other features for text length.} For every setting, access level, and data split, word and sentence counts are nuisance covariates. Inside each fitting fold, multi-output least squares residualizes every other primary surface and Cohere feature on those two counts; the counts are then excluded before standardization and logistic-probe fitting. Values average the five data splits within each setting and then give equal weight to the 15 settings. Change is the adjusted estimate minus the original primary-monitor estimate.}
\label{tab:length_orthogonalized}
\small
\begin{adjustbox}{max width=\textwidth}
\begin{tabular}{@{}L{4.0cm}R{2.4cm}R{2.8cm}R{2.4cm}@{}}
\toprule
Quantity & Primary monitor & After length adjustment & Change \\
\midrule
\multicolumn{4}{l}{\textbf{Access-level AUROC}} \\
$R$ & 0.623 & 0.617 & $-$0.006 \\
$S$ & 0.774 & 0.689 & $-$0.084 \\
$C$ & 0.813 & 0.716 & $-$0.097 \\
$R{+}S$ & 0.779 & 0.734 & $-$0.045 \\
$R{+}S{+}C$ & 0.814 & 0.717 & $-$0.097 \\
$q{+}R$ & 0.750 & 0.727 & $-$0.023 \\
$q{+}R{+}S$ & 0.769 & 0.736 & $-$0.033 \\
$q{+}R{+}C$ & 0.812 & 0.712 & $-$0.100 \\
$q{+}R{+}S{+}C$ & 0.809 & 0.713 & $-$0.097 \\
\addlinespace
\multicolumn{4}{l}{\textbf{Paired AUROC increments}} \\
$\Delta_S$ & 0.156 & 0.117 & $-$0.039 \\
$\Delta_C$ & 0.035 & $-$0.017 & $-$0.052 \\
$\Delta_{\mathrm{public}}$ & 0.191 & 0.101 & $-$0.090 \\
$q\Delta_S$ & 0.019 & 0.010 & $-$0.009 \\
$q\Delta_{C\mid R}$ & 0.062 & $-$0.015 & $-$0.077 \\
$q\Delta_C$ & 0.041 & $-$0.023 & $-$0.064 \\
$q\Delta_{S\mid R,C}$ & $-$0.002 & 0.001 & +0.003 \\
$q(C{-}S)$ & 0.043 & $-$0.024 & $-$0.067 \\
$q\Delta_{\mathrm{public}}$ & 0.060 & $-$0.014 & $-$0.073 \\
\bottomrule
\end{tabular}
\end{adjustbox}
\end{table}

\begin{table}[H]
\centering
\caption{\textbf{Public embeddings fit without silent truncation.} Cohere embed-v4.0 was applied with silent truncation disabled. Counts refer to items at the full public level $R+S+C$, pooled across the five target models for each benchmark.}
\label{tab:embedding_qc}
\small
\begin{adjustbox}{max width=\textwidth}
\begin{tabular}{@{}L{3.0cm}R{2.2cm}R{2.2cm}R{1.8cm}R{1.8cm}@{}}
\toprule
Benchmark & $R+S+C$ items & Whole-context fit & Mean chunks & Max chunks \\
\midrule
GPQA-Diamond & 990 & 100\% & 1.00 & 1 \\
MMLU-Pro & 60,160 & 100\% & 1.00 & 1 \\
Omni-MATH-2-Filtered & 20,905 & 100\% & 1.00 & 1 \\
\bottomrule
\end{tabular}
\end{adjustbox}
\end{table}

\begin{table}[H]
\centering
\caption{\textbf{Stripping prompt-overlapping text from traces and summaries leaves public AUROC essentially unchanged.} Paired AUROC change when 8--14-word spans copied from the task prompt are removed from the visible summary and trace before feature extraction, averaged over five data splits ($\pm$ SD across data splits). The final row averages per-setting means across all 15 settings ($\pm$ between-setting SD).}
\label{tab:noqcopy_controls}
\small
\begin{adjustbox}{max width=\textwidth}
\begin{tabular}{@{}L{2.2cm}L{2.0cm}R{1.85cm}R{1.85cm}R{1.85cm}R{1.85cm}R{1.85cm}R{1.85cm}@{}}
\toprule
Benchmark & Model & $S$ & $C$ & $R{+}S$ & $R{+}C$ & $S{+}C$ & $R{+}S{+}C$ \\
\midrule
GPQA-Diamond & Qwen3-4B & +0.006 $\pm$ 0.020 & $-$0.004 $\pm$ 0.019 & +0.016 $\pm$ 0.014 & $-$0.008 $\pm$ 0.019 & $-$0.005 $\pm$ 0.012 & $-$0.003 $\pm$ 0.021 \\
GPQA-Diamond & Qwen3-8B & $-$0.003 $\pm$ 0.010 & +0.007 $\pm$ 0.012 & $-$0.006 $\pm$ 0.010 & +0.004 $\pm$ 0.005 & +0.003 $\pm$ 0.005 & +0.000 $\pm$ 0.009 \\
GPQA-Diamond & Qwen3-14B & +0.019 $\pm$ 0.046 & +0.014 $\pm$ 0.030 & $-$0.013 $\pm$ 0.019 & $-$0.019 $\pm$ 0.033 & $-$0.014 $\pm$ 0.053 & $-$0.021 $\pm$ 0.037 \\
GPQA-Diamond & gpt-oss-20b & $-$0.027 $\pm$ 0.067 & $-$0.004 $\pm$ 0.029 & $-$0.030 $\pm$ 0.080 & +0.007 $\pm$ 0.042 & +0.002 $\pm$ 0.014 & $-$0.001 $\pm$ 0.020 \\
GPQA-Diamond & gpt-oss-120b & +0.005 $\pm$ 0.016 & $-$0.002 $\pm$ 0.020 & +0.004 $\pm$ 0.007 & $-$0.002 $\pm$ 0.032 & +0.005 $\pm$ 0.020 & $-$0.007 $\pm$ 0.019 \\
\addlinespace
MMLU-Pro & Qwen3-4B & +0.000 $\pm$ 0.001 & +0.000 $\pm$ 0.003 & +0.000 $\pm$ 0.000 & +0.000 $\pm$ 0.003 & $-$0.000 $\pm$ 0.002 & $-$0.000 $\pm$ 0.002 \\
MMLU-Pro & Qwen3-8B & $-$0.000 $\pm$ 0.001 & +0.001 $\pm$ 0.001 & +0.000 $\pm$ 0.000 & +0.001 $\pm$ 0.001 & $-$0.000 $\pm$ 0.001 & $-$0.000 $\pm$ 0.001 \\
\bottomrule
\end{tabular}
\end{adjustbox}
\end{table}

\input{tables/table_masked_controls}

\clearpage
% ---- Prompt-conditioned and within-item analyses ----
\input{tables/table_prompt_controls_15cell}

\input{tables/table_prompt_residual_contrasts}

\input{tables/table_prompt_conditioned_shapley}

\input{tables/table_open_prompt_factorial}

\begin{table}[H]
\centering
\caption{\textbf{The complete open factorial shows context-dependent marginal gains.} Mean paired AUROC changes across the 15 settings for the surface-feature-plus-TF-IDF monitor. Intervals are descriptive 95\% setting-bootstrap intervals over the observed settings, not population-generalization intervals; the last column counts settings with a positive mean contrast. Summary-first and trace-first marginal gains use the same held-out items and data splits. Here, `after' denotes set inclusion in the factorial contrast, not a change in the physical presentation order of the displayed text.}
\label{tab:open_prompt_contrasts}
\small
\begin{adjustbox}{max width=\textwidth}
\begin{tabular}{@{}L{6.2cm}R{2.0cm}R{3.0cm}R{2.0cm}@{}}
\toprule
Contrast & Mean $\Delta$ AUROC & 95\% descriptive interval & Positive settings \\
\midrule
Add $S$ after $q{+}R$ & +0.047 & [+0.027, +0.064] & 14/15 \\
Add $C$ after $q{+}R{+}S$ & +0.048 & [+0.039, +0.059] & 15/15 \\
Add $C$ after $q{+}R$ & +0.094 & [+0.078, +0.110] & 15/15 \\
Add $S$ after $q{+}R{+}C$ & +0.000 & [$-$0.001, +0.001] & 12/15 \\
$q{+}R{+}C$ minus $q{+}R{+}S$ & +0.047 & [+0.038, +0.058] & 15/15 \\
$q{+}R{+}S{+}C$ minus $q{+}R$ & +0.094 & [+0.079, +0.110] & 15/15 \\
\bottomrule
\end{tabular}
\end{adjustbox}
\end{table}

\input{tables/table_open_prompt_shapley}

\input{tables/table_multirun_within_item}

\input{tables/table_multirun_prompt_visible}

\input{tables/table_within_item_length_controls}

\begin{table}[H]
\centering
\caption{\textbf{A conclusion-region trace extract is compared with the model-written summary on repeated runs.} Entries give across-item AUROC / within-item AUROC; square brackets give the 95\% item-bootstrap interval for within-item AUROC. Self-summary uses the original preservation-tuned summary, whereas Last90 deterministically places up to the final 90 whitespace-delimited trace words in the summary slot and therefore uses the entire trace when it is shorter. It is a conclusion-region-biased control rather than a pure ending-only test. Monitors use the same primary surface-plus-Cohere feature family, are trained on run 1, and score held-out MMLU-Pro items across three independent runs. Both AUROCs use the same post-outcome discordant items. Across-item AUROC pools runs across those items, while within-item AUROC compares correct and incorrect runs of one item.}
\label{tab:multirun_last90}
\small
\begin{adjustbox}{max width=\textwidth}
\begin{tabular}{@{}L{2.4cm}R{2.5cm}R{2.5cm}R{2.5cm}R{2.5cm}@{}}
\toprule
Model & $S$ self-summary & $S$ Last90 & $R{+}S$ self-summary & $R{+}S$ Last90 \\
\midrule
Qwen3-4B & \shortstack[r]{0.536 / 0.533\\{}[0.507, 0.557]} & \shortstack[r]{0.550 / 0.559\\{}[0.534, 0.584]} & \shortstack[r]{0.533 / 0.545\\{}[0.519, 0.569]} & \shortstack[r]{0.541 / 0.543\\{}[0.518, 0.569]} \\
Qwen3-8B & \shortstack[r]{0.536 / 0.531\\{}[0.505, 0.557]} & \shortstack[r]{0.558 / 0.544\\{}[0.519, 0.570]} & \shortstack[r]{0.525 / 0.519\\{}[0.493, 0.545]} & \shortstack[r]{0.528 / 0.542\\{}[0.518, 0.566]} \\
Qwen3-14B & \shortstack[r]{0.515 / 0.503\\{}[0.477, 0.529]} & \shortstack[r]{0.524 / 0.525\\{}[0.498, 0.551]} & \shortstack[r]{0.512 / 0.505\\{}[0.478, 0.530]} & \shortstack[r]{0.525 / 0.522\\{}[0.496, 0.548]} \\
gpt-oss-20b & \shortstack[r]{0.536 / 0.526\\{}[0.504, 0.548]} & \shortstack[r]{0.548 / 0.536\\{}[0.514, 0.558]} & \shortstack[r]{0.533 / 0.515\\{}[0.494, 0.537]} & \shortstack[r]{0.548 / 0.538\\{}[0.516, 0.560]} \\
gpt-oss-120b & \shortstack[r]{0.534 / 0.527\\{}[0.501, 0.553]} & \shortstack[r]{0.556 / 0.565\\{}[0.539, 0.593]} & \shortstack[r]{0.534 / 0.526\\{}[0.499, 0.553]} & \shortstack[r]{0.554 / 0.574\\{}[0.547, 0.600]} \\
\bottomrule
\end{tabular}
\end{adjustbox}
\end{table}

\input{tables/table_open_prompt_multirun}

\begin{table}[H]
\centering
\caption{\textbf{A conclusion-region trace extract is compared with the model-written summary on repeated runs.} Entries give across-item AUROC / within-item AUROC; square brackets give the 95\% item-bootstrap interval for within-item AUROC. Self-summary uses the original preservation-tuned summary, whereas Last90 deterministically places up to the final 90 whitespace-delimited trace words in the summary slot and therefore uses the entire trace when it is shorter. It is a conclusion-region-biased control rather than a pure ending-only test. Monitors use the fully open surface-plus-TF--IDF feature family, are trained on run 1, and score held-out MMLU-Pro items across three independent runs. Both AUROCs use the same post-outcome discordant items. Across-item AUROC pools runs across those items, while within-item AUROC compares correct and incorrect runs of one item.}
\label{tab:multirun_last90_open}
\small
\begin{adjustbox}{max width=\textwidth}
\begin{tabular}{@{}L{2.4cm}R{2.5cm}R{2.5cm}R{2.5cm}R{2.5cm}@{}}
\toprule
Model & $S$ self-summary & $S$ Last90 & $R{+}S$ self-summary & $R{+}S$ Last90 \\
\midrule
Qwen3-4B & \shortstack[r]{0.575 / 0.565\\{}[0.539, 0.590]} & \shortstack[r]{0.597 / 0.592\\{}[0.566, 0.617]} & \shortstack[r]{0.574 / 0.568\\{}[0.542, 0.592]} & \shortstack[r]{0.596 / 0.589\\{}[0.563, 0.614]} \\
Qwen3-8B & \shortstack[r]{0.563 / 0.569\\{}[0.543, 0.595]} & \shortstack[r]{0.580 / 0.554\\{}[0.528, 0.581]} & \shortstack[r]{0.547 / 0.551\\{}[0.524, 0.578]} & \shortstack[r]{0.553 / 0.544\\{}[0.518, 0.571]} \\
Qwen3-14B & \shortstack[r]{0.543 / 0.543\\{}[0.515, 0.569]} & \shortstack[r]{0.563 / 0.562\\{}[0.536, 0.588]} & \shortstack[r]{0.546 / 0.542\\{}[0.513, 0.568]} & \shortstack[r]{0.561 / 0.559\\{}[0.533, 0.585]} \\
gpt-oss-20b & \shortstack[r]{0.561 / 0.565\\{}[0.543, 0.587]} & \shortstack[r]{0.575 / 0.558\\{}[0.536, 0.580]} & \shortstack[r]{0.562 / 0.571\\{}[0.549, 0.593]} & \shortstack[r]{0.576 / 0.558\\{}[0.536, 0.581]} \\
gpt-oss-120b & \shortstack[r]{0.565 / 0.567\\{}[0.542, 0.593]} & \shortstack[r]{0.588 / 0.596\\{}[0.571, 0.622]} & \shortstack[r]{0.566 / 0.572\\{}[0.546, 0.597]} & \shortstack[r]{0.589 / 0.596\\{}[0.570, 0.622]} \\
\bottomrule
\end{tabular}
\end{adjustbox}
\end{table}

\input{tables/table_item_conditional_pairwise}

\input{tables/table_item_conditional_pairwise_prompt}

\input{tables/table_item_conditional_pairwise_open}

\input{tables/table_discordant_sensitivity}

\input{tables/table_discordant_item_profile}

\begin{table}[t]
\centering
\caption{\textbf{The fixed-prompt GPT-5-mini reader is evaluated on all ten tested combinations of the response, summary, trace, and prompt.} Across-item AUROC ranks runs across questions; within-item AUROC compares runs of the same question. Both estimates use the same 1{,}389 analyzed outcome-discordant MMLU-Pro questions, three gpt-oss-20b runs per item, and the fixed-prompt \texttt{gpt-5-mini-2025-08-07} reader. Brackets give paired 95\% item-bootstrap intervals (5{,}000 resamples). Each comparison subtracts the second named display from the first on every paired bootstrap draw.}
\label{tab:llm_judge_factorial}
\small
\begin{adjustbox}{max width=\textwidth}
\begin{tabular}{@{}L{6.5cm}R{3.7cm}R{3.7cm}@{}}
\toprule
Display or comparison & Across-item estimate [95\% CI] & Within-item estimate [95\% CI] \\
\midrule
\multicolumn{3}{@{}l}{\textit{Display AUROC}} \\
Response ($R$) & \shortstack[r]{0.501\\{}[0.485, 0.517]} & \shortstack[r]{0.503\\{}[0.486, 0.521]} \\
Summary ($S$) & \shortstack[r]{0.673\\{}[0.655, 0.690]} & \shortstack[r]{0.649\\{}[0.628, 0.669]} \\
Reasoning trace ($C$) & \shortstack[r]{0.682\\{}[0.667, 0.698]} & \shortstack[r]{0.683\\{}[0.663, 0.702]} \\
Response + summary & \shortstack[r]{0.676\\{}[0.659, 0.694]} & \shortstack[r]{0.656\\{}[0.635, 0.677]} \\
Response + reasoning trace & \shortstack[r]{0.700\\{}[0.684, 0.716]} & \shortstack[r]{0.690\\{}[0.670, 0.710]} \\
Response + summary + reasoning trace & \shortstack[r]{0.704\\{}[0.688, 0.721]} & \shortstack[r]{0.693\\{}[0.673, 0.713]} \\
\bottomrule
\end{tabular}
\end{adjustbox}
\end{table}

\clearpage
% ---- Provider-side replay features ----
\input{tables/table_internal_decomposition_15cell}

\begin{table}[H]
\centering
\caption{\textbf{The mean hidden-state gain is largest with the run's own trace and is lost under shuffled replay.} Held-out test AUROC of the $R{+}S{+}C{+}Z_{\mathrm{hidden}}$ monitor when the hidden state is extracted from replays of different contexts: the standard task prompt plus the run's reasoning trace ($q{+}C$), the prompt alone, the trace alone, the prompt plus final response, the prompt plus the aggressively masked trace, and the prompt plus a trace from a different item, reassigned using a fixed random seed. The public $R{+}S{+}C$ column repeats the comparison without replay. The headline describes the mean over the reported subset; the pattern need not hold in every setting. Values average five data splits ($\pm$ SD across data splits) for all five models on GPQA-Diamond and Qwen3-8B on MMLU-Pro and Omni-MATH-2-Filtered.}
\label{tab:z_replay_ablation}
\small
\begin{adjustbox}{max width=\textwidth}
\begin{tabular}{@{}L{2.0cm}L{1.8cm}R{1.7cm}R{1.7cm}R{1.7cm}R{1.7cm}R{1.7cm}R{1.7cm}R{1.7cm}@{}}
\toprule
Benchmark & Model & $R{+}S{+}C$ & $q{+}C$ & $q$ only & $C$ only & $q{+}R$ & $q{+}$masked $C$ & $q{+}$shuffled $C$ \\
\midrule
GPQA-Diamond & gpt-oss-120b & 0.770 $\pm$ 0.057 & 0.798 $\pm$ 0.069 & 0.767 $\pm$ 0.077 & 0.830 $\pm$ 0.058 & 0.789 $\pm$ 0.064 & 0.713 $\pm$ 0.044 & 0.672 $\pm$ 0.144 \\
GPQA-Diamond & gpt-oss-20b & 0.707 $\pm$ 0.096 & 0.771 $\pm$ 0.075 & 0.693 $\pm$ 0.094 & 0.730 $\pm$ 0.089 & 0.658 $\pm$ 0.079 & 0.708 $\pm$ 0.062 & 0.674 $\pm$ 0.063 \\
GPQA-Diamond & Qwen3-14B & 0.800 $\pm$ 0.058 & 0.774 $\pm$ 0.080 & 0.724 $\pm$ 0.102 & 0.797 $\pm$ 0.086 & 0.758 $\pm$ 0.132 & 0.780 $\pm$ 0.102 & 0.721 $\pm$ 0.059 \\
GPQA-Diamond & Qwen3-4B & 0.732 $\pm$ 0.067 & 0.738 $\pm$ 0.070 & 0.660 $\pm$ 0.080 & 0.698 $\pm$ 0.103 & 0.656 $\pm$ 0.041 & 0.746 $\pm$ 0.069 & 0.685 $\pm$ 0.032 \\
GPQA-Diamond & Qwen3-8B & 0.839 $\pm$ 0.062 & 0.858 $\pm$ 0.034 & 0.828 $\pm$ 0.060 & 0.849 $\pm$ 0.038 & 0.754 $\pm$ 0.056 & 0.854 $\pm$ 0.035 & 0.829 $\pm$ 0.052 \\
\addlinespace
MMLU-Pro & Qwen3-8B & 0.821 $\pm$ 0.004 & 0.861 $\pm$ 0.007 & 0.833 $\pm$ 0.010 & 0.857 $\pm$ 0.008 & 0.842 $\pm$ 0.005 & 0.857 $\pm$ 0.007 & 0.822 $\pm$ 0.006 \\
\addlinespace
Omni-MATH-2-Filtered & Qwen3-8B & 0.875 $\pm$ 0.022 & 0.912 $\pm$ 0.012 & 0.874 $\pm$ 0.019 & 0.907 $\pm$ 0.010 & 0.908 $\pm$ 0.014 & 0.914 $\pm$ 0.010 & 0.866 $\pm$ 0.018 \\
\bottomrule
\end{tabular}
\end{adjustbox}
\end{table}

\input{tables/table_replay_qc}

\clearpage
% ---- Transfer across models and benchmarks ----
\input{tables/table_transfer_matrix}

\clearpage
% ---- Additional diagnostics and audits ----
\begin{table}[H]
\centering
\caption{\textbf{Higher-access monitors concentrate errors in the flagged tail.} Error rate among retained runs when the monitor keeps the indicated fraction of test runs ranked by predicted correctness and flags the rest, averaged over data splits and the 15 settings. Coverage 1.0 is the unconditional error rate.}
\label{tab:selective_risk}
\small
\begin{adjustbox}{max width=\textwidth}
\begin{tabular}{@{}L{2.6cm}R{1.8cm}R{1.8cm}R{1.8cm}R{1.8cm}R{1.8cm}@{}}
\toprule
Level & 50\% & 70\% & 80\% & 90\% & 100\% \\
\midrule
$R$ & 0.243 & 0.266 & 0.277 & 0.297 & 0.316 \\
$R{+}S$ & 0.139 & 0.203 & 0.240 & 0.277 & 0.316 \\
$R{+}S{+}C$ & 0.114 & 0.185 & 0.227 & 0.274 & 0.316 \\
$R{+}S{+}C{+}Z$ & 0.098 & 0.168 & 0.217 & 0.267 & 0.316 \\
\bottomrule
\end{tabular}
\end{adjustbox}
\end{table}

\input{tables/table_qc_counts}
\begin{table}[H]
\centering
\caption{\textbf{GPT-5-mini agrees with the deterministic checker on exact or numeric matches and evaluates the remaining cases.} This comparison covers the stored Omni-MATH-2-Filtered correctness labels from all 20{,}905 judgments (4{,}181 items $\times$ five target models, primary run). The deterministic exact/numeric checker resolves 20{,}310 comparisons. Each percentage uses the relevant comparison count as its denominator; \textemdash{} marks cases where a share is not defined.}
\label{tab:judge_audit}
\small
\begin{adjustbox}{max width=\textwidth}
\begin{tabular}{@{}L{8.6cm}R{2.4cm}R{1.4cm}@{}}
\toprule
Quantity & Count & Share \\
\midrule
Judgments with a binary decision & 20{,}905 & 100.0\% \\
\quad Judged equivalent & 14{,}206 & 68.0\% \\
\quad Judged not equivalent & 6{,}699 & 32.0\% \\
\quad Uncertain or error cases & 0 & 0.0\% \\
\addlinespace
Comparisons resolved by the deterministic checker & 20{,}310 & \textemdash \\
\quad Judge--checker agreement & 14{,}146 & 69.7\% \\
\quad Agreement on deterministic exact/numeric matches & 7{,}859/7{,}865 & 99.9\% \\
\quad Deterministic mismatches judged equivalent & 6{,}158/12{,}445 & 49.5\% \\
\bottomrule
\end{tabular}
\end{adjustbox}
\end{table}

\input{tables/table_failure_stratification}

\input{tables/table_response_audit}

\begin{table}[H]
\centering
\caption{\textbf{Output censoring changes equal-weight headline AUROC by at most 0.010.} Blank-trace exclusion retains items whose whitespace-stripped reasoning trace is nonempty. The strict parsed-output rule retains items with parser status ``ok'' and nonempty whitespace-stripped values for the reasoning trace, response, and extracted answer. AUROC is recomputed within each setting and data split after censoring, averaged over five data splits within each model--benchmark setting, and then averaged equally over the 15 settings. $\Delta$ is censored minus uncensored AUROC; upper-block deltas are reported to four decimals because several are smaller than $0.001$, and the per-setting block uses three decimals. The final column reports retained / uncensored held-out predictions across items and data splits for each access level. The lower block reports per-setting deltas for the two Omni-MATH-2-Filtered settings with concentrated empty traces.}
\label{tab:censoring_sensitivity}
\small
\begin{adjustbox}{max width=\textwidth}
\begin{tabular}{@{}L{3.0cm}L{1.7cm}R{1.8cm}R{1.8cm}R{1.4cm}R{2.8cm}@{}}
\toprule
Censoring rule & Access level & Uncensored & Censored & $\Delta$ & Scored predictions \\
\midrule
Blank-trace exclusion & $R$ & 0.623 & 0.621 & $-$0.0024 & 80{,}789 / 82{,}100 \\
 & $R{+}S$ & 0.779 & 0.776 & $-$0.0030 & 80{,}789 / 82{,}100 \\
 & $R{+}S{+}C$ & 0.814 & 0.814 & $-$0.0002 & 80{,}789 / 82{,}100 \\
 & $R{+}S{+}C{+}Z$ & 0.840 & 0.839 & $-$0.0013 & 80{,}789 / 82{,}100 \\
\addlinespace
Strict parsed-output censor & $R$ & 0.623 & 0.613 & $-$0.0098 & 79{,}930 / 82{,}100 \\
 & $R{+}S$ & 0.779 & 0.772 & $-$0.0070 & 79{,}930 / 82{,}100 \\
 & $R{+}S{+}C$ & 0.814 & 0.810 & $-$0.0035 & 79{,}930 / 82{,}100 \\
 & $R{+}S{+}C{+}Z$ & 0.840 & 0.835 & $-$0.0049 & 79{,}930 / 82{,}100 \\
\midrule
\multicolumn{6}{@{}l}{\textit{Affected-setting $\Delta$ (censored minus uncensored AUROC)}} \\
Censoring rule & Setting & $R$ & $R{+}S$ & $R{+}S{+}C$ & $R{+}S{+}C{+}Z$ \\
\cmidrule(lr){1-6}
Blank-trace exclusion & Qwen3-8B / Omni-MATH-2-Filtered & $-$0.019 & $-$0.029 & $-$0.008 & $-$0.011 \\
 & Qwen3-14B / Omni-MATH-2-Filtered & $-$0.007 & $-$0.012 & $-$0.003 & $-$0.003 \\
\addlinespace
Strict parsed-output censor & Qwen3-8B / Omni-MATH-2-Filtered & $-$0.059 & $-$0.052 & $-$0.029 & $-$0.033 \\
 & Qwen3-14B / Omni-MATH-2-Filtered & $-$0.039 & $-$0.035 & $-$0.019 & $-$0.019 \\
\bottomrule
\end{tabular}
\end{adjustbox}
\end{table}

\begin{table}[H]
\centering
\caption{\textbf{Sparse models select both public feature sources.} Values are selection rates across the 15 model--benchmark settings and five random splits. Plain LASSO selects individual standardized features. Sparse group LASSO applies a penalty by feature source while retaining sparsity within each source. Both models include surface features, Cohere embeddings, and answer-probability features when available. Raw hidden states are evaluated separately in a sensitivity analysis using replay features.}
\label{tab:feature_selection_audits}
\small
\begin{adjustbox}{max width=\textwidth}
\begin{tabular}{@{}L{2.7cm}L{2.2cm}R{1.55cm}R{1.55cm}R{1.55cm}R{1.55cm}@{}}
\toprule
Selection method & Feature source & $R$ & $R+S$ & $R+S+C$ & $R+S+C+Z$ \\
\midrule
Plain LASSO & Surface & 86\% & 96\% & 99\% & 97\% \\
 & Cohere & 92\% & 100\% & 100\% & 100\% \\
 & $Z_{\mathrm{prob}}$ & \textemdash & \textemdash & \textemdash & 76\% \\
\addlinespace
Sparse group LASSO & Surface & 100\% & 97\% & 100\% & 100\% \\
 & Cohere & 93\% & 92\% & 75\% & 84\% \\
 & $Z_{\mathrm{prob}}$ & \textemdash & \textemdash & \textemdash & 79\% \\
\bottomrule
\end{tabular}
\end{adjustbox}
\end{table}

\input{tables/table_sparse_feature_examples}

\input{tables/table_summary_extract_features}

\end{document}